%% file: main.tex
\documentclass[10pt,twocolumn,letterpaper]{article}

\usepackage{3dv}

\usepackage[utf8]{inputenc} %
\usepackage[T1]{fontenc}    %
\usepackage{hyperref}       %
\usepackage{url}            %
\usepackage{booktabs}       %
\usepackage{amsmath}
\usepackage{amsfonts}       %
\usepackage{nicefrac}       %
\usepackage{microtype}      %
\usepackage[table,xcdraw,usenames, dvipsnames]{xcolor}         %
\usepackage{tocbibind}
\usepackage[toc,page]{appendix}
\usepackage{comment}
\usepackage{pifont}%
\usepackage{graphicx}
\newcommand{\cmark}{\textcolor{OliveGreen}{\ding{51}}}
\newcommand{\xmark}{\textcolor{red}{\ding{55}}}
\usepackage{times}
\usepackage[labelformat=simple]{subcaption}
\renewcommand\thesubfigure{\alph{subfigure})}
\usepackage{epsfig}
\usepackage{enumitem}
\usepackage{multirow}
\usepackage{arydshln}	%
\usepackage{rotfloat}
\usepackage{caption}
\floatstyle{plain} 
\restylefloat{figure}
\newcommand{\boldparagraph}[1]{\vspace{0.1em}\noindent{\bf #1} }

\newcommand{\figbottomspace}{3pt}

\threedvfinalcopy %
\def\threedvPaperID{138} %

\ifthreedvfinal\fi
\begin{document}

\title{NVS-MonoDepth: Improving Monocular Depth Prediction\\ with Novel View Synthesis}

\author{\hspace{-1em}
Zuria Bauer$^1$\!,
Zuoyue Li$^2$\!,
Sergio Orts-Escolano$^1$\!,
Miguel Cazorla$^1$\!,
Marc Pollefeys$^{2,4}$\!,
Martin R. Oswald$^{2,3}$\\[3pt]
$^1$University of Alicante \qquad $^2$ETH Zurich \qquad $^3$University of Amsterdam \qquad $^4$Microsoft\\
\hspace{-1.4em}
{\tt\small \{zuria.bauer, sorts, miguel.cazorla\}@ua.es; \{li.zuoyue, marc.pollefeys, moswald\}@inf.ethz.ch}
}

\maketitle
\thispagestyle{empty}

\begin{abstract}

Building upon the recent progress in novel view synthesis, we propose its application to improve monocular depth estimation.
In particular, we propose a novel training method split in three main steps.
First, the prediction results of a monocular depth network are warped to an additional view point.
Second, we apply an additional image synthesis network, which corrects and improves the quality of the warped RGB image.
The output of this network is required to look as similar as possible to the ground-truth view by minimizing the pixel-wise RGB reconstruction error.
Third, we reapply the same monocular depth estimation onto the synthesized second view point and ensure that the depth predictions are consistent with the associated ground truth depth.
Experimental results prove that our method achieves state-of-the-art or comparable performance on the KITTI and NYU-Depth-v2 datasets with a lightweight and simple vanilla U-Net architecture.

\end{abstract}

\vspace{-1em}

\section{Introduction} \label{sec:introduction}
Monocular depth estimation has been an important topic in the computer vision community for years, and has been widely applied in many tasks such as 3D reconstruction, semantic segmentation, object detection, autonomous driving, or improving super-resolution on low-quality images. 
Recently, several papers~\cite{niklaus2019ken, chen2019monocular, wiles2020synsin} which used novel view synthesis (NVS) have demonstrated that accurate geometry priors can well benefit the synthesis quality.
SynSin~\cite{wiles2020synsin} showed that a rough but plausible depth estimation can be obtained even without depth supervision and that a depth estimation network can obtain supervision signals from an end-to-end view synthesis pipeline.
Furthermore, the recent NVS methods using implicit representations~\cite{Mildenhall2020, Brualla2020} do not use the depth information to optimize a neural radiance field, but the geometry can still be well learned showing that view constraints can guide a network to learn the depth of the image.

In this paper, we focus on monocular depth prediction and propose a new training pipeline for monocular depth prediction architectures which uses novel view consistency constraints as a supervisory signal in addition to the common depth supervision.
To this end, our pipeline is trained with two neighboring viewpoints, which are simultaneously predicted during training to provide further network supervision via differentiable rendering of both the RGB image and the depth data.
We evaluate our method on various indoor and outdoor datasets and investigate the key elements of our pipeline, carrying out an ablation study that shows the benefits of the proposed training strategy.
Our approach yields lower RMSE scores than other state-of-the-art methods and outperforms them on all metrics on the NYU-Depth-v2~\cite{NYUV2} dataset, while yielding on par performance with the best method on the KITTI~\cite{Kitti} dataset.

\vspace{1em}
\noindent
Overall, our \textbf{contributions} are two-fold:
\begin{enumerate}[itemsep=0pt,topsep=3pt,leftmargin=*]
  \item We propose to use novel-view synthesis as an additional supervisory signal to improve the training of a monocular depth estimation network. 
  To this end, we propose two loss functions that augment the traditional depth supervision.
  \item We present comprehensive experiments on both indoor and outdoor datasets that demonstrate the benefits of our approach, as well as an ablation study which empirically justifies our design choices.
\end{enumerate}
We will make all source code and trained models publicly available upon publication to ensure reproducibility.

\section{Related Work} \label{sec:relatedworks}

This literature review focuses on the main related fields, namely depth estimation and novel view synthesis.

\boldparagraph{Depth estimation.}
Originally, the task of depth estimation from 2D image data relied on classical stereo vision~\cite{Silberman2012} approaches.
Below, we present a review of the monocular centered methods. %
Existing approaches can be roughly divided into supervised, self-supervised and unsupervised models. %
Here, we focus on supervised methods. %

A pioneering work which took advantage of supervised learning for depth estimation from a single monocular image appeared in 2005 by Saxena~\etal~\cite{Saxena2005}.
Their model used a discriminatively-trained Markov random field that incorporated local and global features from the image.
Since then, various approaches to provide additional consistency to the prediction have been tested.
The works of \cite{Liu2010, Wang2015, Wang2020} opted to use semantic segmentation as an additional key feature to guide the training. 
\cite{Karsch2012, Konrad2012, Kerl2013} built upon the hypothesis of similarity between image pairs, assuming that two similar images were more likely to also have a similar 3D structure.

Numerous works~\cite{Liu2015, Li2015, Roy2016, Laina2016, LiKY2016, Li2017, Gan2018, Medens2020} based their proposed architectures on fully convolutional neural networks (CNN).
\cite{Liu2015} paired the network with a learning scheme which learns the unary and pairwise potentials of continuous Conditional Random Fields (CRFs).
\cite{Li2015} used a DCNN model to map patches from multi-scale images.
\cite{Laina2016} proposed a modified ResNet-50 architecture with novel upsampling blocks allowing higher resolution. 
\cite{LiKY2016} captured scene details by considering information contained in depth gradients -- with this purpose, they proposed a fast-to-train two-streamed CNN to regress the depth and depth gradients.
\cite{Roy2016} combined CNNs with Regression Forest to predict depth.
\cite{Li2017} aimed to predict depth pixel-wise for a single color image.
\cite{Gan2018} integrated an affinity layer into a CNN, being able to combine learned absolute and relative features in a fully end-to-end model. 
\cite{Medens2020} proposed a new lightweight and fast supervised CNN architecture combined with novel feature extraction models which are designed for real-world autonomous navigation.
\cite{shu2020featdepth} proposed a feature-metric loss defined on the feature-level representation in their entangled dual task of depth and pose estimation, leading to an improved accuracy and demonstrated its effectiveness.

\cite{Yuanzhouhan2016, Xu2018, Wang2015, Liu2015} made use of the potentials of continuous CRFs to fuse multi-scale information obtained from different CNN layers that are then used to obtain  the  final  depth  estimation.
\cite{Jung2017, Huang2017} took advantage of recent advances in Generative Adversarial Networks (GANs)~\cite{GANs} to sequentially estimate global and local structures of the depth images.
\cite{Facil2019, Wu2019Cam} made use of the internal camera parameters to improve the generalization capabilities of the architectures.

\cite{Eigen2014} introduced a coarse-scale network refined by a fine-scale network.
Both networks were applied to the original input, but, in addition, the coarse network’s output is passed to the fine network as additional first-layer image features. %
\cite{Fu2018} introduced a network architecture to obtain high-resolution depth maps, using a spacing-increasing discretization (SID) strategy for depth and recasting depth network learning as an ordinal regression problem. 
\cite{Lee2019} introduced a network composed of a dense feature extractor (the base network), a contextual information extractor (ASPP), local planar guidance layers and their dense connection for final depth estimation. 

AdaBins~\cite{AdaBins2020} proposed a transformer-based architecture block that divides the depth range into bins whose center value is estimated adaptively per image.
The final depth values are estimated as linear combinations of the bin centers.
Until now, this work defines the state of the art on the monocular depth prediction benchmarks.
Finally, \cite{Song2021} proposed a simple but effective scheme by incorporating the Laplacian pyramid into the decoder architecture.

While most previous works proposed various network architectures for performance improvements, we propose a novel training scheme.
Our method is complementary to most approaches and can potentially be used with many architectures to augment their training.
Moreover, a common practice in monocular depth estimation is the use of a deep, often pretrained encoder network like VGG-16~\cite{Eigen2015}, ResNet-50~\cite{Laina2016,Godard2016,Kuznietsov2017,Gan2018,Ramamonjisoa2019,Song2021}, ResNet-101~\cite{Fu2018}, ResNext-101~\cite{Yin2019,Lee2019}, or SeNet-154~\cite{Hu2018,Chen2019structure} with large parameter counts.
In our experiments, we show that a simple U-Net~\cite{Ronneberger2015} without any architectural modifications can achieve state-of-the-art or comparable performance using our proposed training scheme.

\boldparagraph{Novel view synthesis.}
Traditional novel view synthesis approaches are typically based on multi-view reconstruction using a geometric formulation \cite{Debevec96, Fitzgibbon2003, Seitz2006}, while most recent approaches rely more on deep neural networks which can even make predictions from a single input image.
Some works do not directly build geometry for the source image, but resort to the scene flows. 
\cite{sun2018multiview} directly output flows followed by a weighted aggregation based on the self-learned confidence.
\cite{park2017transformation} performs transformation on the 3D latent space, using the target view's coarse RGB and visibility map as intermediate representations. 
\cite{chen2019monocular} follows such a transformation, but supervises the network to predict the target depth and further calculate the flows.
More works have better performance using to build geometry. 
\cite{niklaus2019ken} generates Ken Burns effects based on the supervised depth prediction on the source image. 
\cite{wiles2020synsin} shows that a plausible dense depth map can even be obtained through its end-to-end pipeline without depth supervision demonstrating the benefits of novel view synthesis for geometric reconstruction.

In contrast to our hybrid approach, \cite{Godard2016, Zhou2017, Wang2017} use novel view synthesis as main supervision and are fully unsupervised. The main difference to our approach can be condensed to their explicit usage of novel view synthesis to create stereo/multi-view depth predictions, while our approach only uses them for consistency at training time.

A recent popular NVS approach is via a neural radiance field, NeRF~\cite{Mildenhall2020} and many derived works such as \cite{Brualla2020} which used multi-view data to train a latent appearance code to enable learning a neural scene representation.
\cite{Park2020} used deformable videos using a second MLP applying a deformation for each frame of the video.
\cite{Yu2020} proposed a multi-image approach at test time. 
\cite{Liu2020} organized the scene into a sparse voxel octree to speed up rendering by a factor of 10.
\cite{Zhang2020} handled unbounded scenes.
NeRF-based models typically require several input images and expensive optimization, which makes them less suitable for our targeted monocular depth prediction problem.
Rather than striving for perfect novel view synthesis, we merely use the technology as a tool for better network training.

\begin{figure*}[!ht]
    \vspace*{-5pt}
	\includegraphics[width=\textwidth]{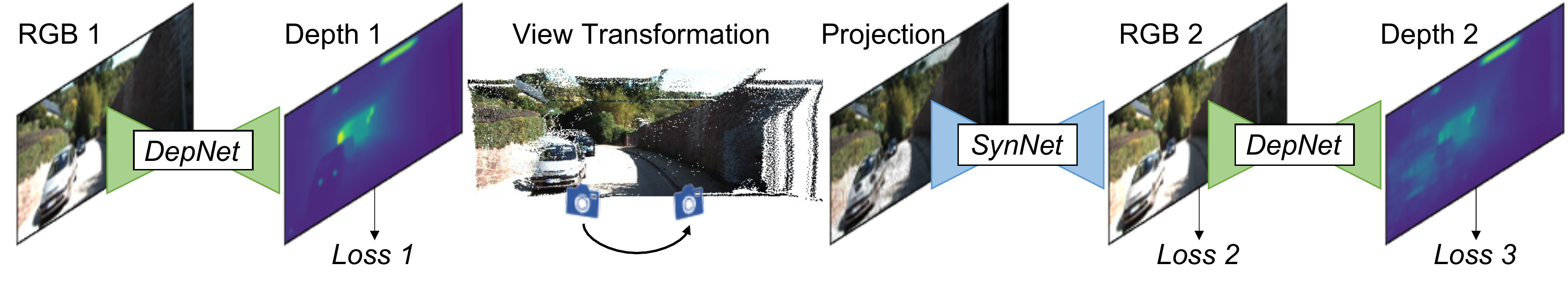}\\
	\vspace*{-24pt}
	\caption{\textbf{Overview of our pipeline.} The figure shows the training scheme of the pipeline with two additional loss function that augment the traditional supervised depth loss (Loss 1).
	The depth prediction of \textit{DepNet} is used to warp the reference view (RGB 1) to a target view (RGB 2).
	Due to occlusions and missing data this image gets completed with \textit{SynNet} to provide a high-quality image prediction for the target view point and is supervised with RGB 2 (Loss 2). By applying the same \textit{DepNet} again on the prediction of the target view we can further supervise the network with its corresponding GT depth.
	During test time, only \textit{DepNet} is used for the monocular depth estimation. RGB 1 and RGB 2 are typically consecutive view points from a video or corresponding stereo images.}
	\label{fig:pipeline}
\end{figure*}

\section{Methodology} \label{sec:methodology}

The main motivation for our approach is to provide additional supervisory signals from novel view synthesis constraints.
The prediction of monocular depth networks may sometimes lack geometrical consistency with other view points.
We address this issue by using novel view synthesis to provide additional geometrical consistency.%
The proposed pipeline follows a simple design, using light architectures to avoid memory shortages. 
Also, the structure of our pipeline is flexible with respect to user needs, since additional losses can be simply added or removed. 
A detailed analysis of all these elements can be found below. 

\subsection{Pipeline} \label{subsec:pipeline}
We provide a detailed overview of our proposed pipeline in Figure~\ref{fig:pipeline}.
The pipeline consists of three major building blocks (a monocular depth prediction network -- \textit{DepNet}, a View Transformation procedure, and an image synthesis network -- \textit{SynNet}) aligned to provide additional consistency to the basic monocular network.
The data flow through the pipeline is as follows.
The RGB image from the source viewpoint (RGB 1) is used as the input of the depth network (\textit{DepNet}). 
The output depth prediction together with the known camera intrinsics and the pose (extrinsics) are utilized to create a 3D point cloud.
This 3D representation is then reprojected directly to the target view point (RGB 2) to obtain the initial 2D projection, which is used as the input for the synthesis network (\textit{SynNet}). 
\textit{SynNet} mainly fills occlusions and holes that were created due to the reprojection such that a complete RGB image of the target viewpoint (RGB 2) is estimated.
Finally, the synthesized RGB of the target viewpoint is again used as an input for the same monocular depth network (\textit{DepNet}).
With the depth prediction for the second RGB viewpoint, we have an additional loss function to supervise the depth prediction network.
In the following we detail all building blocks and provide additional information about the loss functions used for training the pipeline.

\boldparagraph{Monocular Depth Network -- \textit{DepNet}.}
Since the proposed pipeline uses two networks which can create memory shortages we used light architectures, with a small number of layers.
To this end, we utilize a standard U-Net~\cite{Ronneberger2015} architecture.
It consists of a downsampling path and an upsampling path which give the U-shaped form to the network.
The encoder and decoder part of the U-Net
are connected with skip connections such that the upsampling path has additional concatenations with the high-resolution features obtained during the downsampling.
Some benefits of this architecture are that it is lightweight, it can be effectively trained with a small amount of images and has proven to be able to outperform state-of-the art networks in its field. 
Another advantage of this architecture is its fast training and inference capability, e.g. the depth prediction inference for a $256\times768$ image from the test set, takes an average time of 36ms on an Nvidia 1080Ti. %

Thus, the monocular depth architecture is a standard U-Net~\cite{Ronneberger2015}, with input size $256\times256\times3$ for the Replica~\cite{Replica} and NYU-Depth-v2~\cite{NYUV2} datasets and a size of $256\times768\times3$ for the KITTI~\cite{Kitti} dataset. 
The output sizes are $256\times256\times1$ and $256\times768\times1$, respectively. 
The network has a depth of 7 layers, downsampling the image to $1\times1\times1024$ and $1\times3\times512$. 
Downsampling the image to this small size for any other task than classification is not needed, since the final feature does not provide enough information for the network.
Nevertheless, we experimented with fewer layers, downsampling the image to a size of $8\times8$, and the random artifacts the network created on the output were much higher than with downsampling the image to $1\times1$. 
In between the layers are the skip connections that provide high-resolution features to the decoder part.
A more detailed description of the architectures can be found in the supplementary material.

\boldparagraph{View Transformation.}
From the predicted depth, 3D points in world coordinates are obtained through a unprojection procedure using the intrinsics of the camera and its pose (extrinsics). 
In the re-projection procedure, the points' world coordinates are first transformed to the coordinates of the novel view, followed by a conventional z-buffer to obtain the 2D image. 
The pixels without any projected points are initialized with zero for each channel.

\boldparagraph{Synthesis Network -- \textit{SynNet}.}
The third key component of the proposed pipeline is an image synthesis network.
The purpose of this network is to improve the quality of the input image after warping it to the second view point.
Due to disocclusions and missing data, the output of the view transformation might be incomplete and noisy.
Example input, outputs pairs of this network are later shown in Fig.~\ref{fig:Pipeline0}.
Since we had the same prerequisites as for the monocular depth architecture, we decided on a similar architecture for this network.
The synthesis network is a standard U-Net~\cite{Ronneberger2015} architecture with also 7 layers.
The input size of the network is $256\times256\times3$ or $256\times768\times3$ respectively for the Replica~\cite{Replica}, NYU-Depth-v2~\cite{NYUV2} and KITTI~\cite{Kitti} datasets.
The output of the network are RGB images with size $256\times256\times3$ or, $256\times768\times3$ respectively.

\subsection{Loss functions}  \label{subsec:loss}
In the following we detail all utilized loss functions.

\boldparagraph{Pixel-wise losses.}
For the monocular depth network loss, $\mathcal{L}_1$ we use the traditional L1 loss for ground truth supervision: 
\begin{equation}
  \mathcal{L}_1 = \frac{1}{N}\sum_{i=1}^{N} |y_i - \hat{y_i}| \enspace .
\end{equation}
The loss summarizes the absolute pixel-wise differences between the target depth prediction $y_i$ and the predicted depth map $\hat{y_i}$.
This value is the divided by the number of valid pixels $N$ -- those valid pixels are extracted from the ground truth depth map, excluding the locations where no information was provided.

The loss to train the synthesis network ($\mathcal{L}_2$) is an L1 loss between the target RGB image and the RGB image predicted by the network.
In this case, we used the entirety of the image pixels without any masking.

The final loss applied to the output of the pipeline after the second depth prediction ($\mathcal{L}_3$) aims to preserve the accuracy of the second viewpoint and also provide an additional discrimination factor on the synthesis network.
This is also an L1 loss function, but it is applied to the second view point instead of the first one.
This third loss supervises the same \textit{DepNet} twice, before and after the view transformation, but due to the view transformation in the loop, small prediction errors may lead to larger prediction errors in the second view after the view transformation.
In sum, this defines a loss for the depth prediction network which is much more sensitive to small view point variations than the standard L1 loss on the input view alone. 

\boldparagraph{Overall loss.}
The overall loss function is the weighted sum of all the previously listed individual loss functions. %
The weights can be used to control the training by enhancing or diminishing the importance of different networks, \eg, one can choose to enhance the depth network and provide less importance to the synthesis.
This loss function guarantees backpropagation over the entire pipeline and can be formalized as follows:
\begin{equation}
  \mathcal{L} = \alpha \cdot \mathcal{L}_1 + \beta \cdot \mathcal{L}_2 + \alpha \cdot \mathcal{L}_3 \enspace , %
\end{equation}
where $\alpha$ and $\beta$ are the weighting hyper-parameters. %
\begin{table*}[!htb]
    \vspace*{-5pt}
  \centering
  \footnotesize
  \setlength{\tabcolsep}{7.7pt} %
  \newcommand{\ul}[1]{{\textcolor{blue}{{#1}}}}
  \begin{tabular}{llcccccccc}
    \toprule
    Model & Backbone & \#params (M)$\downarrow$ & REL$\downarrow$ & RMSE$\downarrow$ & RMSE$_\text{log}\!\!\downarrow$ & Sq.Rel.$\downarrow$ & $\delta_1\!\uparrow$ & $\delta_2\!\uparrow$ & $\delta_3\!\uparrow$\\
    \midrule
    Saxena~\etal~\cite{Saxena2009}         &         -       &  -  & 0.280 & 8.734 & 0.361 & 3.012 & 0.601 & 0.820 & 0.926 \\
    Eigen~\etal~\cite{Eigen2014}           &         -       &  -  & 0.190 & 7.156 & 0.270 & 1.515 & 0.692 & 0.899 & 0.967 \\
    Liu~\etal~\cite{Liu2015}               &         -       &  40 & 0.217 & 6.986 & 0.287 & 1.841 & 0.647 & 0.882 & 0.961 \\
    Godard~\etal~\cite{Godard2016}         & ResNet-50       &  31 & 0.085 & 3.938 & 0.135 & 0.427 & 0.916 & 0.980 & 0.994 \\
    Kuznietsov~\etal~\cite{Kuznietsov2017} & ResNet-50       &  -  & 0.138 & 3.610 & 0.138 & \bf 0.121 & 0.906 & 0.989 & 0.995 \\
    Gan~\etal~\cite{Gan2018}               & ResNet-50       &  -  & 0.098 & 3.933 & 0.173 & 0.666 & 0.890 & 0.964 & 0.985 \\
    Fu~\etal~\cite{Fu2018}                 & ResNet-101      & 110 & 0.072 & 2.727 & 0.120 & 0.307 & 0.932 & 0.984 & 0.994 \\
    Yin~\etal~\cite{Yin2019}               & ResNeXt-101     & 114 & 0.072 & 3.258 & 0.117 &   -   & 0.938 & 0.990 & \ul{0.998} \\
    BTS~\cite{Lee2019}                     & ResNeXt-101     & 113 & 0.064 & 2.540 & 0.100 & 0.254 & 0.950 & 0.993 & \bf 0.999 \\
    Song~\etal~\cite{Song2021}             & ResNet-50       &  -  & 0.059 & \ul{2.446} & 0.091 & 0.212 & 0.962 & \ul{0.994} & \bf 0.999 \\
    AdaBins~\cite{AdaBins2020}             & EfficientNet-B5 &  78 & 0.058 &  \bf 2.360 & \bf 0.088 & \ul{0.190} & \bf 0.964 & \bf 0.995 & \bf 0.999\\[0.3pt]
    \hdashline \noalign{\vskip 2pt}
    U-Net baseline (\textit{DepNet})       & U-Net           &  54 & \ul{0.057} & 3.023 & 0.104 & 0.441 & 0.936 & 0.975 & 0.991 \\
    \hdashline \noalign{\vskip 2pt}
    \textbf{NVS-MonoDepth (Ours)}          & U-Net           &  54 & \bf 0.031 & 2.702 & \ul{0.089} & 0.292 & \ul{0.963} & 0.989 & 0.997 \\[0.3pt]
    \bottomrule
  \end{tabular}
  \vspace{-7pt}
  \caption{\textbf{State-of-the-art comparison on the KITTI~\cite{Kitti} dataset.} For reference, we additionally show the results of our U-Net baseline in the second-to-last row, which is the same network, but trained without the proposed NVS losses. The reported numbers are from the corresponding original papers. Best results are shown in \textbf{bold} and second best results in \ul{blue}.}
  \label{tab:kitti_results}
  \vspace*{\figbottomspace}
\end{table*}

\section{Experiments}  \label{sec:experimentation}
We carried out a variety of monocular depth estimation experiments on indoor as well as outdoor scenes.  
We first briefly describe the data and evaluation metrics, then we present quantitative comparisons to state-of-the-art monocular depth estimation methods.

All our experiments have been carried out on two GPUs: an Nvidia 1080Ti and an Nvidia Titan X.
Our method was implemented in TensorFlow~\cite{tensorflow2015}.
For training, we used Adam as the optimizer, with a learning rate of $10^{-4}$ and a batch size of 8 to comply with the GPU memory limitations.
The typical training span for one epoch is between 20 and 30 minutes.

Note that all the qualitative results shown in this paper are images from the Eigen test splits of the KITTI~\cite{Kitti} or NYU-Depth-v2~\cite{NYUV2} datasets. 
In case of the Replica~\cite{Replica} dataset, we selected our own unseen test split (details are provided in Section~\ref{subsec:datasets}).

\subsection{Datasets and evaluation metrics} \label{subsec:datasets}
\boldparagraph{Datasets.}
To train and assess our method, we considered a variety of datasets used for monocular depth estimation.
Since we propose to train with multiple input images with overlapping view points we focus on multi-view datasets with ground truth depth.

We first discuss a number of datasets suitable for our multi-view setting which are mostly not used for monocular depth estimation.

The main focus when choosing a dataset relied on the size of the dataset, if it provides stereo pairs or sequences, if camera poses are provided for each frame, and if the dataset is outdoor or indoor.
From the publicly available datasets the most suitable for indoor scenarios were: SceneNET-RGBD~\cite{SCENENET}, Matterport~3D~\cite{Matterport3D}, SUNCG~\cite{SUNCG}, ScanNet~\cite{SCANNET}, Sun3D~\cite{SUN3D}, InteriorNet~\cite{InteriorNet18}, Middlebury~\cite{Middlebury} and Replica~\cite{Replica}.
The standard benchmark dataset NYU-Depth-v2~\cite{NYUV2} does not originally provide camera pose information, but since it is the benchmark dataset for monocular depth prediction, we wanted to include this dataset and used the camera pose information provided by \cite{Teed2018} for 13776 samples pairs.
We trained our method on those samples and afterwards evaluated on the Eigen~\cite{Eigen2014} split provided for the NYU-Depth-v2~\cite{NYUV2} dataset.
Additionally, we used the Replica dataset~\cite{Replica}, to train our pipeline on a synthetic dataset with perfect camera poses.
We split the provided 18 scenes in train (15), validation (2) and test sets (1). 
We extracted from the scenes and camera motion paths pairs of images with a maximum camera movement of 5 degrees in all the possible directions, including zoom.
The network was trained with 10k images for training, and was validated and tested on 1k images that were never seen during training. 

For the outdoor datasets, suitable ones were: Synthia~\cite{Synthia}, %
KITTI~\cite{Kitti}, UASOL~\cite{UASOL}, ETH3D~\cite{ETH3D} and Tanks and Temples~\cite{TanksAndTemples}.
Synthia~\cite{Synthia} is also a synthetic dataset for indoor environments. %
ETH3D~\cite{ETH3D} was not considered as it provides too few images to train the network properly. %
Tanks and Temples~\cite{TanksAndTemples} mostly provides scans of objects with only few complete scenes, making it less suitable for our setting.
UASOL~\cite{UASOL} would be a suitable dataset, but the camera poses provided by the dataset are position tracking which uses the first frame of each sequence as the fixed world coordinates and a vision-based algorithm is used to compute the rotation and translation between two consecutive frames.
Since these results are less accurate than the camera poses provided by the KITTI~\cite{Kitti} dataset, we decided to perform our experimentation on that dataset.
Moreover, it is also one of the monocular depth benchmark datasets.
To train the network on the KITTI~\cite{Kitti} dataset, we used the specific Eigen~\cite{Eigen2014} split which consist on 22k images for training, 888 for validation and 689 for test.

The main qualitative experimentation has been carried out on the KITTI~\cite{Kitti} dataset. 

\boldparagraph{Evaluation metrics.}
To evaluate the network against the state-of-the-art, we used the metrics from \cite{Eigen2014}.
Given the ground truth depth image $y$ with $N$ valid pixels and the predicted depth image $\hat{y}$, the metrics are defined as follows: 
relative error (REL): $\frac{1}{N} \sum_{i=1}^N \frac{|y_i - \hat{y_i}|}{y}$; 
root mean squared error (RMSE): $\sqrt{\frac{1}{N}\sum_{i=1}^N(y_i - \hat{y_i})^2}$; 
log root mean squared error (RMSE$_\text{log}$): $\sqrt{\frac{1}{N}\sum_{i=1}^N|\log{y_i}-\log{\hat{y_i}}|^2}$; 
squared relative difference (Sq. Rel.): $\frac{1}{N}\sum_{i=1}^N\frac{|y_i - \hat{y_i}|^2}{y_i}$; and 
threshold accuracy ($\delta_j$): fraction of $y_i$ such that $max(\frac{y_i}{\hat{y_i}}, \frac{\hat{y_i}}{y_i})=\delta<1.25^j$ for $j \in \{1, 2, 3\}$.

\subsection{Comparison to the state of the art} \label{subsec:stateofthear}
\boldparagraph{Quantitative results.}
We provide quantitative depth prediction results for the KITTI~\cite{Kitti} and NYU-Depth-v2~\cite{NYUV2} datasets compared to state-of-the-art methods.
Since the Replica~\cite{Replica} dataset is not part of the monocular depth benchmark, we did not assess  other state-of-the-art methods on this dataset but performed ablations studies.

For the KITTI~\cite{Kitti} and NYU-Depth-v2~\cite{NYUV2} datasets, we calculated the evaluation metrics by clipping the prediction values to the maximum depth of the sensors that filmed those datasets. 
In case of the KITTI~\cite{Kitti} dataset, the minimum value is 0.01 meters and the maximum 80 meters.
For the NYU-Depth-v2~\cite{NYUV2}, the minimum value is 0.01 meters and the maximum is 10 meters.
This clipping is also common in previous works such as \cite{Fonder2021M4Depth}, \cite{Laina2016}, \cite{AdaBins2020} or \cite{Lee2019}.

\boldparagraph{KITTI dataset.} 
Table~\ref{tab:kitti_results} lists the performance metrics on the KITTI~\cite{Kitti} dataset.
All the networks were trained on the KITTI-Eigen training set and evaluated on the KITTI-Eigen test set.
On this dataset, our method yields comparable performance to the currently best state-of-the-art method.
Our method yields a significantly lower relative error (REL) on all datasets.
Based on the RMSE results, it can be noticed that our network performs worse with the larger residuals but performs better for small ones as indicated by the RMSE$_\text{log}$ score and by the delta values - for $\delta_1$ and $\delta_3$ the performance is $0.1\%$ worse than the state-of-the-art method, and for $\delta_2$, $0.6\%$ worse. 
Note that the number of network parameters is substantially smaller than the ones of the best performing approaches.

\boldparagraph{NYU-Depth-v2 dataset.} 
Table~\ref{tab:nyu_results} lists the performance metrics on the NYU-Depth-v2~\cite{NYUV2} dataset. 
Despite the simple network architecture, our method outperforms all state-of-the-art networks trained on this dataset which indicates the effectivity of the proposed training scheme for monocular depth estimation.
The proposed network is able to predict $4.8\%$ more accurate than the state-of-the-art network, having also the smallest RMSE results. 
On the other side, the proposed method also achieves the highest results on all the $\delta$ thresholds, proving its effectiveness also on indoor datasets.

\begin{table}[tb]
    \vspace*{-5pt}
  \centering
  \scriptsize
  \setlength{\tabcolsep}{2pt}
  \renewcommand{\arraystretch}{1.1}
  \newcommand{\ul}[1]{{\textcolor{blue}{{#1}}}}
  \begin{tabular}{llcccccccc}
    \toprule
    Model & Backbone & \#params$\downarrow$ & REL$\downarrow$ & RMSE$\downarrow$ & $\delta_1\!\uparrow$ & $\delta_2\!\uparrow$ & $\delta_3\!\uparrow$\\
    \midrule
    Eigen~\etal~\cite{Eigen2014}        &         -       & 141 M & 0.158 & 0.641 & 0.769 & 0.950 & 0.988 \\
    Laina~\etal~\cite{Laina2016}        & ResNet-50       &  64 M & 0.127 & 0.573 & 0.811 & 0.953 & 0.989\\
    Hao~\etal\cite{Hao2018}             & ResNet-101      &  60 M & 0.127 & 0.555 & 0.841 & 0.966 & 0.991\\
    Lee~\etal~\cite{Lee2011}            &         -       & 119 M & 0.131 & 0.538 & 0.837 & 0.971 & 0.994\\
    Fu~\etal~\cite{Fu2018}              & ResNet-101      & 110 M & 0.115 & 0.509 & 0.828 & 0.965 & 0.992\\
    SharpNet~\cite{Ramamonjisoa2019}    & ResNet-50       &  80 M & 0.139 & 0.502 & 0.836 & 0.966 & 0.993\\
    Hu~\etal~\cite{Hu2018}              & SENet-154       & 157 M & 0.115 & 0.530 & 0.866 & 0.975 & 0.993\\
    Chen~\etal~\cite{Chen2019structure} & SENet-154       & 210 M & 0.111 & 0.514 & 0.878 & 0.977 & 0.994\\
    Yin~\etal~\cite{Yin2019}            & ResNeXt-101     & 110 M & 0.108 & 0.416 & 0.875 & 0.976 & 0.994\\
    BTS~\cite{Lee2019}                  & DenseNet-161    &  47 M & 0.110 & 0.392 & 0.885 & 0.978 & 0.994\\
    DAV~\cite{Huynh2020}                & DRN-D-22        &  25 M & 0.108 & 0.412 & 0.882 & 0.980 & \ul{0.996}\\
    AdaBins~\cite{AdaBins2020}          & EfficientNet-B5 &  78 M & \ul{0.103} & \ul{0.364} & \ul{0.903} & \ul{0.984} & \bf 0.997\\
    \hdashline \noalign{\vskip 2pt}
    U-Net (\textit{DepNet})             & U-Net           &  54 M & 0.132 & 0.571 & 0.815 & 0.839 & 0.854 \\
    \hdashline \noalign{\vskip 2pt}
    \textbf{Ours}                       & U-Net           &  54 M & \bf 0.058 & \bf 0.331 & \bf 0.989 & \bf 0.995 & \bf 0.997\\
    \bottomrule
  \end{tabular}
  \vspace{-7pt}
  \caption{\textbf{State-of-the-art comparison on the NYU-Depth-v2~\cite{NYUV2} dataset.} Please note the substantial reduction of the relative error by our approach. The reported numbers are from the corresponding original papers. Best results are shown in \textbf{bold} and second best results in \ul{blue}. }
  \label{tab:nyu_results}
  \vspace*{-1pt}
\end{table}

\boldparagraph{Replica dataset.} 
Table~\ref{tab:ablation_study} lists the performance metrics on the Replica~\cite{Replica} dataset. 
All networks were trained on 15 of the rooms (Apartments 0,1 and 2; frl apartments 1 to 5, office 0 to 3, rooms 1 and 2 and hotel 0), validated on 2 rooms (Office 3 and 4) and tested on 1 room (Room), the amount of data used for each split is provided in Section~\ref{subsec:datasets}.
The scores show reasonable performance on this synthetic indoor dataset as well.
As can be seen in this case, the network proves to be able to improve on large (according to the RMSE) and on small residuals (according to delta values).

\begin{figure}[!tb]
    \centering
    \scriptsize
    \setlength{\tabcolsep}{1pt}
	\renewcommand{\arraystretch}{0.8}
	\newcommand{\sz}{0.154}
	\newcommand{\sh}{1.2cm}
	\newcommand{\gcs}{\hspace{8pt}}  %
	\begin{tabular}{ccc@{\gcs}ccc}
	    RGB & GT & Ours & RGB & GT & Ours \\
        \includegraphics[width=\sz\columnwidth,height=\sh]{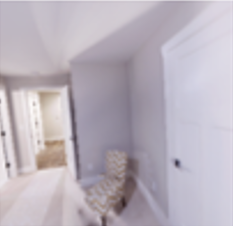} &
        \includegraphics[width=\sz\columnwidth,height=\sh]{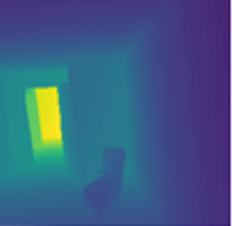} &
        \includegraphics[width=\sz\columnwidth,height=\sh]{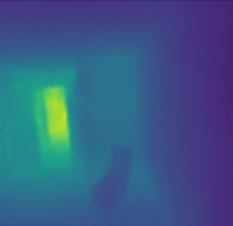} &
        \includegraphics[width=\sz\columnwidth,height=\sh]{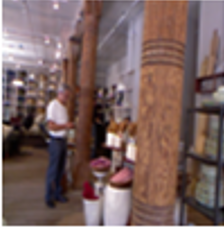} &
        \includegraphics[width=\sz\columnwidth,height=\sh]{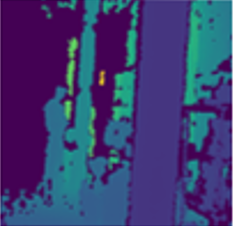} &
        \includegraphics[width=\sz\columnwidth,height=\sh]{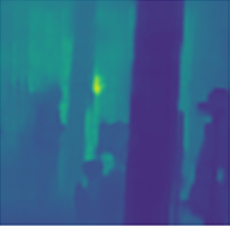} \\
        \includegraphics[width=\sz\columnwidth,height=\sh]{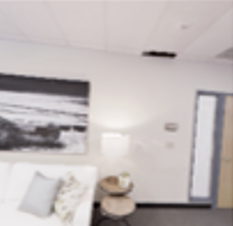} &
        \includegraphics[width=\sz\columnwidth,height=\sh]{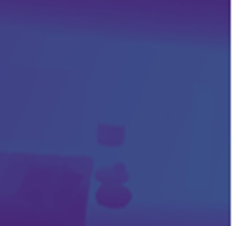} &
        \includegraphics[width=\sz\columnwidth,height=\sh]{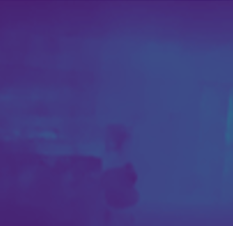} &
        \includegraphics[width=\sz\columnwidth,height=\sh]{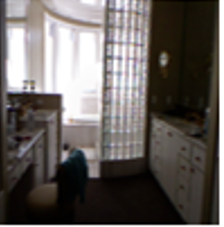} &
        \includegraphics[width=\sz\columnwidth,height=\sh]{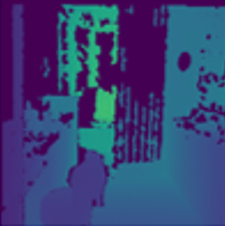} &
        \includegraphics[width=\sz\columnwidth,height=\sh]{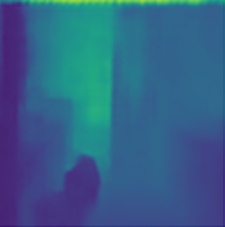} \\
        \includegraphics[width=\sz\columnwidth,height=\sh]{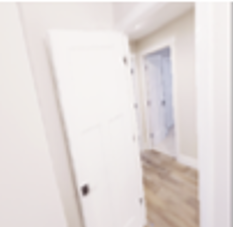} &
        \includegraphics[width=\sz\columnwidth,height=\sh]{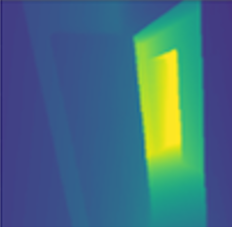} &
        \includegraphics[width=\sz\columnwidth,height=\sh]{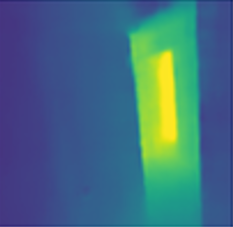} &
        \includegraphics[width=\sz\columnwidth,height=\sh]{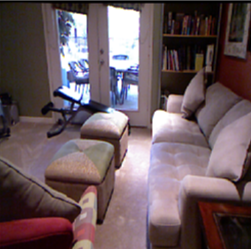} &
        \includegraphics[width=\sz\columnwidth,height=\sh]{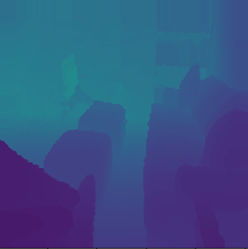} &
        \includegraphics[width=\sz\columnwidth,height=\sh]{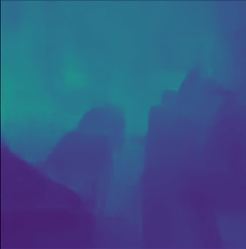} \\
        \multicolumn{3}{c}{Replica~\cite{Replica}} & \multicolumn{3}{c}{NYU-Depth-v2~\cite{NYUV2}}\\[-7pt]
    \end{tabular}    
    \vspace*{-3pt}
    \caption{\textbf{Qualitative results on the Replica~\cite{Replica} and NYU-Depth-v2~\cite{NYUV2} datasets.} We show the predictions of our proposed method compared to the GT images, proving its effectiveness on real and synthetic indoor datasets. Color scale: 0 (purple) to 80 meters (yellow).}
    \label{fig:qualitative}
\end{figure}

\begin{figure*}[!htb]
\vspace*{-10pt}
    \centering
    \scriptsize
    \setlength{\tabcolsep}{1pt}
	\renewcommand{\arraystretch}{0.8}
	\newcommand{\sz}{0.326}
	\newcommand{\sh}{1.38cm}
    \begin{tabular}{lccc}
        \rotatebox{90}{\hspace{9pt} RGB} &
        \includegraphics[width=\sz\linewidth, height=\sh]{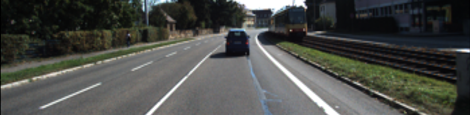}
        \includegraphics[width=\sz\linewidth, height=\sh]{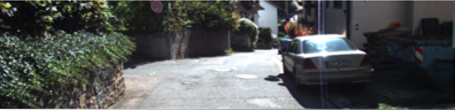}
        \includegraphics[width=\sz\linewidth, height=\sh]{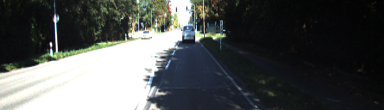}\\
        \rotatebox{90}{\hspace{11pt} GT} &
        \includegraphics[width=\sz\linewidth, height=\sh]{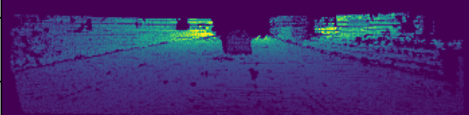}
        \includegraphics[width=\sz\linewidth, height=\sh]{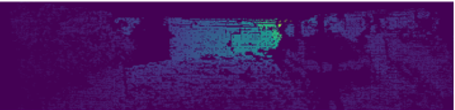}
        \includegraphics[width=\sz\linewidth, height=\sh]{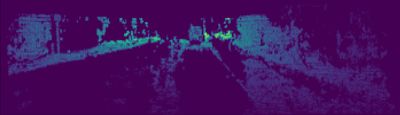}\\
        \rotatebox{90}{\hspace{10pt} Ours} &
        \includegraphics[width=\sz\linewidth, height=\sh]{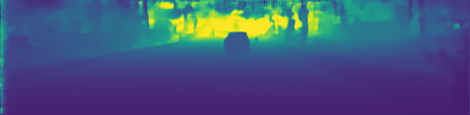}
        \includegraphics[width=\sz\linewidth, height=\sh]{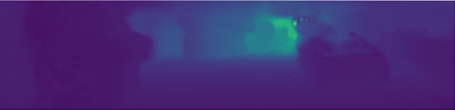}
        \includegraphics[width=\sz\linewidth, height=\sh]{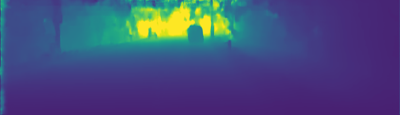}\\
        \rotatebox{90}{\hspace{-2pt} AdaBins~\cite{AdaBins2020}} &
        \includegraphics[width=\sz\linewidth, height=\sh]{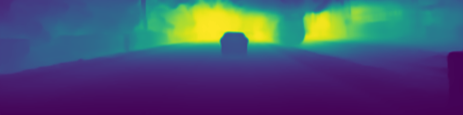}
        \includegraphics[width=\sz\linewidth, height=\sh]{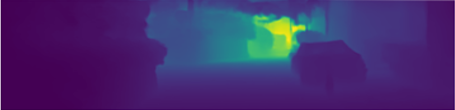}
        \includegraphics[width=\sz\linewidth, height=\sh]{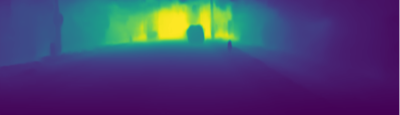}\\[-5pt]
    \end{tabular}
      \vspace{-4pt}
    \caption{\textbf{Qualitative predictions on the KITTI \cite{Kitti} dataset.} We show the performance of \textbf{NVS-MonoDepth} (Ours) compared to AdaBins~\cite{AdaBins2020} leading to similar predictions, but with a much simpler architecture. Color scale: 0 (purple) to 80 meters (yellow).
    }
    \label{fig:qualitative_compared}
    \vspace*{2pt}
\end{figure*}

\boldparagraph{Qualitative results.} 
Qualitative results of the proposed method are provided in Figure~\ref{fig:qualitative} and Figure~\ref{fig:qualitative_compared}. 
Figure~\ref{fig:qualitative} provides a comparison between the GT and the predictions provided by the proposed pipeline on the Replica~\cite{Replica} and NYU-Depth-v2~\cite{NYUV2} dataset.
In Figure~\ref{fig:qualitative_compared}, we compare the GT images provided by the KITTI~\cite{Kitti} dataset with the predictions from our network. 
It can be noticed that the network is able to recover even small details found in the scene.
Furthermore, it is also able to perform equally good on different depth ranges and on different image types (synthetic, real scenario). 
On the negative side, it can be seen that the network creates artifacts based on reflections or other complex lighting conditions.

Additionally, Figure~\ref{fig:qualitative_compared} provides a qualitative comparison between the state-of-the-art network AdaBins~\cite{AdaBins2020} and our predictions. 
Our method tends to perform better on thin structures while AdaBins~\cite{AdaBins2020} has a tendency to oversmooth the surrounding depth as visible at the pole on the left side in the third example. Otherwise, AdaBins~\cite{AdaBins2020} performs much better on the borders of closer objects such as the car in the second example, it also handles better shiny surfaces and specular reflections. 
Overall, the results are comparable, even though we are using a much simpler architecture.
Additional qualitative comparison can be found in the supplementary material.

\begin{figure}[tb]
    \centering
    \scriptsize
    \setlength{\tabcolsep}{1pt}
	\renewcommand{\arraystretch}{0.8}
	\newcommand{\sz}{0.5}
	\newcommand{\sh}{1.38cm}
    \begin{tabular}{cc}
        RGB & GT \\
        \includegraphics[width=\sz\linewidth, height=\sh]{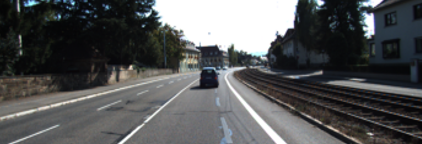} &
        \includegraphics[width=\sz\linewidth, height=\sh]{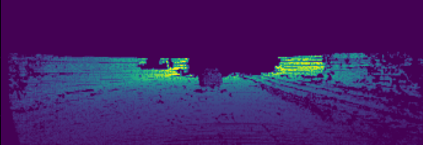}\\
        Prediction &  Masked Prediction \\
        \includegraphics[width=\sz\linewidth, height=\sh]{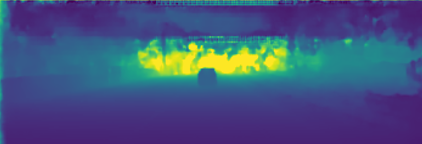} &
        \includegraphics[width=\sz\linewidth, height=\sh]{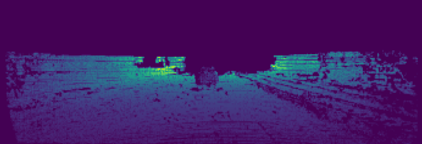}\\     
        Error 1 & Error 2 \\
        \includegraphics[width=\sz\linewidth, height=\sh]{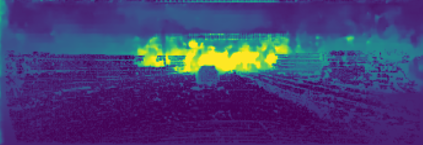} &
        \includegraphics[width=\sz\linewidth, height=\sh]{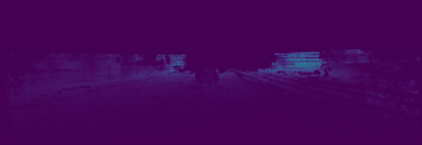}\\[-5pt]
    \end{tabular}
    \vspace{-4pt}
    \caption{\textbf{Qualitative results of the prediction, masked prediction and error images for the KITTI \cite{Kitti} dataset.} The figure provides a qualitative explanation for the cropped predictions provided in the paper and further shows the masking method applied through the training and the evaluation process. Color scale: 0 (purple) to 80 meters (yellow).}

    \vspace*{2pt}
    \label{fig:error}
\end{figure}

All the qualitative results provided from the KITTI~\cite{Kitti} dataset show cropped results. 
This is due to the nature of the dataset, as the sensors used to film the depth sequences provides a depth range from 0.01 to 80 meters -- these values were used to clip the depth predictions by the sensor by the authors.
Figure~\ref{fig:error} details this in a qualitative manner.
In the first line we provide the color image and the GT provided by the dataset; as can be seen, the upper part of this prediction is set to zero since those values are further away than 80 meters.
In the second line, we provide the actual prediction from the \textit{DepNet} after training -- the prediction is a dense depth map with hallucinations on the upper part of the image due to the absence of depth values for this part of the entire dataset.
In the same line, masked prediction is the prediction of \textit{DepNet} after applying a function to ignore the values that the original dataset does not provide (\ie, masking them out).
The last line of the figure provides the error image between the GT and the prediction or the masked prediction, respectively.
As can be seen, for Error 1 which is the GT vs the prediction without alteration, the error values are accurate for the depth information provided by the network, but the error introduced by the depth pixels of the background provides a final error value of 17 meters. 
Instead, in Error 2 where we masked out the values that the actual dataset does not provide, it can be seen that the prediction is quite accurate with an error of 2.91 meters in the entire scene.

Masking invalid pixels in the predictions is a common practice in monocular depth prediction, e.g.~\cite{Eigen2014,Laina2016,Godard2018}.

\subsection{Ablation study}
We further asses the individual building blocks of our pipeline in an ablation study that was performed on all three datasets: KITTI~\cite{Kitti}, NYU-Depth-v2~\cite{NYUV2} and Replica~\cite{Replica}.

Table~\ref{tab:ablation_study} shows the performance scores for three variants of our pipeline. 
For the first variant (``DepNet''), we trained the monocular depth network (\textit{DepNet} only with a supervised loss as the simplest baseline.
In a second variant (``DepNet+SynNet''), we added the view transformation and synthesis network \textit{SynNet} with the image loss in the second view.
The third variant (``Ours'') is our full proposed pipeline, which additionally adds the depth estimation based on the generated novel view output of \textit{SynNet}.

All the architectures were trained in a supervised manner -- we used the additional information for the additional loss functions used throughout the pipeline.
Since the synthesis network has a big impact on the architecture loss, we set its loss weight to $0.5$ such that the main focus stayed on the monocular network.
By adding the synthesis step to the monocular architecture, we can already see a significant improvement on the error metrics, proving that the additional geometrical consistency is valuable for the training. 
In this case, after obtaining the new RGB view, it is passed through the monocular architecture to predict the depth of the second view which is compared with the GT from the second view. 
This leaves us in the end with a complex training pipeline, that, after training, can be used without need of providing any other information than the ground truth depth for evaluation purposes.

The values in Table~\ref{tab:ablation_study} demonstrate that for the KITTI~\cite{Kitti} dataset, the network shows a considerable improvement on all the error values compared to the baseline architecture trained without the proposed losses. The worse performance may be related to dynamic objects and the clipping of large depth values.
Moreover, the KITTI exhibits a larger depth range and error residuals that can affect the method performance.%
The same conclusion can be drawn on the other datasets.
Another conclusion %
is that using the overall loss function makes the network more sensitive to the smaller residuals achieving better delta results, even if it is not capable to outperform the state of the arts on all metrics.

Generally, one can observe that our method has more impact on the RMSE$_\text{log}$ score which emphasizes small residual errors, in contrast to RMSE and Sq.Rel. which emphasize large residuals.
This behavior can be expected since small depth residuals lead to smaller rendering errors in a second view and can be better corrected with a differentiable rendering loss, because large depth errors lead to highly non-local changes in a second view which are hard to be picked up by a network with a local receptive field.

Qualitative results of these experiments are provided in the supplementary material.

\begin{table}[tb]
	\vspace*{4pt}
  \centering
  \scriptsize
  \setlength{\tabcolsep}{1.8pt}
  \renewcommand{\arraystretch}{1.1}
  \begin{tabular}{llccccccc}
    \toprule
    & Model & REL$\downarrow$ & RMSE$\downarrow$ & RMSE$_\text{log}\!\!\downarrow$ & Sq.Rel.$\downarrow$ & $\delta_1\!\uparrow$ & $\delta_2\!\uparrow$ & $\delta_3\!\uparrow$\\
    \midrule
    \multirow{3}{*}{\rotatebox{90}{Replica}}
    & DepNet          & 0.905 & 2.071 &   -   & 1.765 & 0.421 & 0.708 & 0.802\\
    & DepNet + SynNet & 0.403 & \bf 0.775 & 0.423 & \bf 0.276 & 0.658 & 0.796 & 0.832\\
    & \textbf{Ours}   & \bf 0.398 & 0.940 & \bf 0.418 & 0.406 & \bf 0.795 & \bf 0.836 & \bf 0.901\\[0.3pt] 
    \hline \noalign{\vskip 2pt}
    \multirow{3}{*}{\rotatebox{90}{NYU-V2}}
    & DepNet          & 0.132 & 0.571 & 0.147 & 0.915 & 0.815 & 0.839 & 0.854 \\
    & DepNet + SynNet & 0.112 & 0.411 & 0.113 & 0.731 & 0.912 & 0.971 & 0.985 \\
    & \textbf{Ours}   & \bf 0.058 & \bf 0.331 & \bf 0.055 & \bf 0.511 & \bf 0.989 & \bf 0.995 & \bf 0.997\\[0.3pt] 
    
    \hline \noalign{\vskip 2pt}
    \multirow{3}{*}{\rotatebox{90}{KITTI}}
    & DepNet          & 0.057 & 3.023 & 0.104 & 0.441 & 0.936 & 0.975 & 0.991 \\
    & DepNet + SynNet & 0.047 & 3.518 & 0.127 & 0.521 & 0.953 & 0.984 & 0.994 \\
    & \textbf{Ours}   & \bf 0.031 & \bf 2.702 & \bf 0.089 & \bf 0.292 & \bf 0.963 & \bf 0.989 & \bf 0.997 \\
    \bottomrule
  \end{tabular}
  \vspace{-4pt}
  \caption{\textbf{Ablation study.} Comparison between the different variants of our approach on the Replica~\cite{Replica} dataset, NYU-Depth-v2~\cite{NYUV2} dataset and on the KITTI~\cite{Kitti} dataset. Each of our proposed building blocks leads to significant improvements of the baseline across all datasets.}
  \label{tab:ablation_study}
  \vspace*{2pt}
\end{table}

\begin{figure*}[!htb]
\vspace*{-5pt}
    \centering
    \footnotesize
    \setlength{\tabcolsep}{1pt}
	\renewcommand{\arraystretch}{0.8}
	\newcommand{\sz}{0.33}
    \begin{tabular}{ccc}
        RGB 1 & \textit{DepNet} & Depth GT (for \textit{DepNet} supervision) \\
        \includegraphics[width=\sz\linewidth]{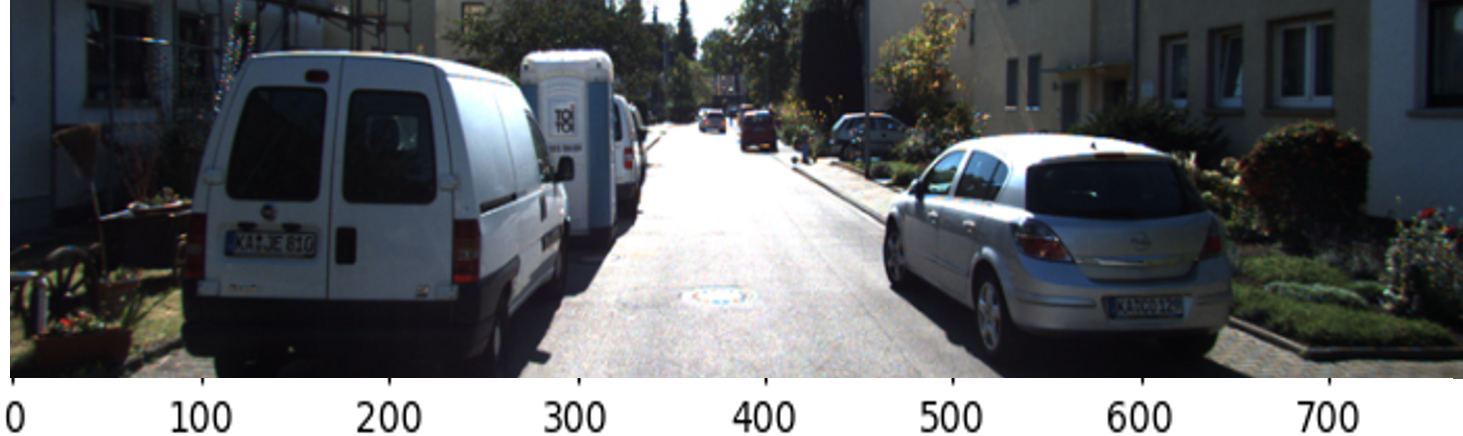} &
        \includegraphics[width=\sz\linewidth]{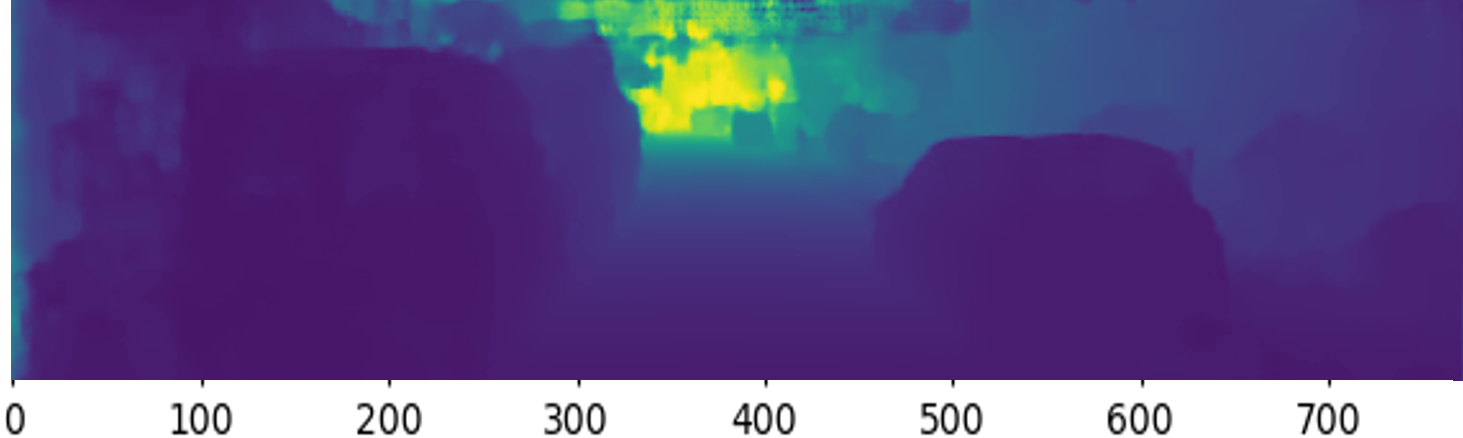} &
        \includegraphics[width=\sz\linewidth]{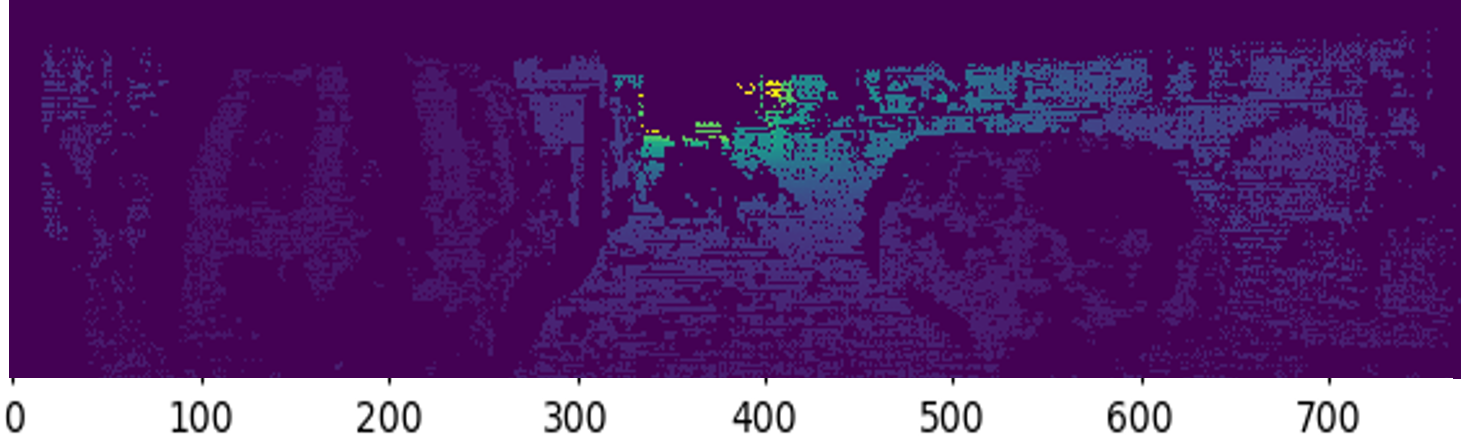}\\   
        View Transformation & \textit{SynNet} & RGB 2 (for \textit{SynNet} supervision)\\
        \includegraphics[width=\sz\linewidth]{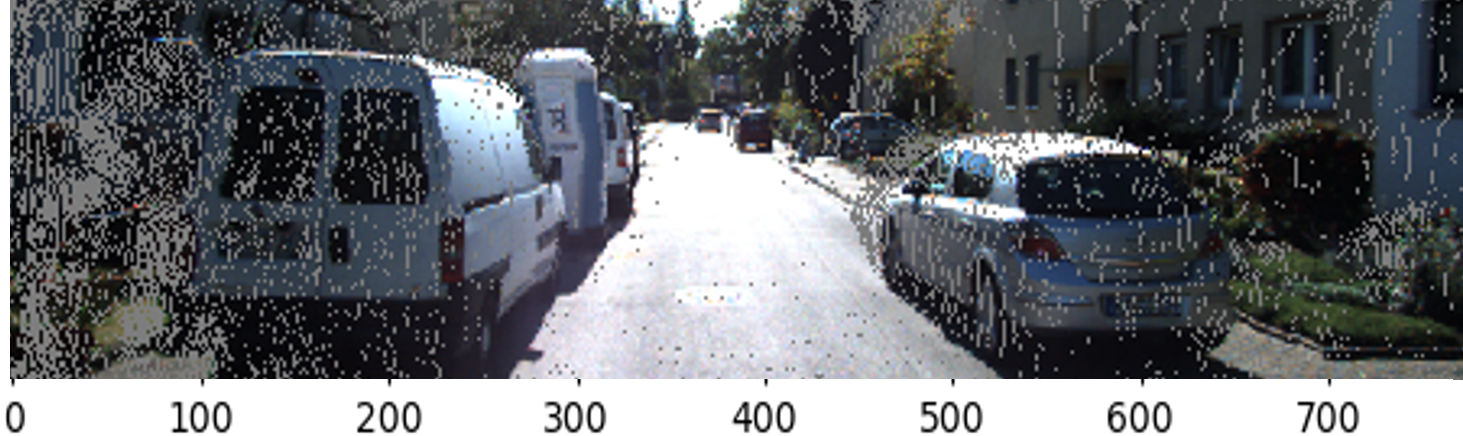} &
        \includegraphics[width=\sz\linewidth]{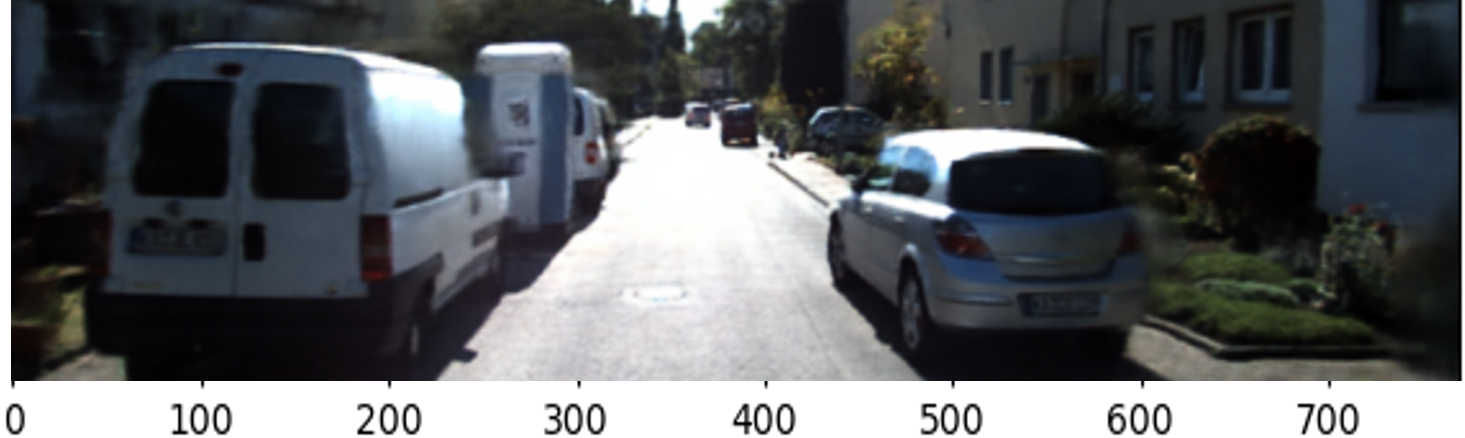} &
        \includegraphics[width=\sz\linewidth]{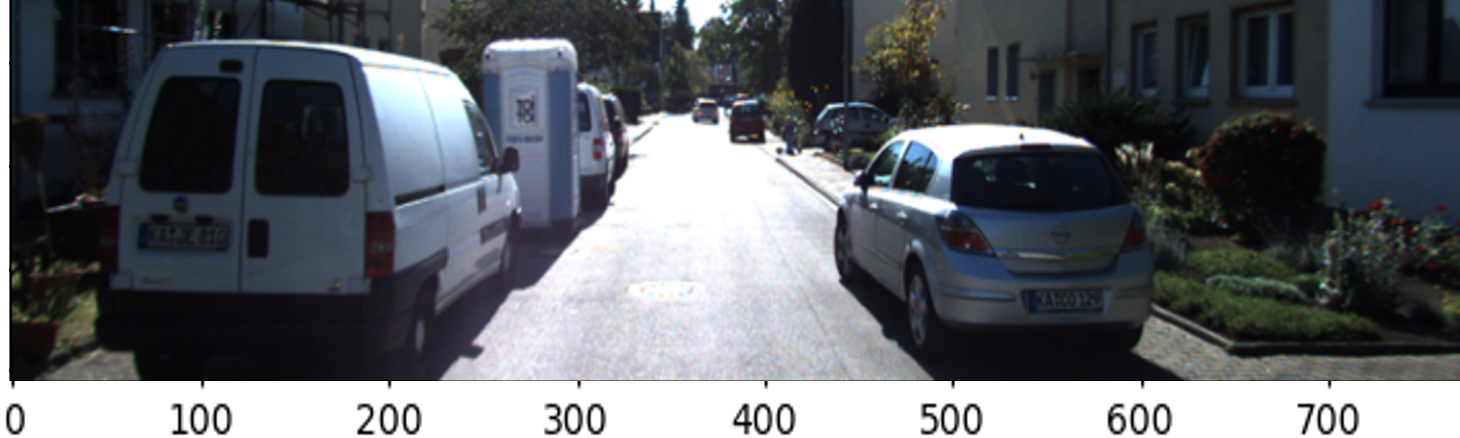}\\[-5pt]        
    \end{tabular} 
        \vspace{-5pt}
    \caption{\textbf{Qualitative predictions from each of the steps of the proposed pipeline on the KITTI \cite{Kitti} dataset}. RGB 1 and RGB 2 are consecutive views or the corresponding stereo image. Color scale: 0 (purple) to 80 meters (yellow).}
    \label{fig:Pipeline0}
\end{figure*}

\begin{figure}[!htb]
    \vspace*{-18pt}
    \centering
    \footnotesize
    \setlength{\tabcolsep}{1pt}
	\renewcommand{\arraystretch}{0.8}
	\newcommand{\sz}{0.48}
    \begin{tabular}{ccc}
        \rotatebox{90}{\hspace{5pt} RGB} &
        \includegraphics[width=\sz\linewidth]{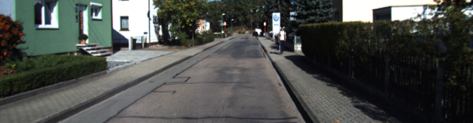} &
        \includegraphics[width=\sz\linewidth]{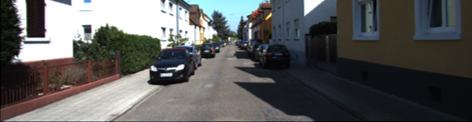} \\
        \rotatebox{90}{\hspace{7pt} GT} &
        \includegraphics[width=\sz\linewidth]{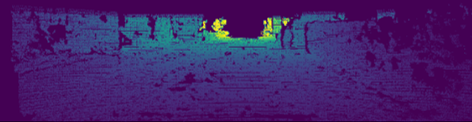} &
        \includegraphics[width=\sz\linewidth]{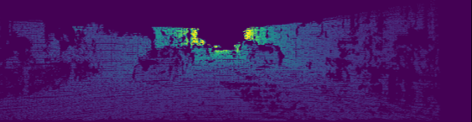}\\
        \rotatebox{90}{\hspace{2pt} \textit{DepNet}} &
        \includegraphics[width=\sz\linewidth]{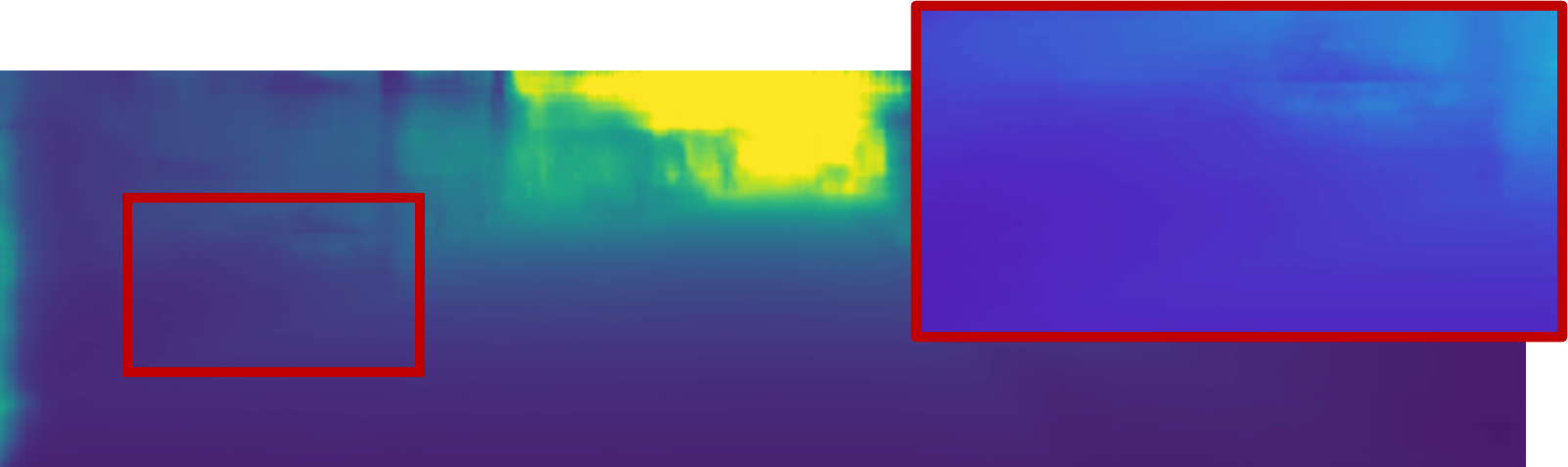} &
        \includegraphics[width=\sz\linewidth]{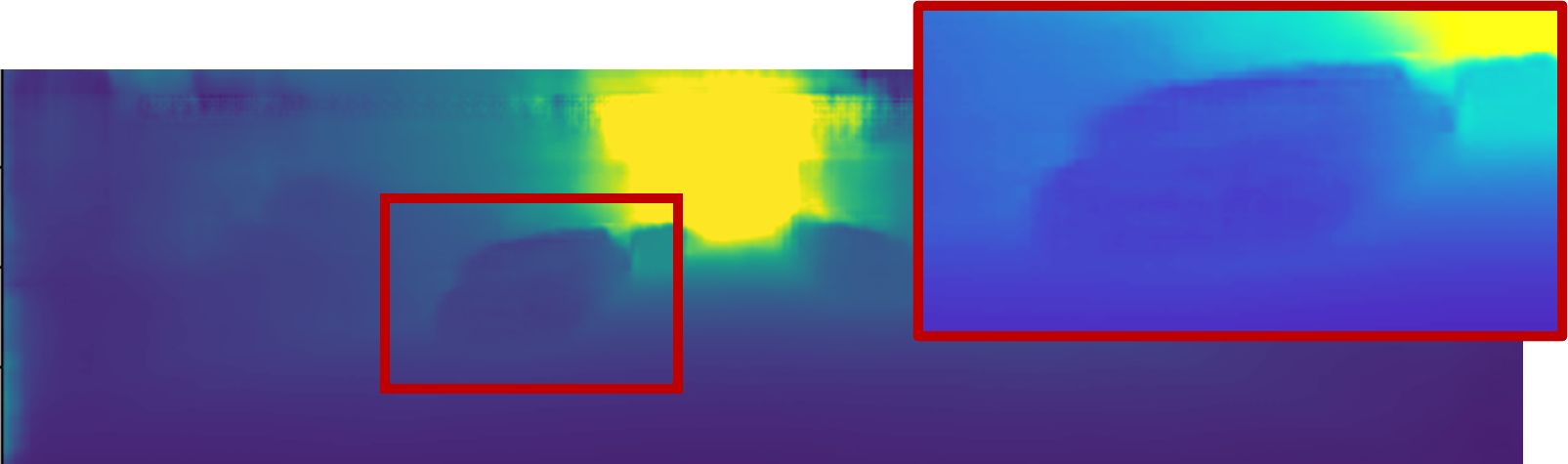}\\
        \rotatebox{90}{\hspace{4pt} Ours} &
        \includegraphics[width=\sz\linewidth]{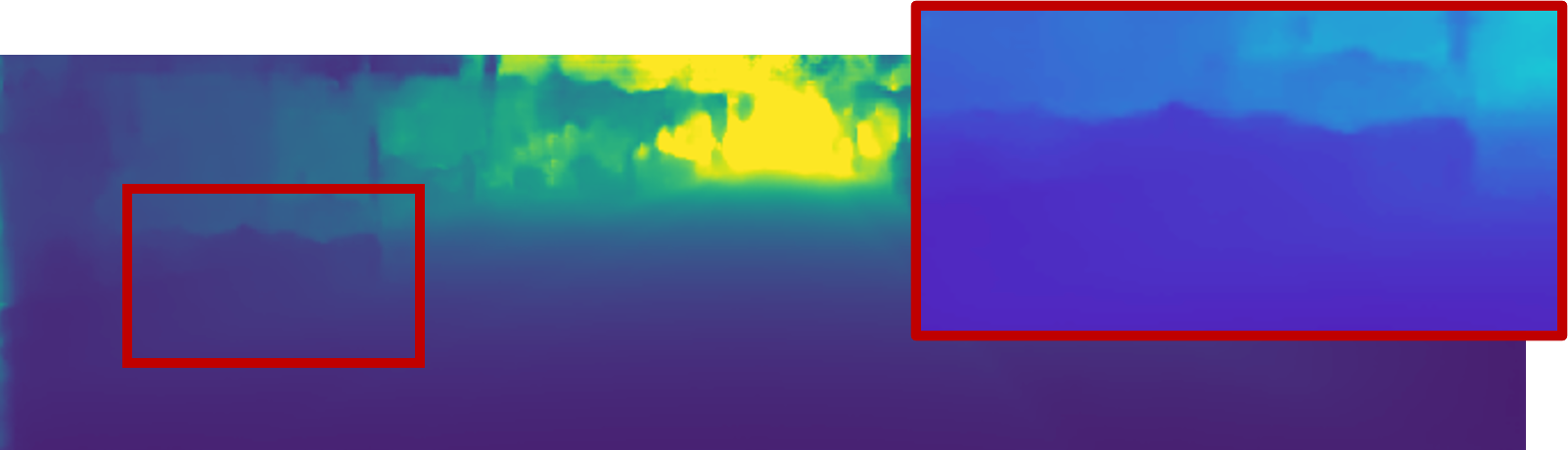} & 
        \includegraphics[width=\sz\linewidth]{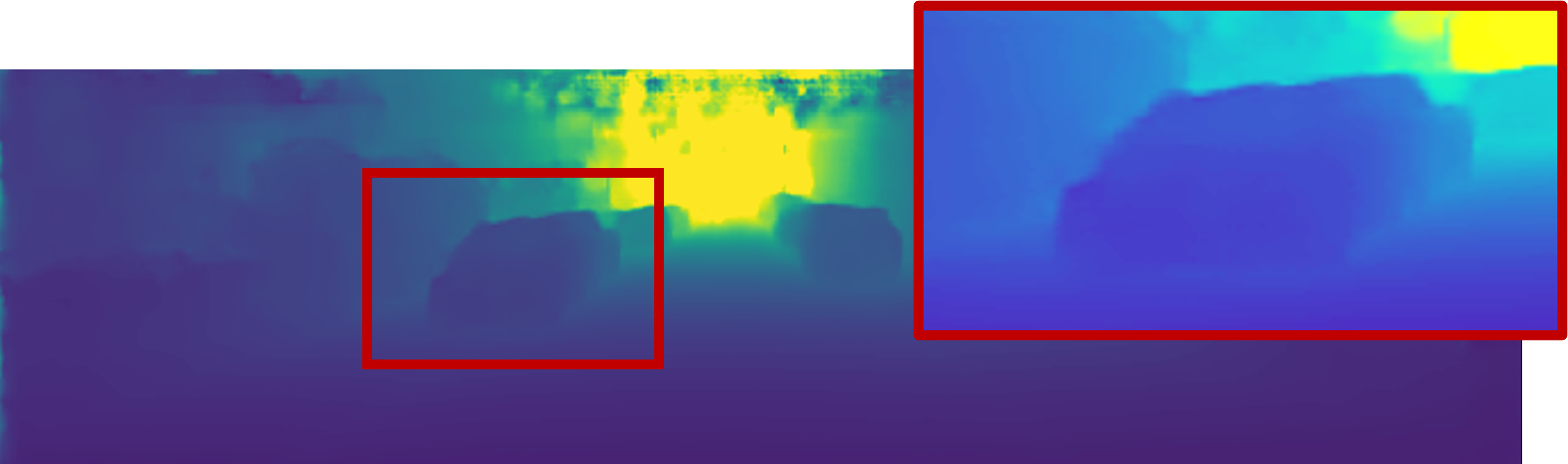}\\[-5pt]
    \end{tabular}
        \vspace{-4pt}
    \caption{\textbf{Qualitative comparison between the vanilla U-Net~\cite{Ronneberger2015} (\textit{DepNet}) and our \textbf{NVS-MonoDepth} variants on the KITTI \cite{Kitti} dataset.} As shown in the close-ups, our method improves the prediction of \textit{DepNet} with better sharpness and finer details. The contrast in the close-ups was adjusted for better visualization.}
    \label{fig:ablation_qualitative}
    \vspace*{\figbottomspace}
\end{figure}

In Figure~\ref{fig:ablation_qualitative} we provide qualitative results between \textit{DepNet} (standard U-Net~\cite{Ronneberger2015} using only the first loss $\mathcal{L}_1$ during training) and the proposed method, \textbf{NVS-MonoDepth}, showing clear improvements on the contours of the predictions by adding the additional steps to the training.

Figure~\ref{fig:Pipeline0} provides qualitative results for each step of the proposed pipeline. As can be seen, the network is able to provide in each of these steps an accurate prediction.
Since it is difficult to see the warping differences, we provided a measuring line below each image to make it easier to see the differences between the two viewpoints.

Related to the training loss function outputs of the different experiments we show a figure with those outputs in the supplementary material.
Based on the loss functions, we can conclude that the monocular baseline is mostly the one that stays the most stable through the training process. By adding the synthesis step and the additional monocular architecture, we generate additional noise on the loss curve but, at the same time, the networks is able to outperform the simple monocular architecture.
Also, this generated noise basically shows that by adding additional steps to the pipeline the network gets more difficult to train on one side, but on the other, the predictions improve achieving a much smaller error. 
More results, additional figures and explanations is provided in the supplementary material.

\section{Conclusions}
We presented a novel training method that makes use of additional loss functions to improve monocular depth estimation. 
The key idea is to use consistency constraints from other views via novel view synthesis as an additional supervisory signal for training in a multi-view setting, while testing only requires monocular input.
This novel training procedure leads to substantial performance improvements compared to traditional supervised training.
Our method achieves comparable results with the state of art for the KITTI~\cite{Kitti} dataset. %
On the NYU-Depth-v2~\cite{NYUV2} dataset our method outperforms all state-of-the-art methods across all metrics.
The ablation of the individual pipeline parts of our building blocks demonstrates significant improvements supporting our design choices empirically across three datasets.
Overall, we demonstrated the effectiveness of the proposed training method to improve the training of a monocular depth estimation network.
As future work and to foster further research in this field, we believe that the proposed training principles can be combined with other monocular depth prediction approaches as also losses to further push their limits.

\noindent
\begin{minipage}{\columnwidth}
	\vspace{3pt}
	\footnotesize
	\noindent
	\textbf{Acknowledgments.}~
	This work has been supported by the Spanish Government PID2019-104818RB-I00 Grant, co-funded by EU Structural Funds.
	Further funding was provided by a Fellowship from the Swiss Data Science Center, Innosuisse funding (Grant No.~34475.1 IP-ICT), and a research grant by FIFA.
\end{minipage}

{
	\small
	\bibliographystyle{ieee_fullname}

\input{main.bbl}
}

\clearpage
\appendix
\begin{appendices}
	\input{supp_main.tex}
\end{appendices}

\end{document}

%% file: supp_main.tex
\bibliographystyle{unsrt}

\renewcommand\thesubfigure{\alph{subfigure})}
\newcommand{\figbottomspacesupp}{-4pt}

\threedvfinalcopy %

\section{Architectures}

\begin{figure*}[!htb]
    \centering
    \vspace*{-1.3cm}
    \includegraphics[width=1.0\textwidth, height=11cm]{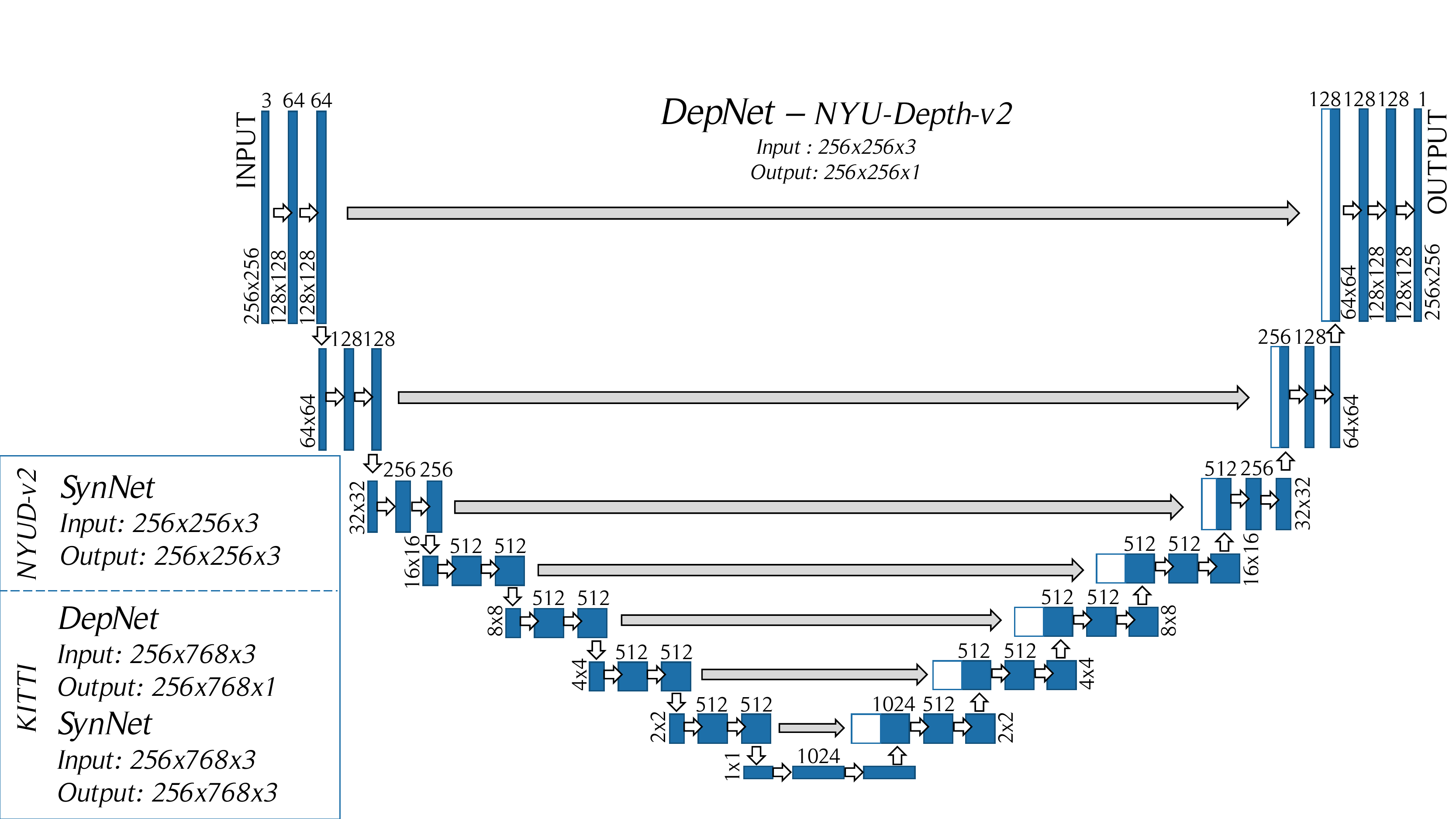}
    \vspace*{-0.5cm}
    \caption{U-Net~\cite{Ronneberger2015} architecture used for the proposed method. The input and output dimensions for each dataset and architecture are provided in the box. The detailed architecture shown in the figure is the \textit{DepNet} network trained with the NYU-Depth-v2~\cite{NYUV2} dataset.}
    \label{fig:detailed_arch}
    \vspace*{-0.1cm}
\end{figure*}

In Figure~\ref{fig:detailed_arch}, we provide a detailed overview of the U-Net~\cite{Ronneberger2015} architectures used for the \textit{DepNet} and \textit{SynNet} modules of our method. 
The figure shows each layer of the \textit{DepNet} network trained on the NYU-Depth-v2~\cite{NYUV2} dataset, providing the input and output sizes of the network, as well as the intermediate sizes. 
As additional information, we provide the input and output sizes for the \textit{SynNet} trained on the NYU-Depth-v2~\cite{NYUV2} and KITTI~\cite{Kitti} datasets, respectively.
The number of layers stays the same in all the cases.

Additionally, in Table~\ref{tab:Datasets}, we provide an extended analysis of the most popular datasets for depth estimation.
Our main criteria when choosing a dataset to test the proposed method on were: scale, stereo images or video sequences, accurate camera poses, and, lastly, if the dataset is outdoor or indoor. 
After this analysis, we settled on the KITTI~\cite{Kitti} and NYU-Depth-v2~\cite{NYUV2} datasets to be able to compare with the state of the art, and, additionally, the Replica~\cite{Replica} dataset to try the method in a synthetic scenario.

\section{Qualitative comparison with state-of-the-art methods on the KITTI dataset}

Figures~\ref{fig:qualitative_compared1} and~\ref{fig:qualitative_compared2} provide additional qualitative results compared to the AdaBins~\cite{AdaBins2020}, BTS~\cite{Lee2019}, LapDepth~\cite{Song2021} and VNL~\cite{Yin2019} methods, proving that \textbf{NVS-MonoDepth} is able to predict comparable results while using a much lighter architecture compared to the state-of-the-art methods.

As can be noted in the provided qualitative results, our method is able to compete against the state-of-the-art methods predicting depth accurately, even at a small detail level.
For example, for bushes on the roadside, state-of-the-art methods typically predict a smoother surface, while our results are more undulating.
We believe this results from the additional novel view supervision signal, which effectively guides the depth estimation network to produce more details. 
On the contrary, VNL~\cite{Yin2019} predicts the most blurred results among several methods, especially for objects in the distance, whose boundaries are more difficult to distinguish from the background.
At the same time, our method has an accurate prediction in details but at the same time it also introduces more obvious artifacts.
At the top of the image, since there is no depth supervision, it is inevitable that there might be some artifacts in this area during inference. AdaBins~\cite{AdaBins2020}'s predictions in this area seem to be the most plausible. 
Our method tends to give noisy results, while LapDepth~\cite{Song2021} prefers a smaller depth.

\section{Training loss curves}
\begin{figure}[!htb]
    \centering
    \includegraphics[width=0.48\textwidth, height=4.4cm]{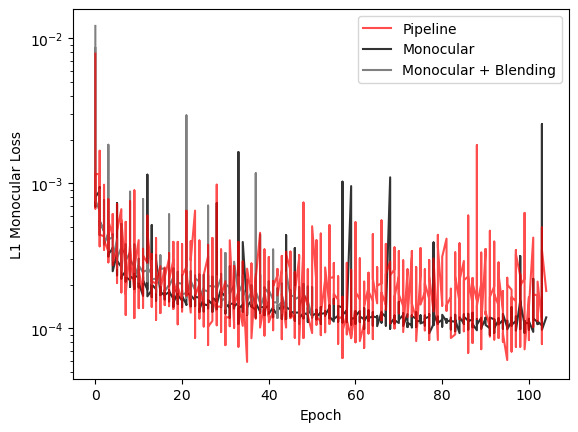}
    \includegraphics[width=0.48\textwidth, height=4.4cm]{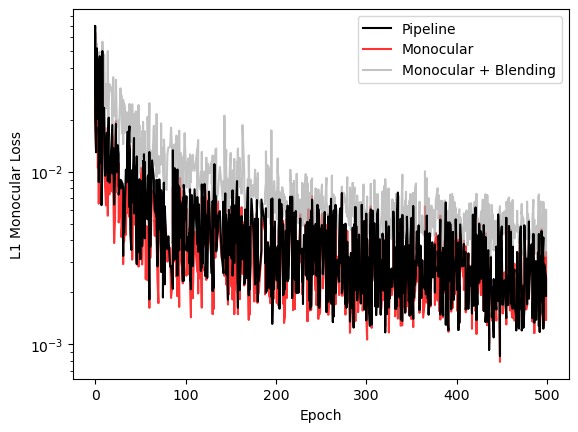}
    \vspace*{-0.3cm}
    \caption{Loss curves from the different experiments performed on the KITTI~\cite{Kitti} dataset (top) and the Replica~\cite{Replica} dataset (bottom).}
    \label{fig:Loss_curves}
\end{figure}

Figure~\ref{fig:Loss_curves} shows the loss curves of the training performed on the Replica~\cite{Replica} and KITTI~\cite{Kitti} datasets.
As already mentioned in the main paper, based on the loss functions, we can conclude that the monocular baseline is mostly the one that stays the most stable through the training process.
By adding the synthesis step and the additional monocular architecture, we generate some noise on the loss curve but, at the same time, the network is able to outperform the simple monocular architecture.
Also, this generated noise shows that, by adding additional steps to the pipeline, the network training gets more difficult, but the predictions improve and achieve a much smaller error. 

\section{Additional qualitative results}

Figure~\ref{fig:sup_NYU} provides additional qualitative results of the architecture trained with the proposed method on the NYU-Depth-v2~\cite{NYUV2} dataset -- we show the performance of the \textbf{NVS-MonoDepth} method compared to the ground truth (GT) images, proving its effectiveness on indoor datasets. 
We additionally provide an error image of the comparison between the GT (with filled out gaps for better comparison) and the prediction of our method, and finally a quantitative mean error of the prediction. 
In both depth and error images, purple represents 0 meter and yellow 10 meter depth. 
As can be noted, the network is able to predict general details in the scene, providing accurate depth prediction, but it struggles with smaller geometries such as the plant leafs.
Another weakness of our network is the handling of low-textured surfaces: in the fourth example, dealing with a lot of similar color shades and also light incidence makes it difficult for the network to predict accurately, providing in the end a worse depth prediction compared to the other examples.

In Figures~\ref{fig:Pipeline_1} and~\ref{fig:Pipeline_2}, we provide qualitative results on the KITTI~\cite{Kitti} dataset for each step of the proposed method, showing the outputs compared to their corresponding GT.
As already stated before, our network handles the depth predictions at the positions where we lack of GT, creating artifacts in most of the prediction backgrounds.
At the same time, the depth prediction in the range from 0 to 80 meters is accurate to the point of providing a good view transformation that is able to serve as input for the \textit{SynNet} to predict new view points.
Related to the new view points, as can be seen, our network is able to predict accurate RGB images.
The main view of the images tends to be sharp and synthesized accurately when compared to the GT; on the sides, where no depth information is provided, the network creates blurry outputs, as can be seen on the tree lines, humans moving or at the street signs in the corners of the image.
We recommend viewing these figures on a monitor zoomed in. 

In Figure~\ref{fig:ablation_qualitative_supp}, we provide a qualitative comparison between the vanilla U-Net~\cite{Ronneberger2015} denoted as \textit{DepNet} and our \textbf{NVS-MonoDepth} method on the KITTI~\cite{Kitti} dataset. 
For better comparison, we provide close-ups on each of the predictions so that the main differences between the two can be better analyzed.
As can be seen, our method adds additional geometrical consistency to the prediction providing more details to the objects.
Also, as can be seen in the case of the street sign or the person on a bike, the prediction is much more accurate than the one provided by the \textit{DepNet} architecture.
In the third example containing a black car, %
even if both networks struggle to predict the car accurately because of reflections, the \textbf{NVS-MonoDepth} method still provides a smoother prediction than the \textit{DepNet} architecture which produces random artifacts on the surface.
\clearpage

\begin{table*}[!htb]
  \centering
  \setlength{\tabcolsep}{18pt}
  \begin{tabular}{lcccccc}
    \toprule
    \multirow{2}*{Dataset} & \multicolumn{2}{c}{Scale} & \multirow{2}*{Stereo / Sequence} & \multirow{2}*{Pose} & \multirow{2}*{Outdoor} \\
    \cmidrule(lr){2-3}
    & Frames & Scenes & & & \\
    \midrule
    SceneNet RGB-D~\cite{SCENENET} & 5M & 57 & \cmark & \cmark &  \xmark \\
    Matterport 3D~\cite{Matterport3D} & 200k & 90 & \cmark & \cmark &  \xmark \\
    Stanford 2D-3D~\cite{STANFORD2D3D} & 70k & 270 & \cmark & \xmark &  \xmark \\
    Microsoft RGB-D~\cite{MICROSOFTRGBD} & 1k p.sq. & 7 & \cmark & \cmark &  \xmark \\
    ViDRILO~\cite{VIDRILO} & 22k & 10 & \cmark &  \xmark &  \xmark \\
    SUNCG~\cite{SUNCG} & - & 45622 & \cmark & \cmark &  \xmark\\
    ScanNet~\cite{SCANNET} & 2.5M & 1513 & \cmark & \cmark &  \xmark\\
    SUN3D~\cite{SUN3D} & - & 415 & \cmark & \cmark & \xmark\\
    InteriorNet~\cite{InteriorNet18} & 5M & 1.7M & \cmark & \cmark &  \xmark \\
    NYU-Depth-v2~\cite{NYUV2} & 1.5k & 464 & \cmark & \xmark &  \xmark\\
    SYNTHIA~\cite{Synthia} & 200k & - & \cmark & \cmark &  \cmark\\
    KITTI~\cite{Kitti} & 1.6k & 400 & \cmark & \cmark &  \cmark \\
    ETH3D~\cite{ETH3D} & 898 & 25 & \cmark & \cmark &  both \\
    MAKE 3D~\cite{Make3D} & 534 & - & \xmark & \xmark &  both \\
    Tanks and Temples~\cite{TanksAndTemples} & 147k & 14 & \cmark & \cmark &  both \\
    Middlebury~\cite{Middlebury} & 40k & 33 & \cmark & \cmark &  \xmark \\
    UASOL~\cite{UASOL} & 48k & 33 & \cmark & \cmark &  \cmark\\
    Diode~\cite{Diode} & 28k & 30 & \xmark & \xmark &  both \\
    Replica~\cite{Replica} & - & 18 & \cmark & \cmark &  \xmark  \\
    \bottomrule
  \end{tabular}
  \caption{Most popular datasets used for depth estimation. The main focus relies on the scale of the dataset, if the dataset is a stereo dataset or the images are sequences, if camera poses are provided and if the dataset is Outdoor or Indoor.}
   \label{tab:Datasets}
\end{table*}

\clearpage
\clearpage
\clearpage
\newpage
\newpage
\begin{sidewaysfigure}
    \centering
    \scriptsize
    \setlength{\tabcolsep}{1pt}
	\renewcommand{\arraystretch}{0.8}
	\newcommand{\sz}{0.26}
	\newcommand{\sh}{1.9cm}
    \vspace*{9cm}
    \hspace*{-20pt}
    \begin{tabular}{lccc}
        \rotatebox{90}{\hspace{12pt} RGB} &
        \includegraphics[width=\sz\linewidth, height=\sh]{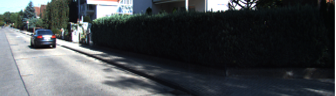}
        \includegraphics[width=\sz\linewidth, height=\sh]{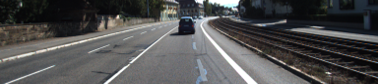}
        \includegraphics[width=\sz\linewidth, height=\sh]{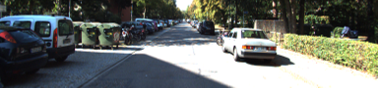}
        \includegraphics[width=\sz\linewidth, height=\sh]{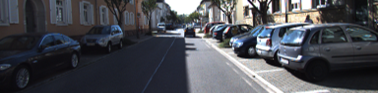}
\\
        \rotatebox{90}{\hspace{14pt} GT} &
        \includegraphics[width=\sz\linewidth, height=\sh]{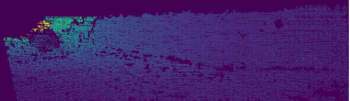}
        \includegraphics[width=\sz\linewidth, height=\sh]{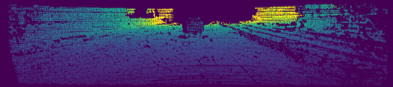}
        \includegraphics[width=\sz\linewidth, height=\sh]{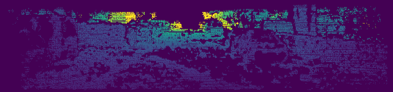}
        \includegraphics[width=\sz\linewidth, height=\sh]{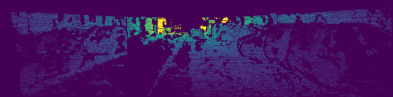}
\\
        \rotatebox{90}{\hspace{14pt} Ours} &
        \includegraphics[width=\sz\linewidth, height=\sh]{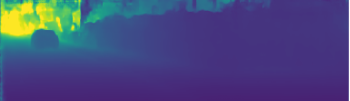}
        \includegraphics[width=\sz\linewidth, height=\sh]{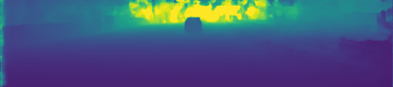}
        \includegraphics[width=\sz\linewidth, height=\sh]{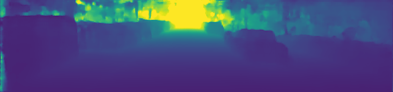}
        \includegraphics[width=\sz\linewidth, height=\sh]{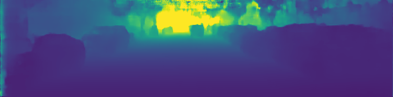}
\\
        \rotatebox{90}{\hspace{9pt} AdaBins~\cite{AdaBins2020}} &
        \includegraphics[width=\sz\linewidth, height=\sh]{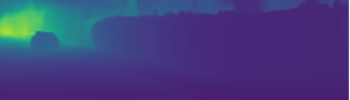}
        \includegraphics[width=\sz\linewidth, height=\sh]{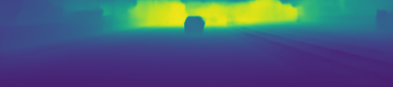}
        \includegraphics[width=\sz\linewidth, height=\sh]{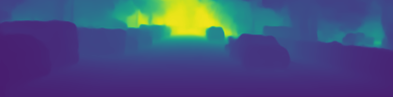}
        \includegraphics[width=\sz\linewidth, height=\sh]{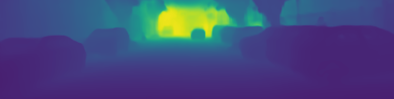}
\\
        \rotatebox{90}{\hspace{10pt} BTS~\cite{Lee2019}} &
        \includegraphics[width=\sz\linewidth, height=\sh]{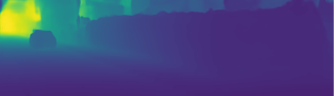}
        \includegraphics[width=\sz\linewidth, height=\sh]{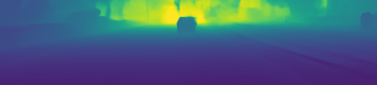}
        \includegraphics[width=\sz\linewidth, height=\sh]{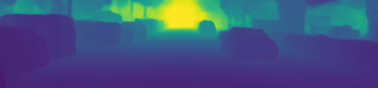}
        \includegraphics[width=\sz\linewidth, height=\sh]{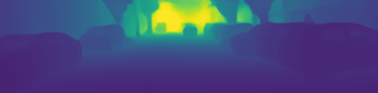}
\\
        \rotatebox{90}{\hspace{7pt} LapDepth~\cite{Song2021}} &
        \includegraphics[width=\sz\linewidth, height=\sh]{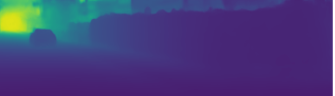}
        \includegraphics[width=\sz\linewidth, height=\sh]{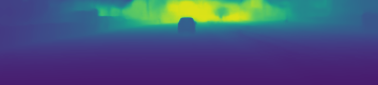}
        \includegraphics[width=\sz\linewidth, height=\sh]{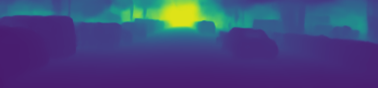}
        \includegraphics[width=\sz\linewidth, height=\sh]{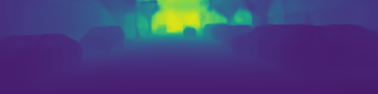}
\\
        \rotatebox{90}{\hspace{10pt} VNL~\cite{Yin2019}} &
        \includegraphics[width=\sz\linewidth, height=\sh]{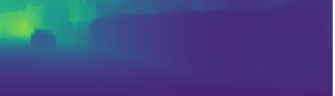}
        \includegraphics[width=\sz\linewidth, height=\sh]{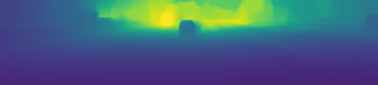}
        \includegraphics[width=\sz\linewidth, height=\sh]{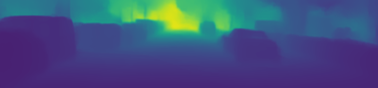}
        \includegraphics[width=\sz\linewidth, height=\sh]{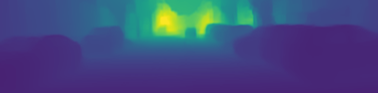}
\\[-5pt]
    \end{tabular}
    \caption{\textbf{Qualitative predictions on the KITTI \cite{Kitti} dataset.} We show the performance of \textbf{NVS-MonoDepth} (Ours) compared to AdaBins~\cite{AdaBins2020}, BTS~\cite{Lee2019}, LapDepth~\cite{Song2021} and VNL~\cite{Yin2019} leading to similar predictions, but with a much simpler architecture. The color scale goes from 0 (purple) to 80 meters (yellow). Best viewed on a monitor zoomed in.
    }
    \label{fig:qualitative_compared1}
    \vspace*{\figbottomspacesupp}
\end{sidewaysfigure}

\clearpage

\begin{sidewaysfigure}
    \centering
    \scriptsize
    \setlength{\tabcolsep}{1pt}
	\renewcommand{\arraystretch}{0.8}
	\newcommand{\sz}{0.26}
	\newcommand{\sh}{1.9cm}
    \vspace*{9cm}
    \hspace*{-20pt}
    \begin{tabular}{lccc}
        \rotatebox{90}{\hspace{12pt} RGB} &
        \includegraphics[width=\sz\linewidth, height=\sh]{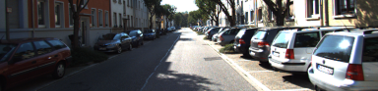}
        \includegraphics[width=\sz\linewidth, height=\sh]{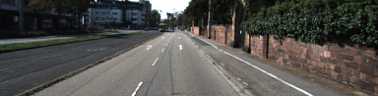}
        \includegraphics[width=\sz\linewidth, height=\sh]{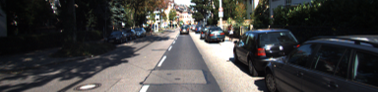}
        \includegraphics[width=\sz\linewidth, height=\sh]{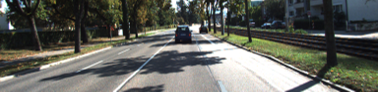}
\\
        \rotatebox{90}{\hspace{14pt} GT} &
        \includegraphics[width=\sz\linewidth, height=\sh]{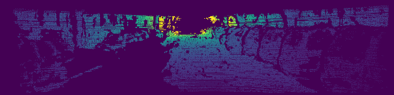}
        \includegraphics[width=\sz\linewidth, height=\sh]{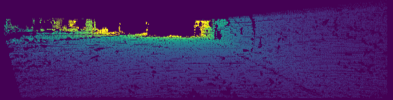}
        \includegraphics[width=\sz\linewidth, height=\sh]{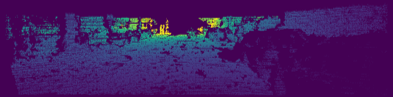}
        \includegraphics[width=\sz\linewidth, height=\sh]{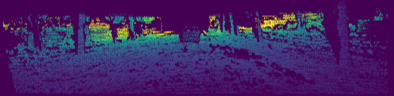}
\\
        \rotatebox{90}{\hspace{14pt} Ours} &
        \includegraphics[width=\sz\linewidth, height=\sh]{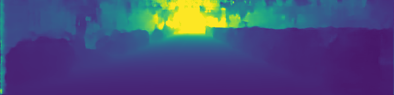}
        \includegraphics[width=\sz\linewidth, height=\sh]{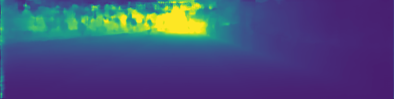}
        \includegraphics[width=\sz\linewidth, height=\sh]{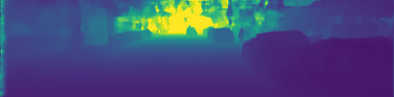}
        \includegraphics[width=\sz\linewidth, height=\sh]{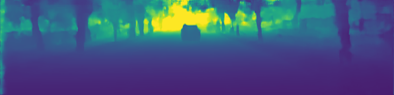}
\\
        \rotatebox{90}{\hspace{9pt} AdaBins~\cite{AdaBins2020}} &
        \includegraphics[width=\sz\linewidth, height=\sh]{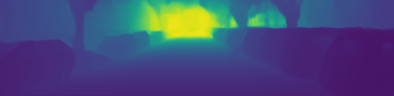}
        \includegraphics[width=\sz\linewidth, height=\sh]{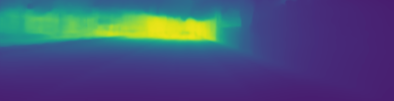}
        \includegraphics[width=\sz\linewidth, height=\sh]{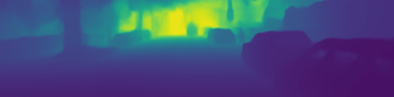}
        \includegraphics[width=\sz\linewidth, height=\sh]{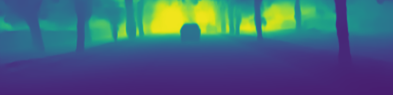}
\\
        \rotatebox{90}{\hspace{10pt} BTS~\cite{Lee2019}} &
        \includegraphics[width=\sz\linewidth, height=\sh]{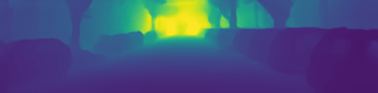}
        \includegraphics[width=\sz\linewidth, height=\sh]{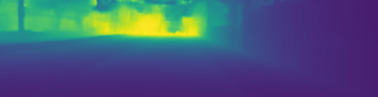}
        \includegraphics[width=\sz\linewidth, height=\sh]{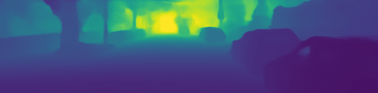}
        \includegraphics[width=\sz\linewidth, height=\sh]{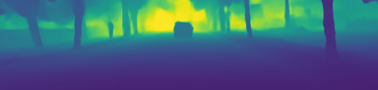}
\\
        \rotatebox{90}{\hspace{7pt} LapDepth~\cite{Song2021}} &
        \includegraphics[width=\sz\linewidth, height=\sh]{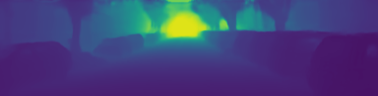}
        \includegraphics[width=\sz\linewidth, height=\sh]{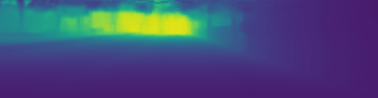}
        \includegraphics[width=\sz\linewidth, height=\sh]{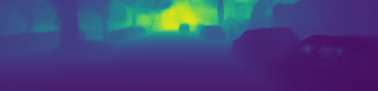}
        \includegraphics[width=\sz\linewidth, height=\sh]{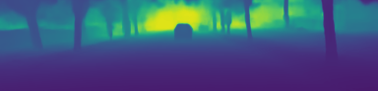}

\\
        \rotatebox{90}{\hspace{10pt} VNL~\cite{Yin2019}} &
        \includegraphics[width=\sz\linewidth, height=\sh]{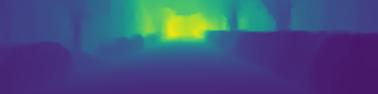}
        \includegraphics[width=\sz\linewidth, height=\sh]{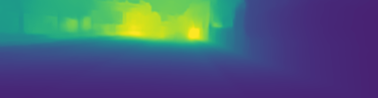}
        \includegraphics[width=\sz\linewidth, height=\sh]{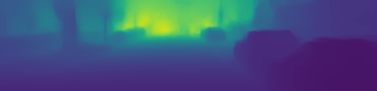}
        \includegraphics[width=\sz\linewidth, height=\sh]{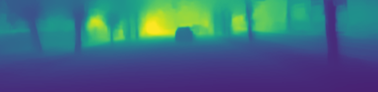}
\\[-5pt]
    \end{tabular}
    \caption{\textbf{Qualitative predictions on the KITTI \cite{Kitti} dataset.} We show the performance of \textbf{NVS-MonoDepth} (Ours) compared to AdaBins~\cite{AdaBins2020}, BTS~\cite{Lee2019}, LapDepth~\cite{Song2021} and VNL~\cite{Yin2019} leading to similar predictions, but with a much simpler architecture. The color scale goes from 0 (purple) to 80 meters (yellow). Best viewed on a monitor zoomed in.
    }
    \label{fig:qualitative_compared2}
    \vspace*{\figbottomspacesupp}
\end{sidewaysfigure}
\clearpage
\begin{figure*}[!htb]
    \centering
    \scriptsize
    \setlength{\tabcolsep}{1pt}
	\renewcommand{\arraystretch}{0.8}
	\newcommand{\sz}{0.25}
	\newcommand{\sh}{4cm}
    \begin{tabular}{ccccc}
        RGB & GT & PREDICTION & ERROR &  \\
        \includegraphics[width=\sz\linewidth, height=\sh]{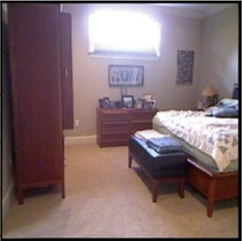} &
        \includegraphics[width=\sz\linewidth, height=\sh]{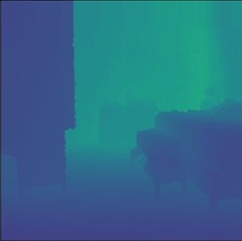} &
        \includegraphics[width=\sz\linewidth, height=\sh]{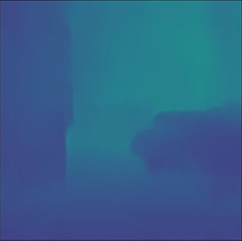} &
        \includegraphics[width=\sz\linewidth, height=\sh]{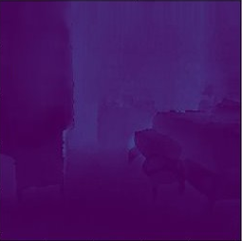} &
        \rotatebox{90}{\hspace{17.5pt} Mean Error: 0.479 meters}\\
        \includegraphics[width=\sz\linewidth, height=\sh]{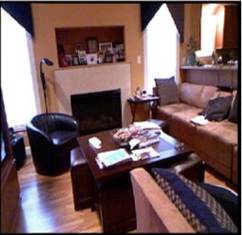} &
        \includegraphics[width=\sz\linewidth, height=\sh]{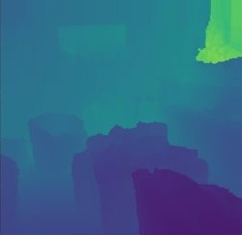} &
        \includegraphics[width=\sz\linewidth, height=\sh]{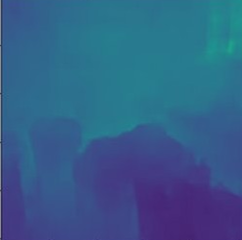} &
        \includegraphics[width=\sz\linewidth, height=\sh]{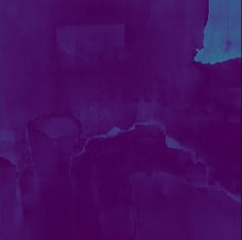} &
        \rotatebox{90}{\hspace{17.5pt} Mean Error: 0.402 meters}\\
        \includegraphics[width=\sz\linewidth, height=\sh]{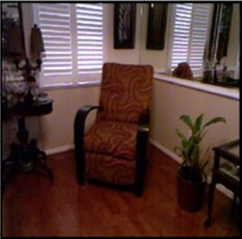} &
        \includegraphics[width=\sz\linewidth, height=\sh]{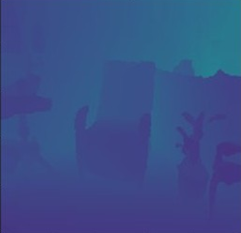} &
        \includegraphics[width=\sz\linewidth, height=\sh]{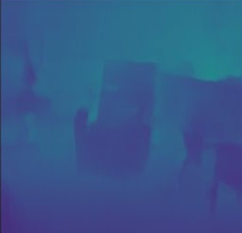} &
        \includegraphics[width=\sz\linewidth, height=\sh]{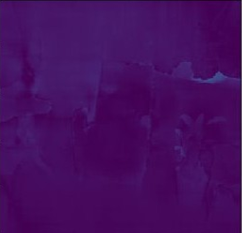} &
        \rotatebox{90}{\hspace{17.5pt} Mean Error: 0.351 meters}\\
        \includegraphics[width=\sz\linewidth, height=\sh]{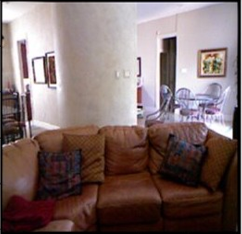} &
        \includegraphics[width=\sz\linewidth, height=\sh]{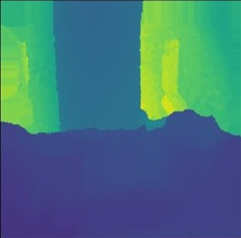} &
        \includegraphics[width=\sz\linewidth, height=\sh]{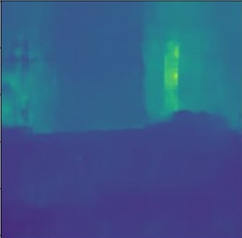} &
        \includegraphics[width=\sz\linewidth, height=\sh]{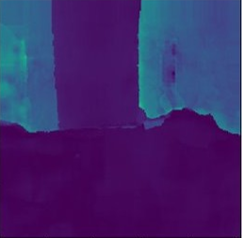} &
        \rotatebox{90}{\hspace{17.5pt} Mean Error: 1.223 meters}\\
        \includegraphics[width=\sz\linewidth, height=\sh]{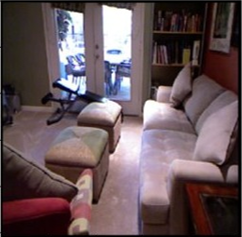} &
        \includegraphics[width=\sz\linewidth, height=\sh]{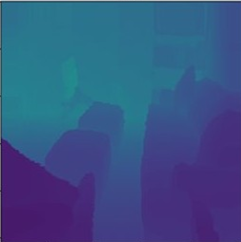} &
        \includegraphics[width=\sz\linewidth, height=\sh]{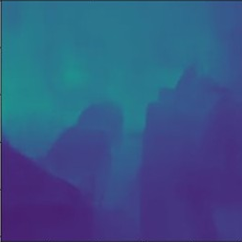} &
        \includegraphics[width=\sz\linewidth, height=\sh]{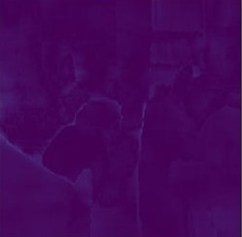} &
        \rotatebox{90}{\hspace{17.5pt} Mean Error: 0.172 meters}\\
\end{tabular}
    \caption{Qualitative predictions from the proposed pipeline on the NYU-Depth-v2 \cite{NYUV2} datasets. The color scale goes from 0 (purple) to 10 (yellow) meters.}
    \label{fig:sup_NYU}
\end{figure*}

\clearpage
\begin{figure*}[!htb]
    \centering
    \footnotesize
    \setlength{\tabcolsep}{1pt}
	\renewcommand{\arraystretch}{0.8}
	\newcommand{\sz}{0.33}
	\newcommand{\tabspace}{6pt}
	\vspace*{-14pt}
    \begin{tabular}{ccc}
        RGB 1 & \textit{DepNet} & Depth GT (for \textit{DepNet} supervision)\\
        \includegraphics[width=\sz\linewidth]{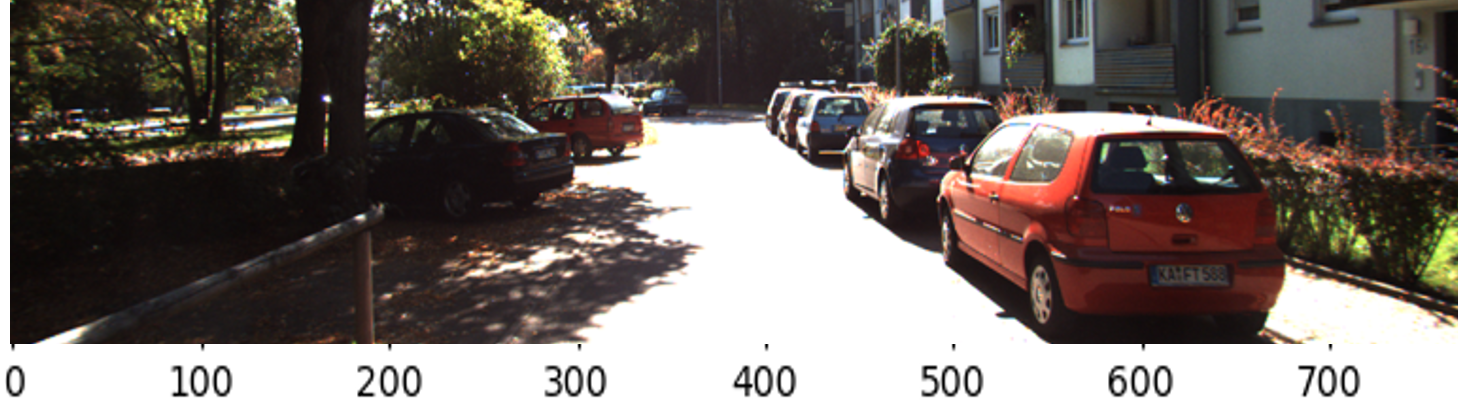} &
        \includegraphics[width=\sz\linewidth]{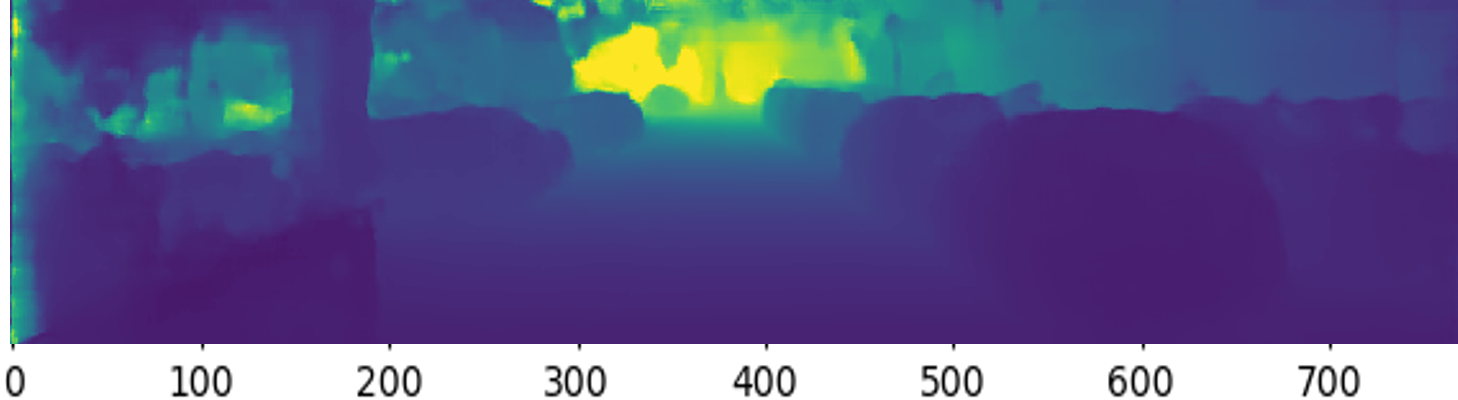} &
        \includegraphics[width=\sz\linewidth]{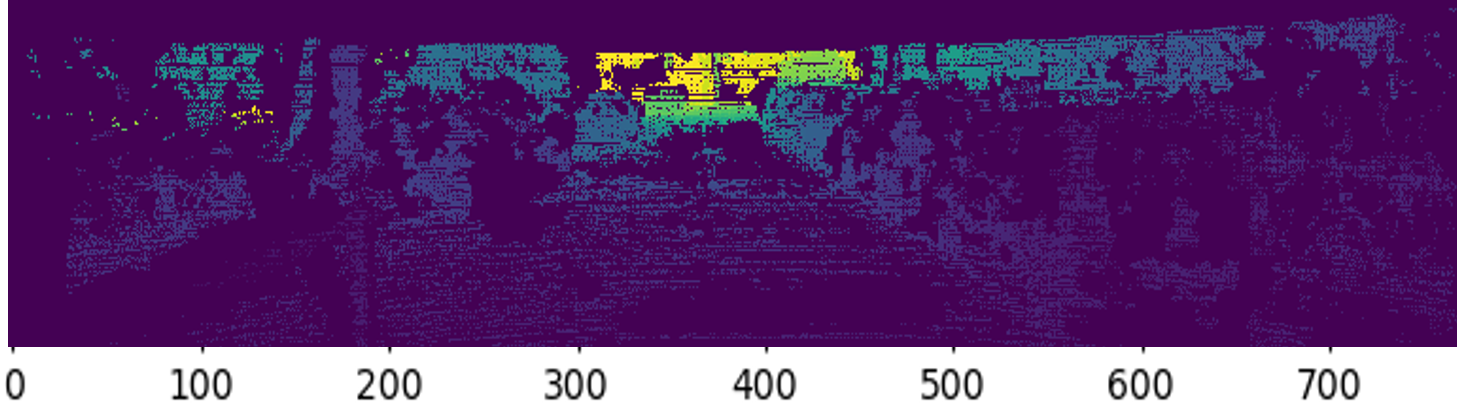}\\   
        View Transformation & \textit{SynNet} & RGB 2 (for \textit{SynNet} supervision)\\
        \includegraphics[width=\sz\linewidth]{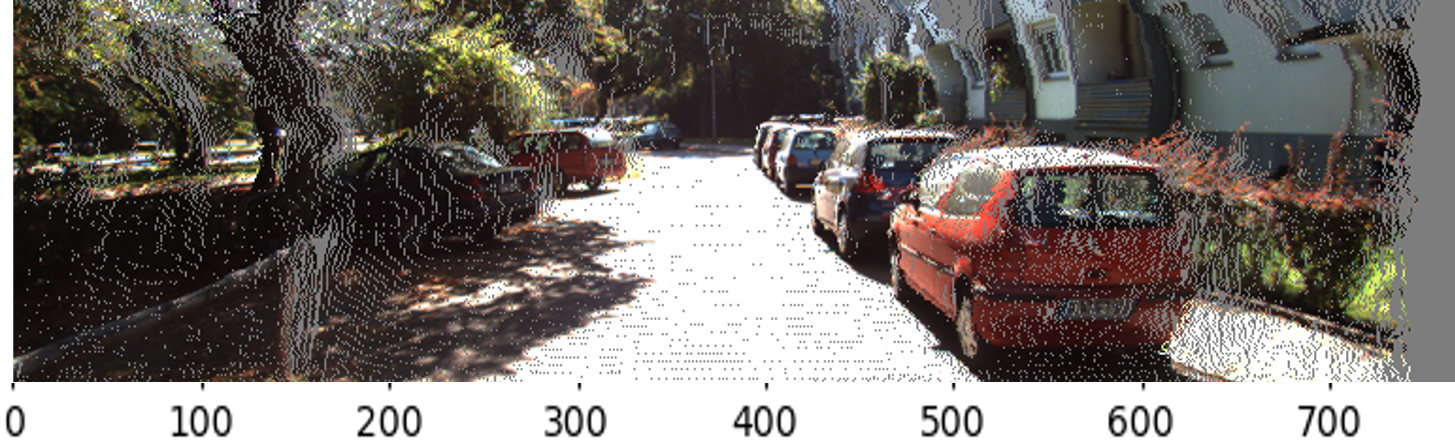} &
        \includegraphics[width=\sz\linewidth]{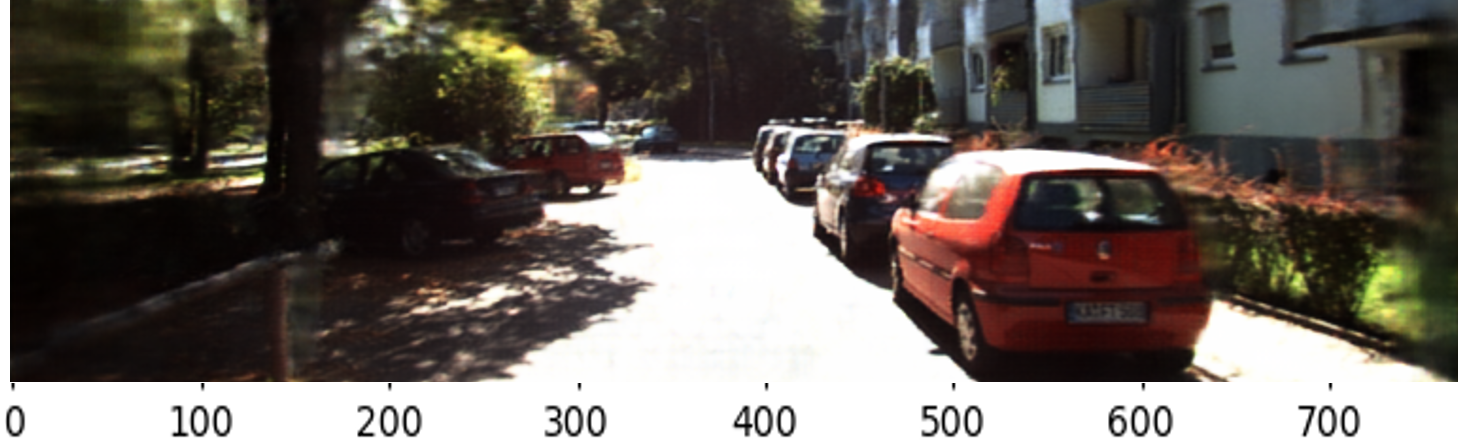} &
        \includegraphics[width=\sz\linewidth]{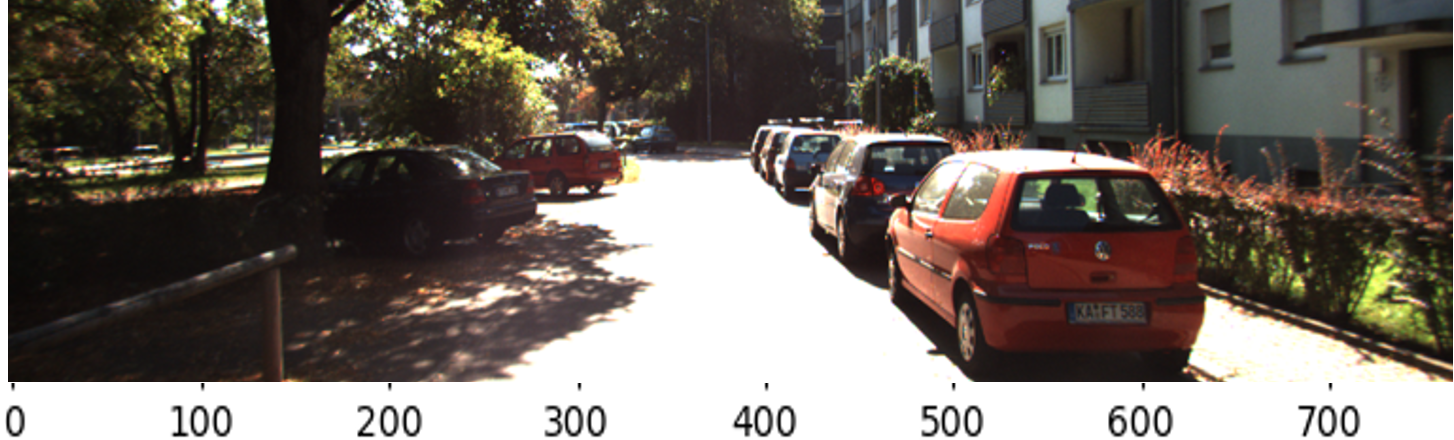}\\[\tabspace]
    \end{tabular}
    \begin{tabular}{ccc}
        RGB 1 & \textit{DepNet} & Depth GT (for \textit{DepNet} supervision)\\
        \includegraphics[width=\sz\linewidth]{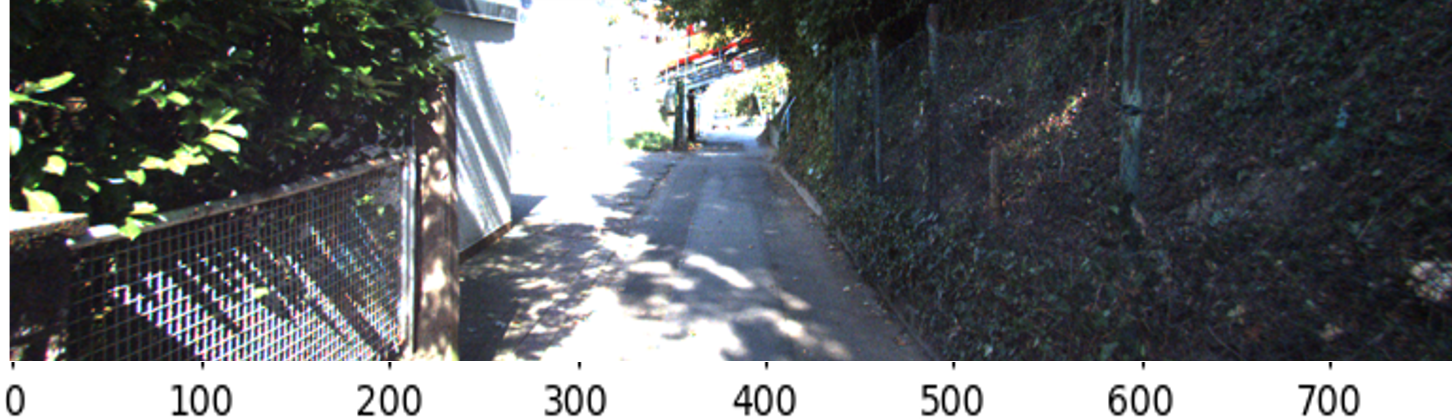} &
        \includegraphics[width=\sz\linewidth]{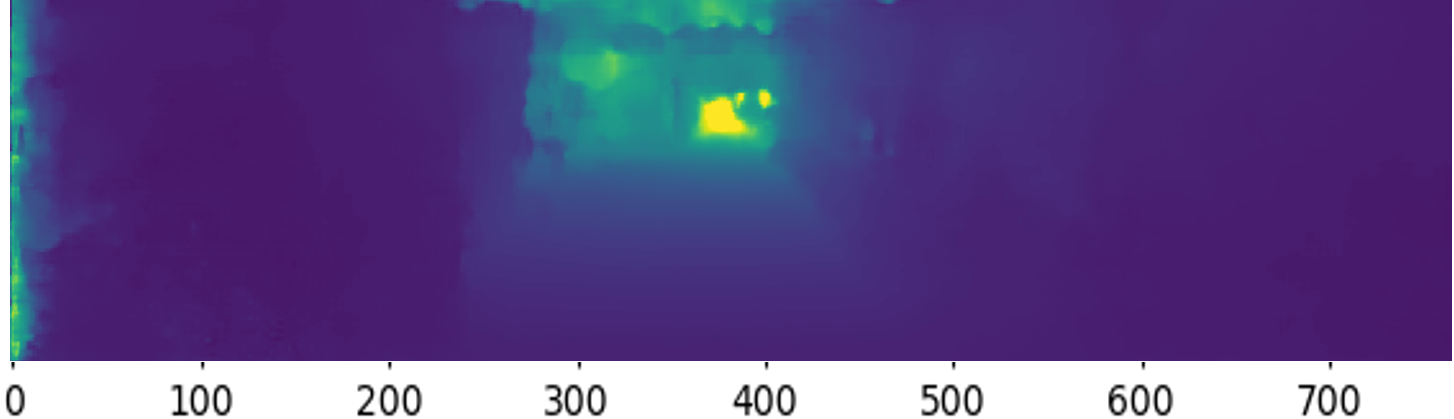} &
        \includegraphics[width=\sz\linewidth]{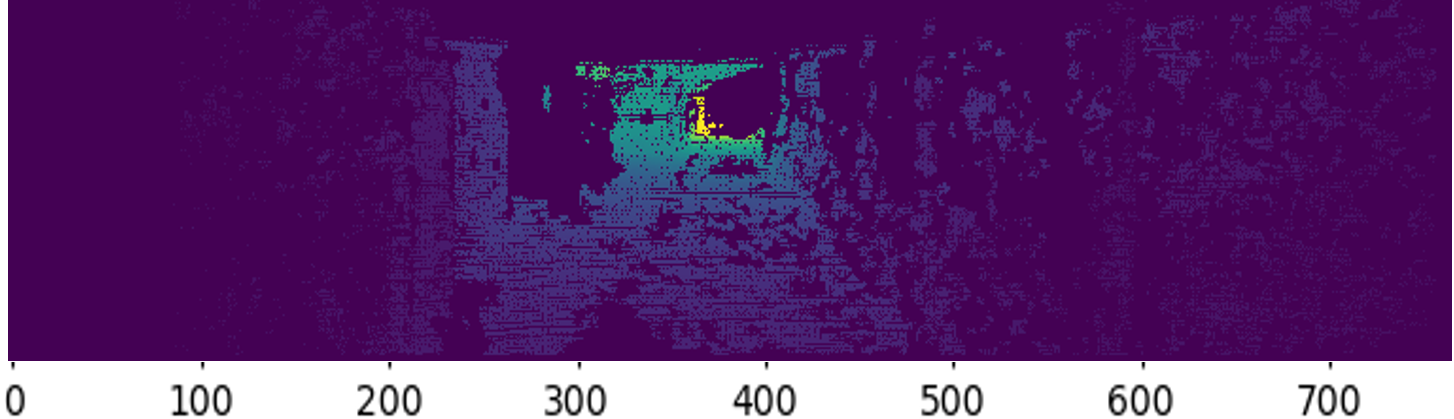}\\   
        View Transformation & \textit{SynNet} & RGB 2 (for \textit{SynNet} supervision)\\
        \includegraphics[width=\sz\linewidth]{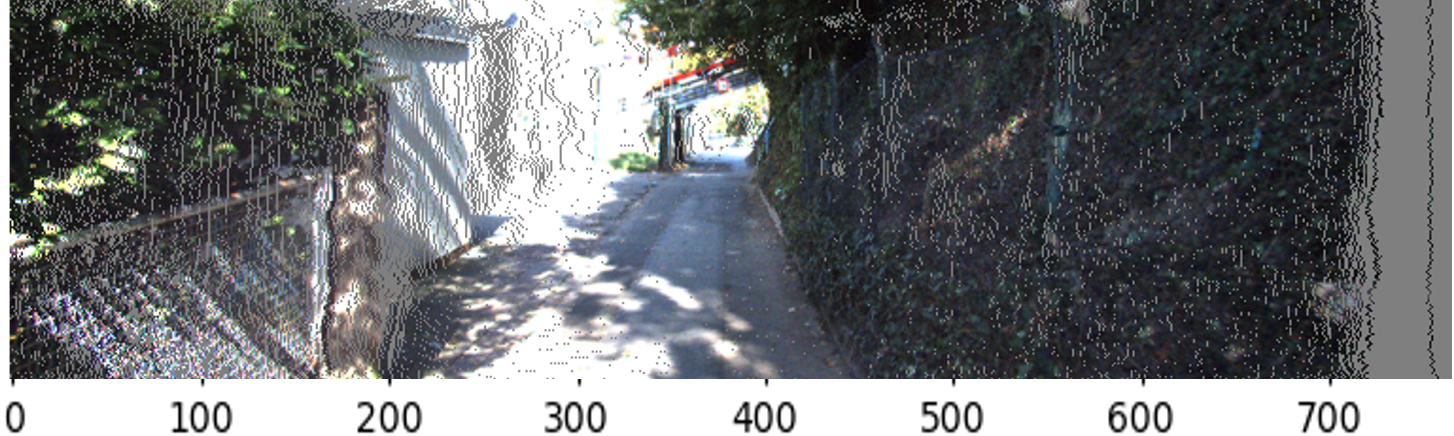} &
        \includegraphics[width=\sz\linewidth]{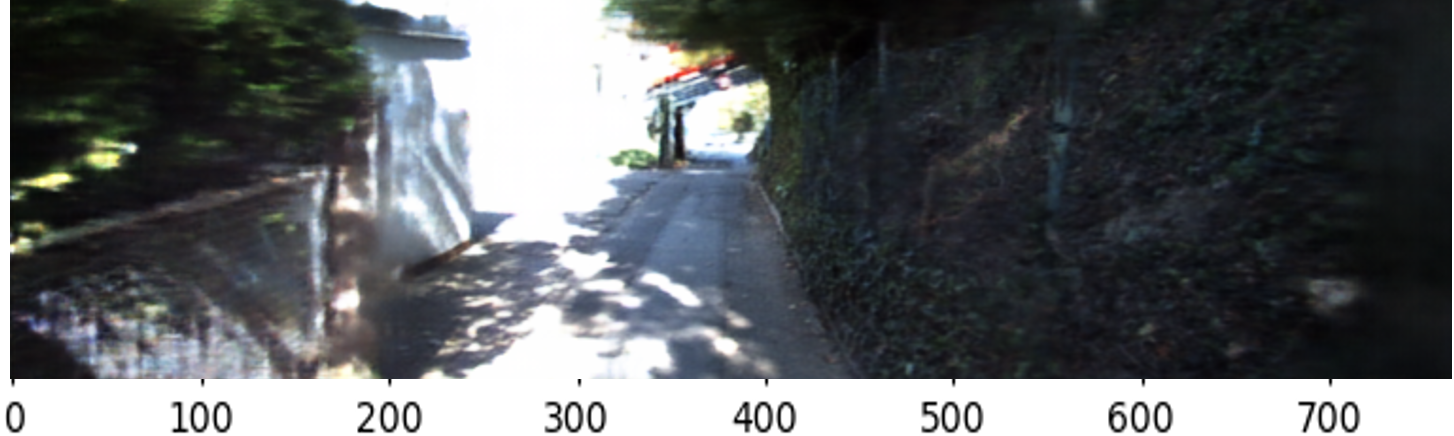} &
        \includegraphics[width=\sz\linewidth]{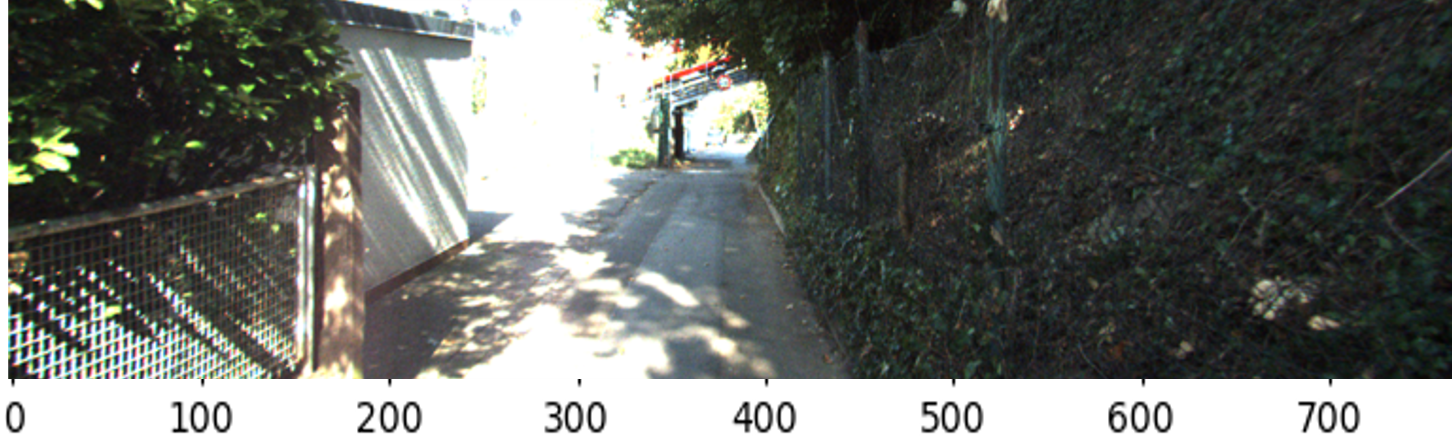}\\[\tabspace]
    \end{tabular} 
    \begin{tabular}{ccc}
        RGB 1 & \textit{DepNet} & Depth GT (for \textit{DepNet} supervision)\\
        \includegraphics[width=\sz\linewidth]{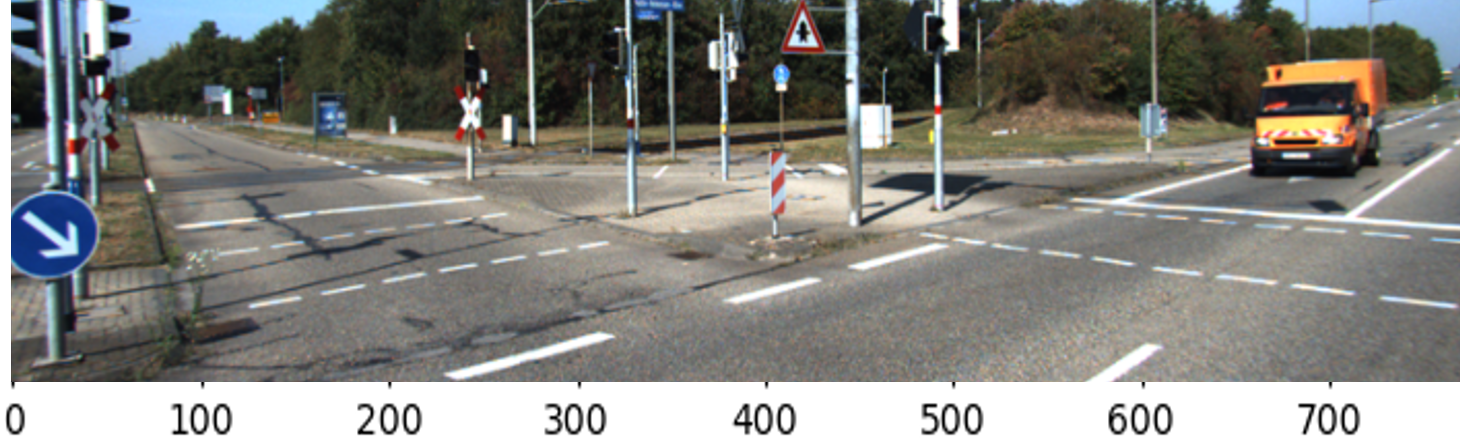} &
        \includegraphics[width=\sz\linewidth]{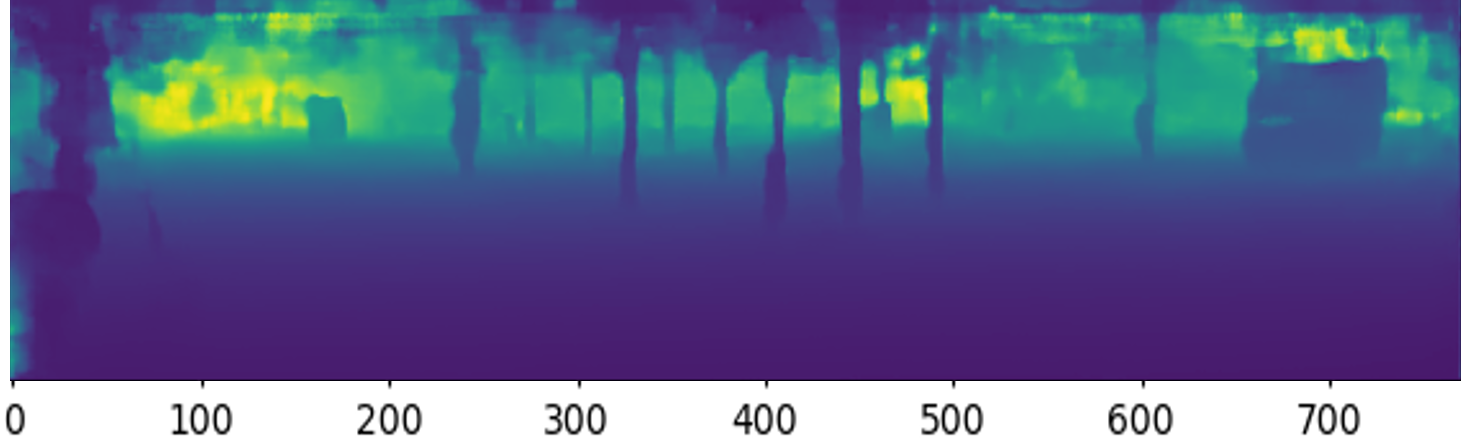} &
        \includegraphics[width=\sz\linewidth]{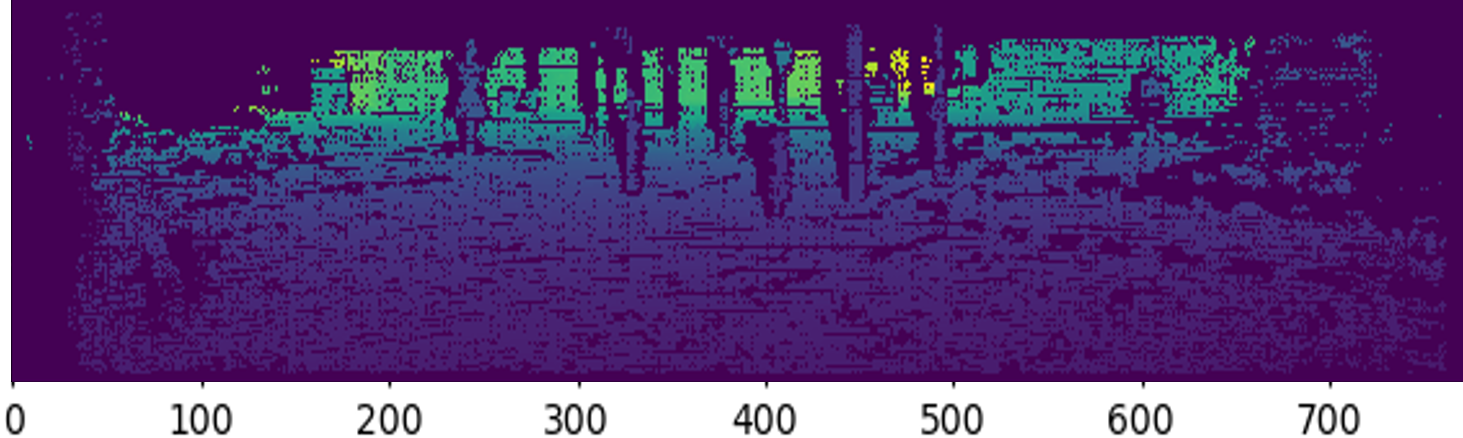}\\   
         View Transformation & \textit{SynNet} & RGB 2 (for \textit{SynNet} supervision)\\
        \includegraphics[width=\sz\linewidth]{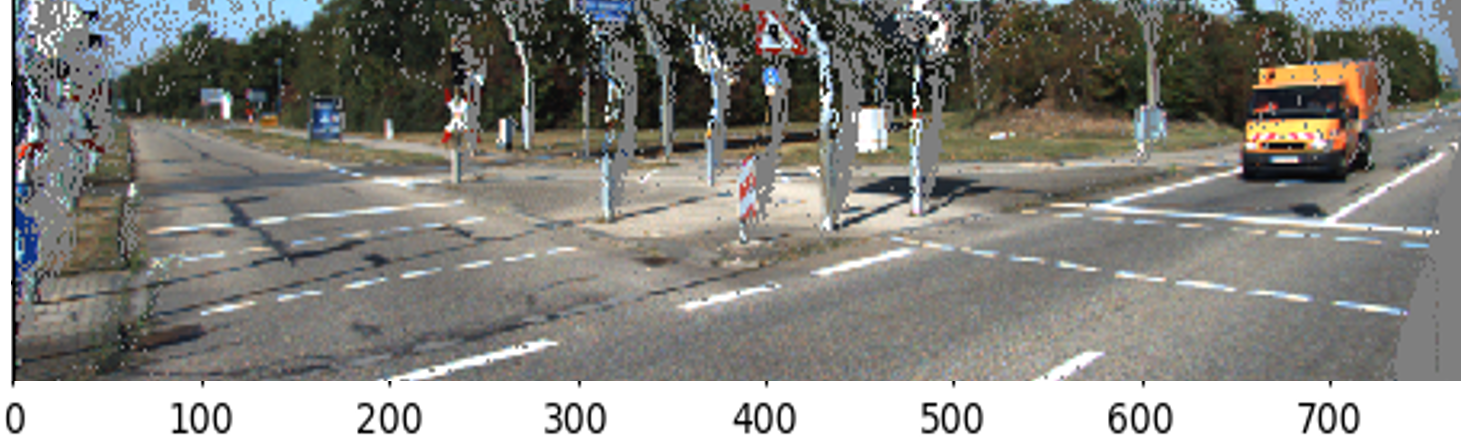} &
        \includegraphics[width=\sz\linewidth]{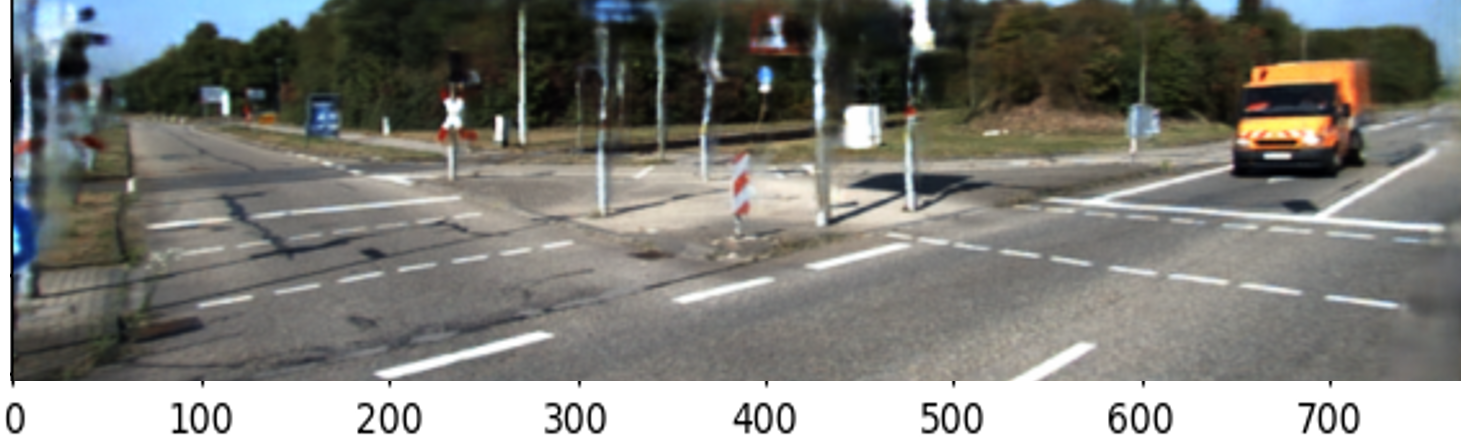} &
        \includegraphics[width=\sz\linewidth]{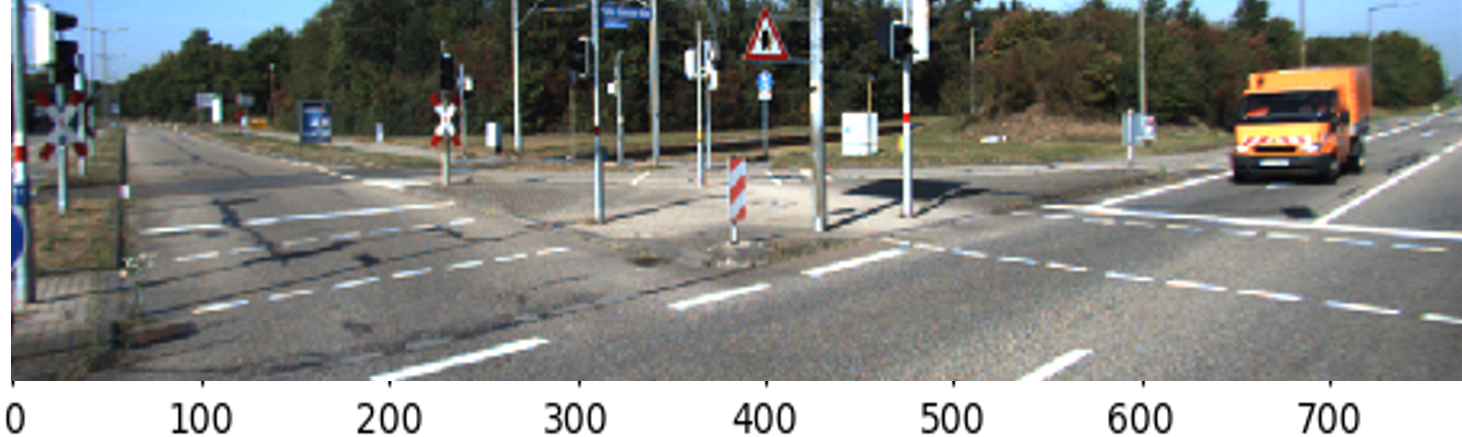}\\[\tabspace]
    \end{tabular} 
    \begin{tabular}{ccc}
        RGB 1 & \textit{DepNet} & Depth GT (for \textit{DepNet} supervision)\\
        \includegraphics[width=\sz\linewidth]{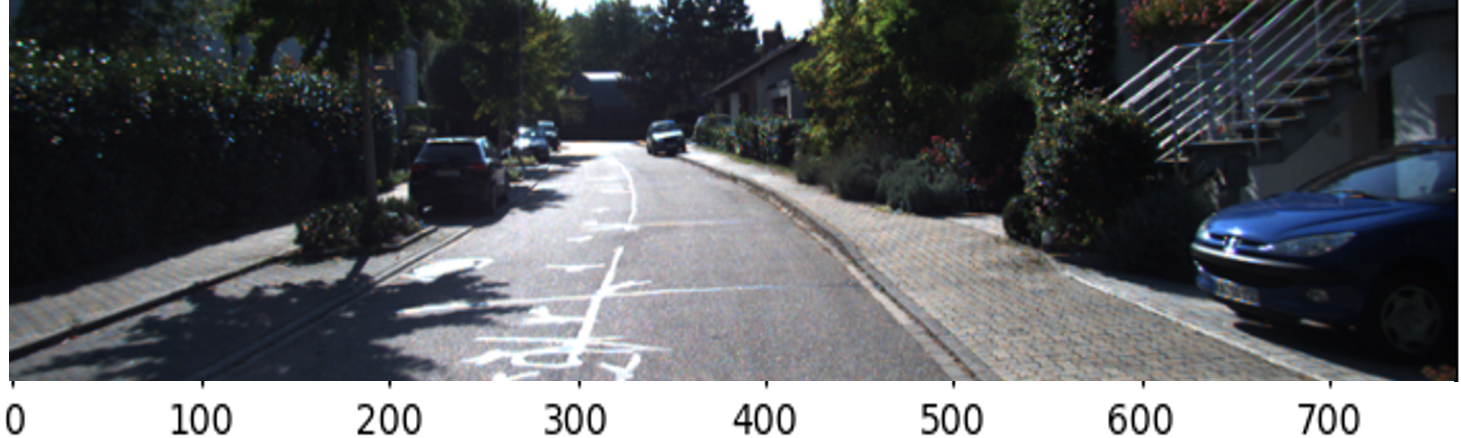} &
        \includegraphics[width=\sz\linewidth]{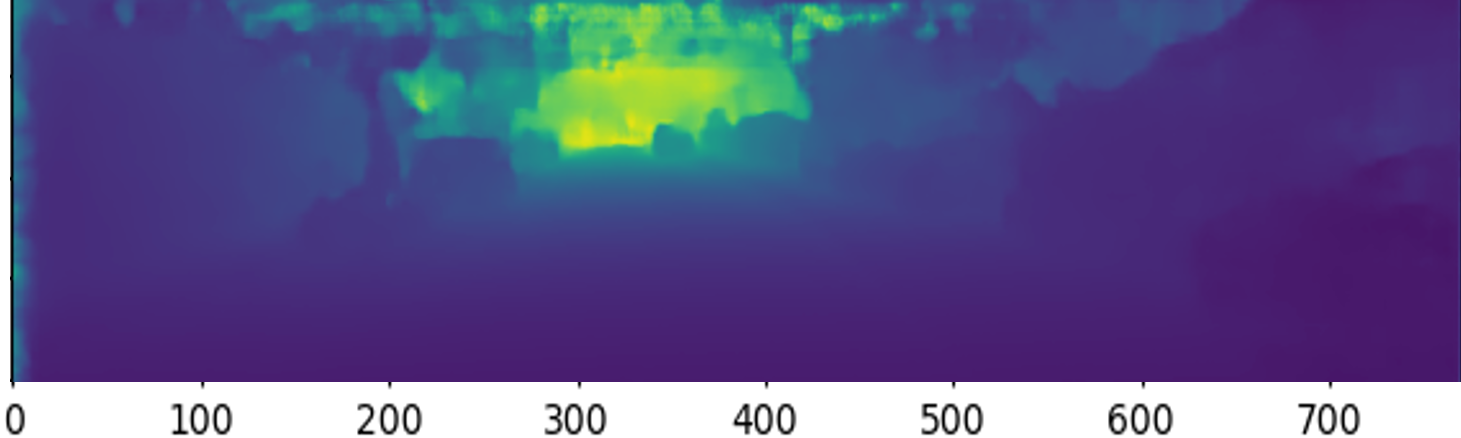} &
        \includegraphics[width=\sz\linewidth]{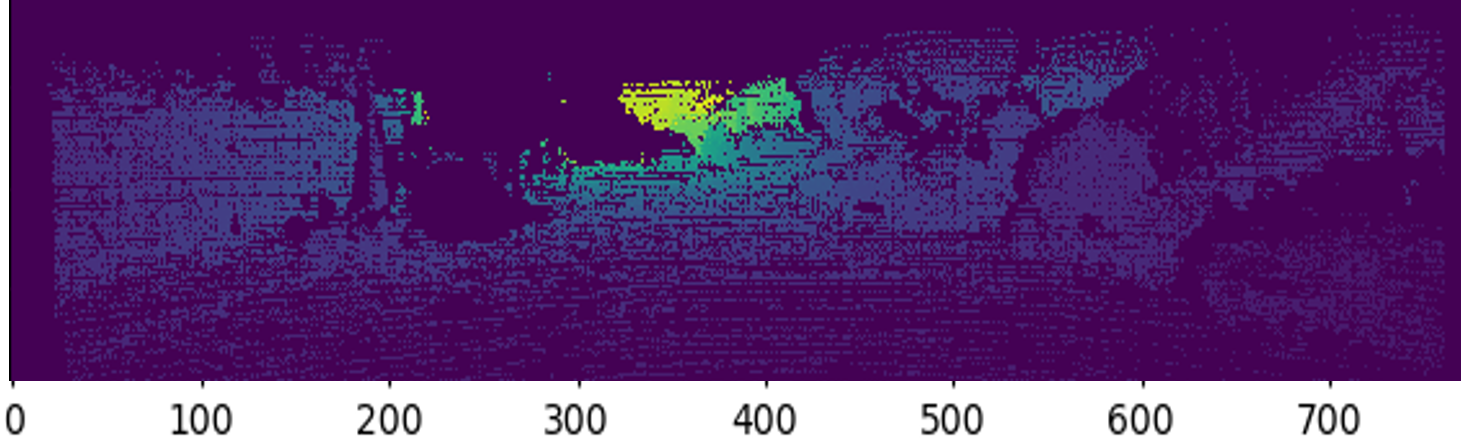}\\   
         View Transformation & \textit{SynNet} & RGB 2 (for \textit{SynNet} supervision)\\
        \includegraphics[width=\sz\linewidth]{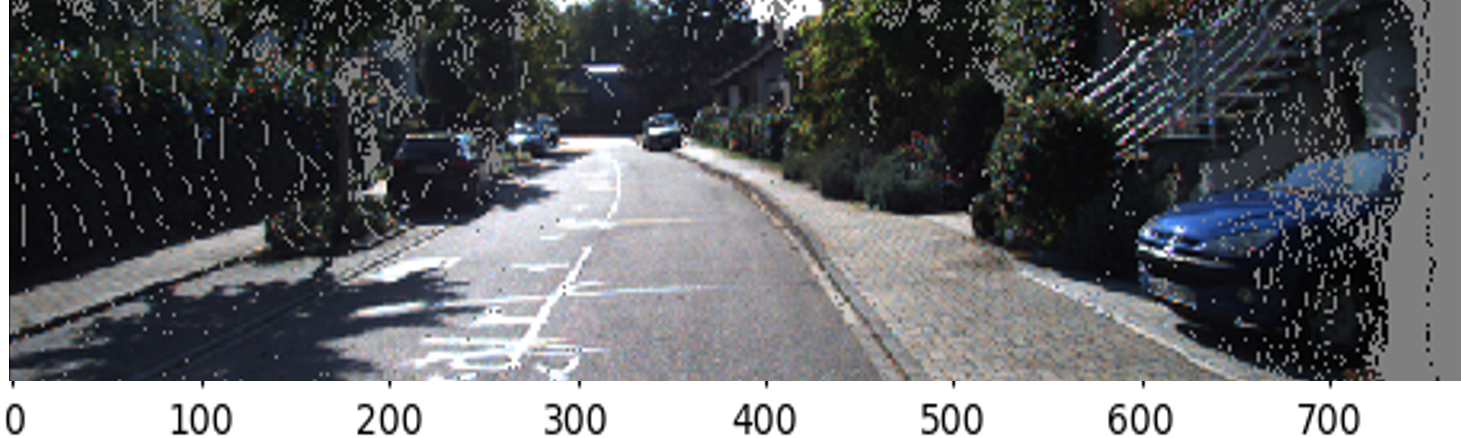} &
        \includegraphics[width=\sz\linewidth]{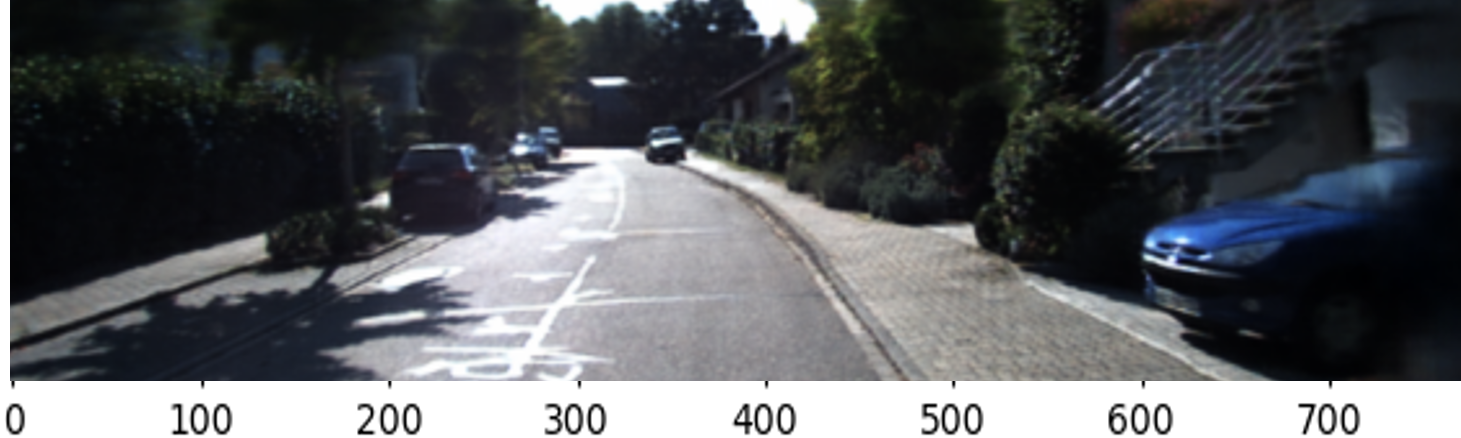} &
        \includegraphics[width=\sz\linewidth]{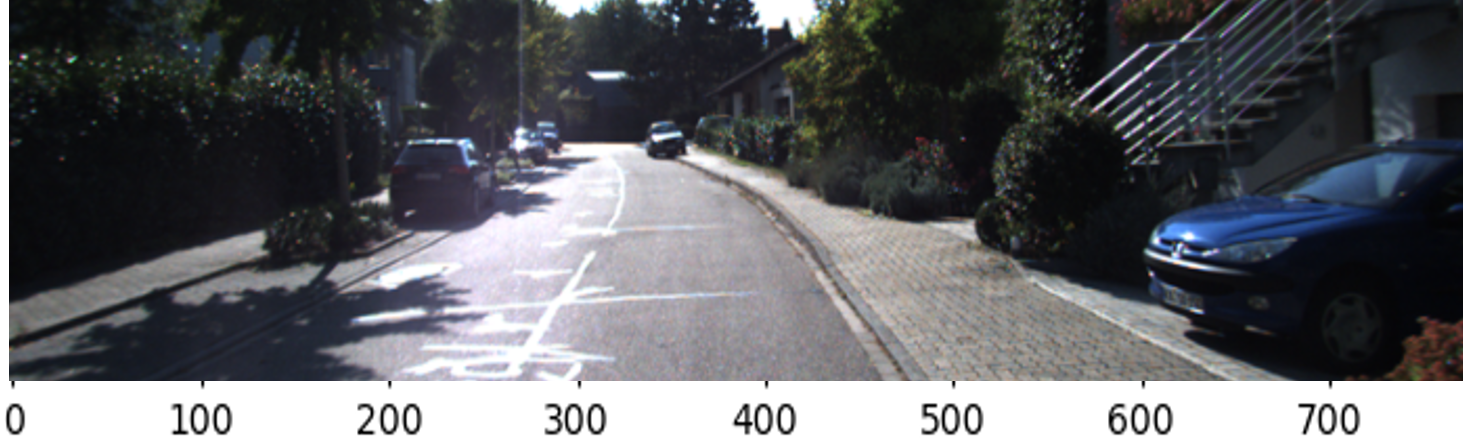}\\[\tabspace]
    \end{tabular} 
    \begin{tabular}{ccc}
        RGB 1 & \textit{DepNet} & Depth GT (for \textit{DepNet} supervision)\\
        \includegraphics[width=\sz\linewidth]{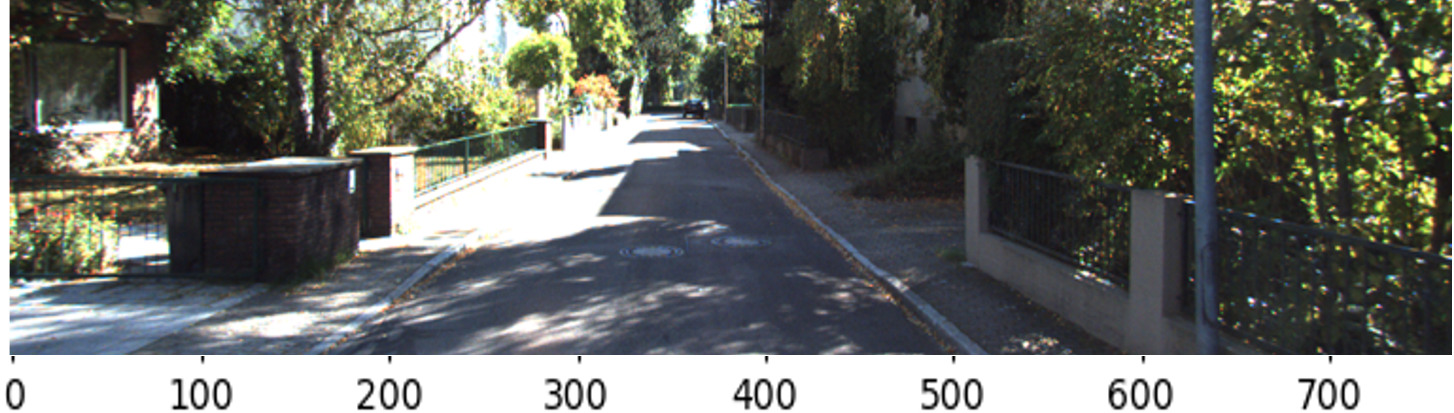} &
        \includegraphics[width=\sz\linewidth]{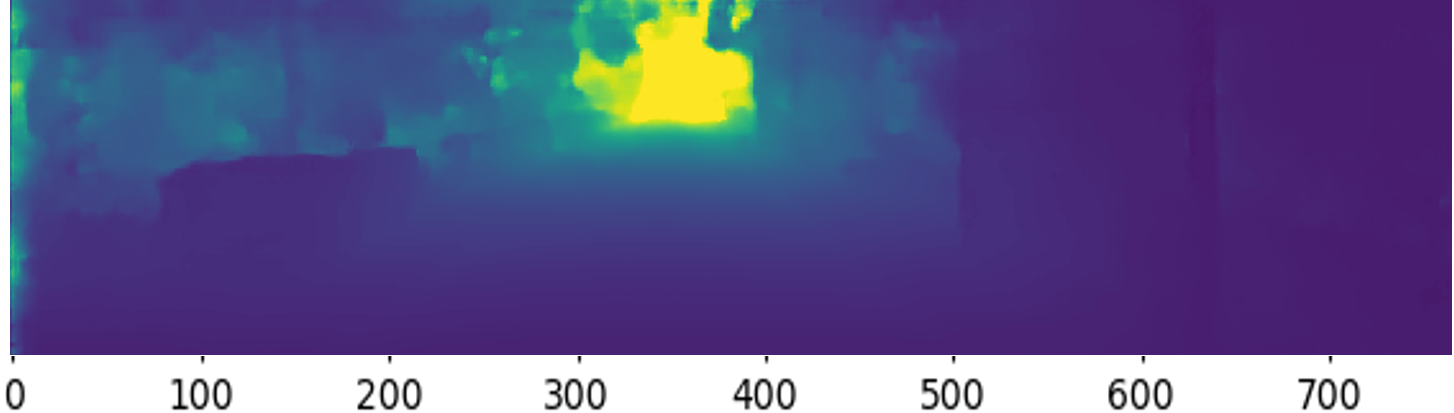} &
        \includegraphics[width=\sz\linewidth]{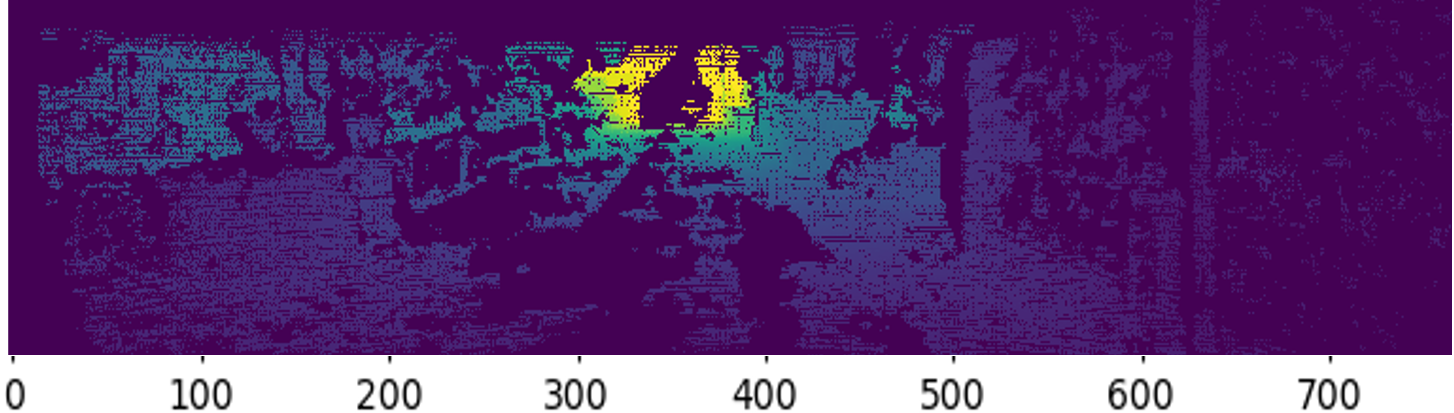}\\   
         View Transformation & \textit{SynNet} & RGB 2 (for \textit{SynNet} supervision)\\
        \includegraphics[width=\sz\linewidth]{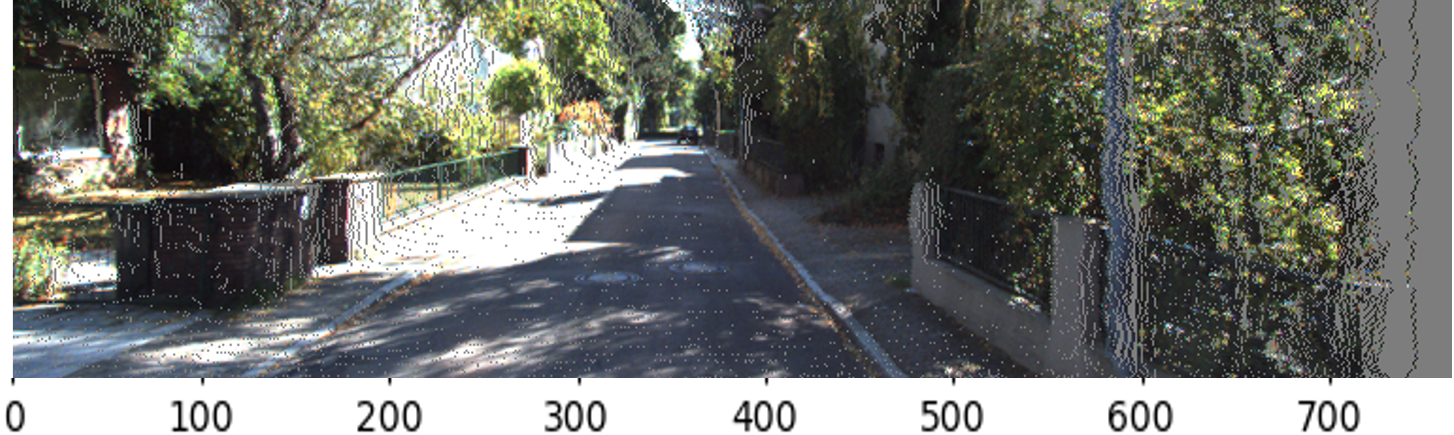} &
        \includegraphics[width=\sz\linewidth]{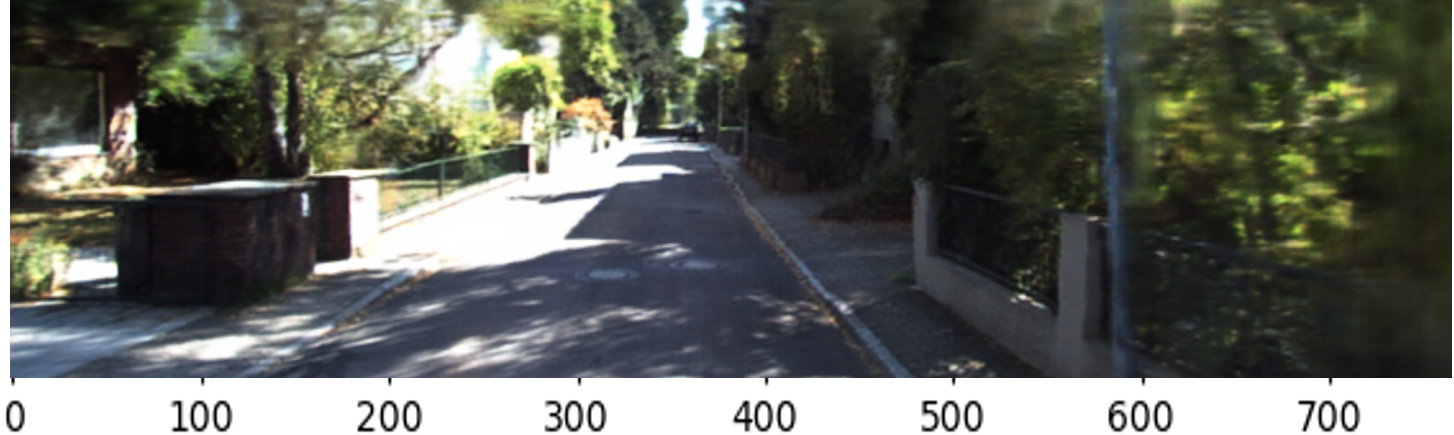} &
        \includegraphics[width=\sz\linewidth]{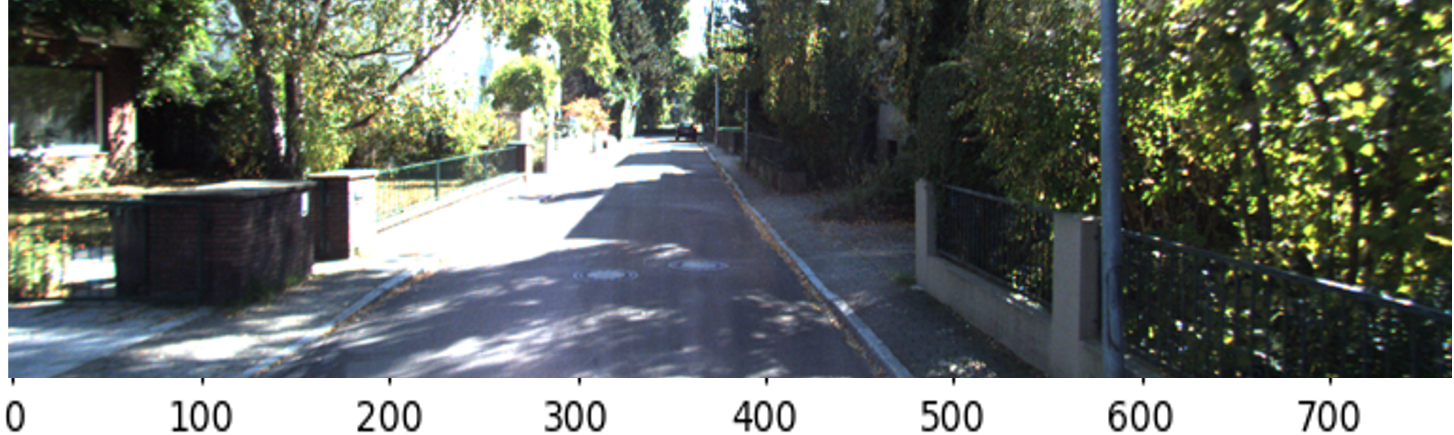}\\[\tabspace]
    \end{tabular} 
    \vspace*{-3pt}
    \caption{Qualitative predictions from each of the steps of the proposed pipeline on the KITTI \cite{Kitti} dataset. RGB 1 and RGB 2 are consecutive views or the corresponding stereo image. Color scale goes from 0 (purple) to 80 meters (yellow). For better appreciation, we recommend to zoom in on each image.}
    \label{fig:Pipeline_1}
    \vspace*{-13pt}
\end{figure*}

\clearpage
\begin{figure*}[!htb]
    \centering
    \footnotesize
    \setlength{\tabcolsep}{1pt}
	\renewcommand{\arraystretch}{0.8}
	\newcommand{\sz}{0.33}
	\newcommand{\tabspace}{6pt}
	\vspace*{-14pt}
    \begin{tabular}{ccc}
        RGB 1 & \textit{DepNet} & Depth GT (for \textit{DepNet} supervision)\\
        \includegraphics[width=\sz\linewidth]{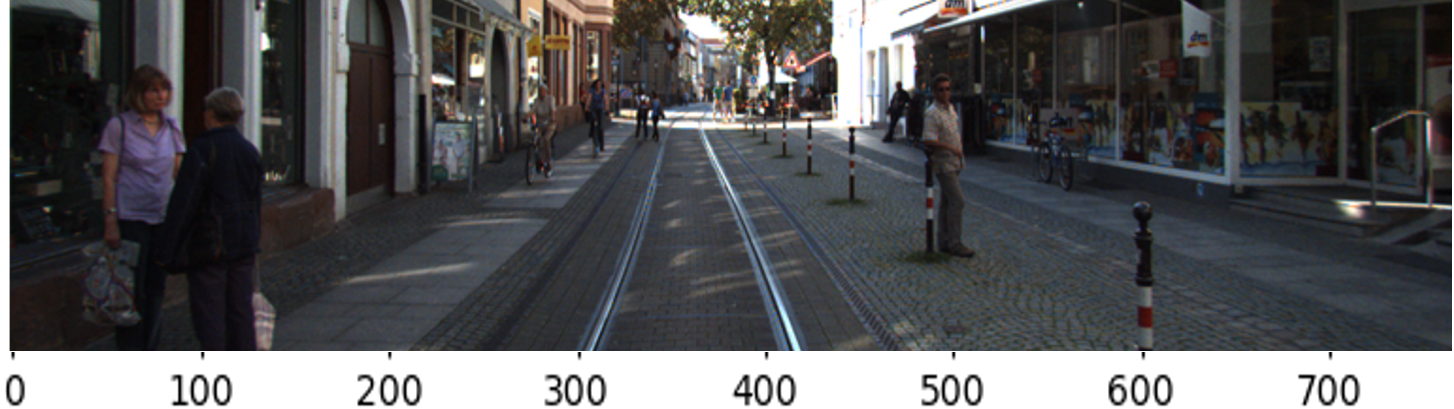} &
        \includegraphics[width=\sz\linewidth]{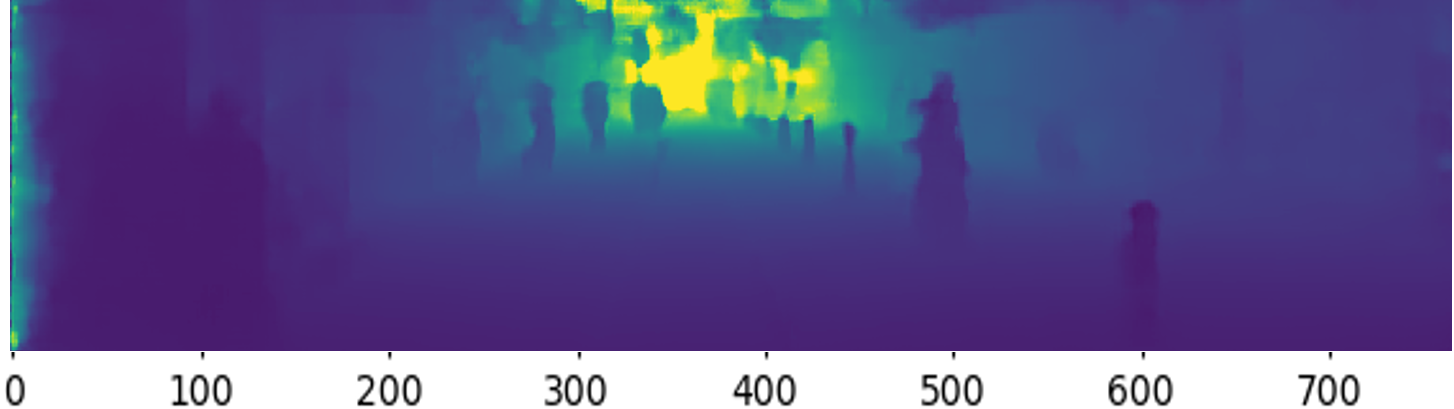} &
        \includegraphics[width=\sz\linewidth]{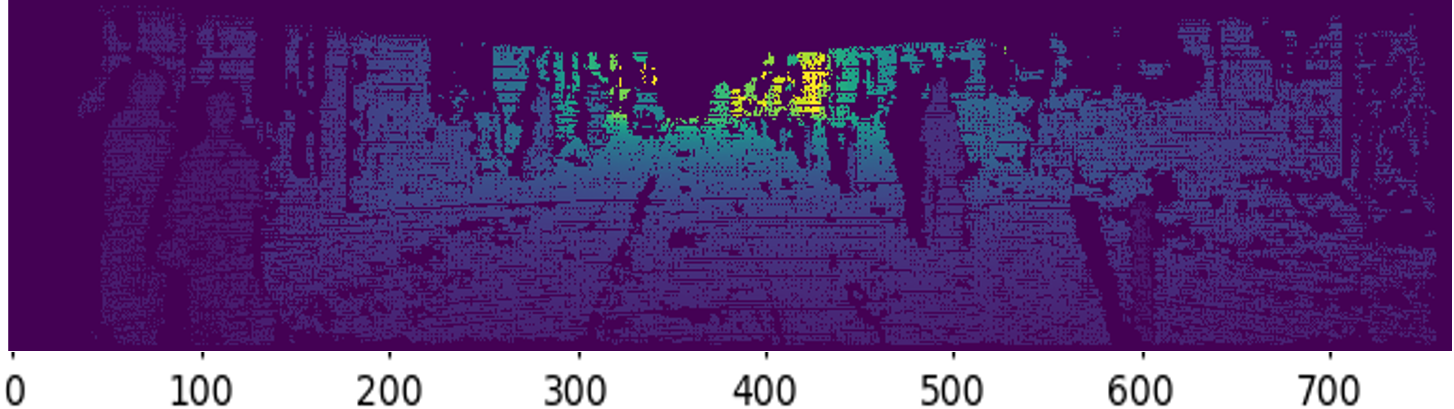}\\   
        View Transformation & \textit{SynNet} & RGB 2 (for \textit{SynNet} supervision)\\
        \includegraphics[width=\sz\linewidth]{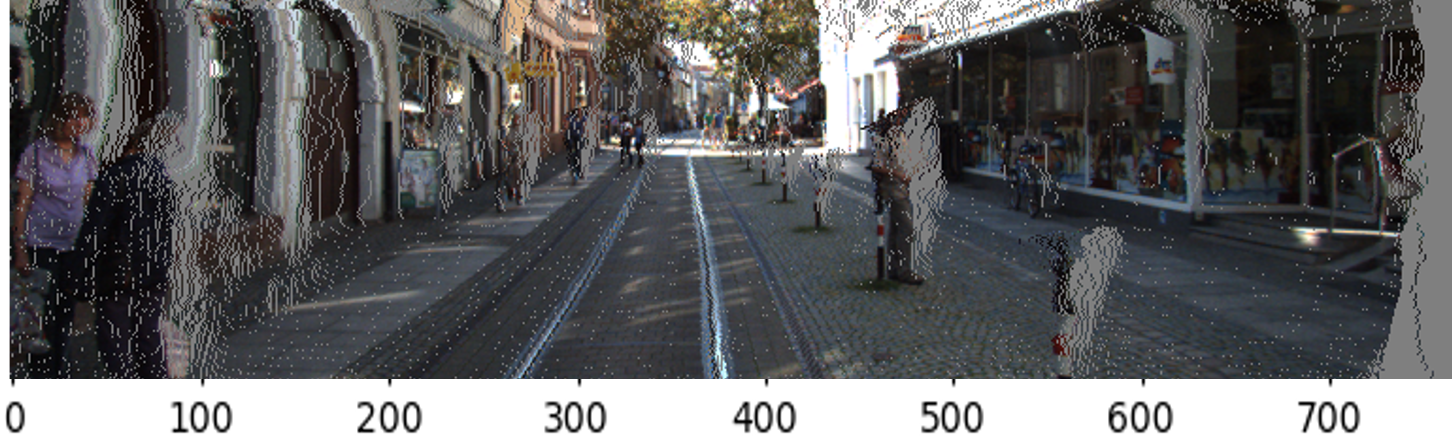} &
        \includegraphics[width=\sz\linewidth]{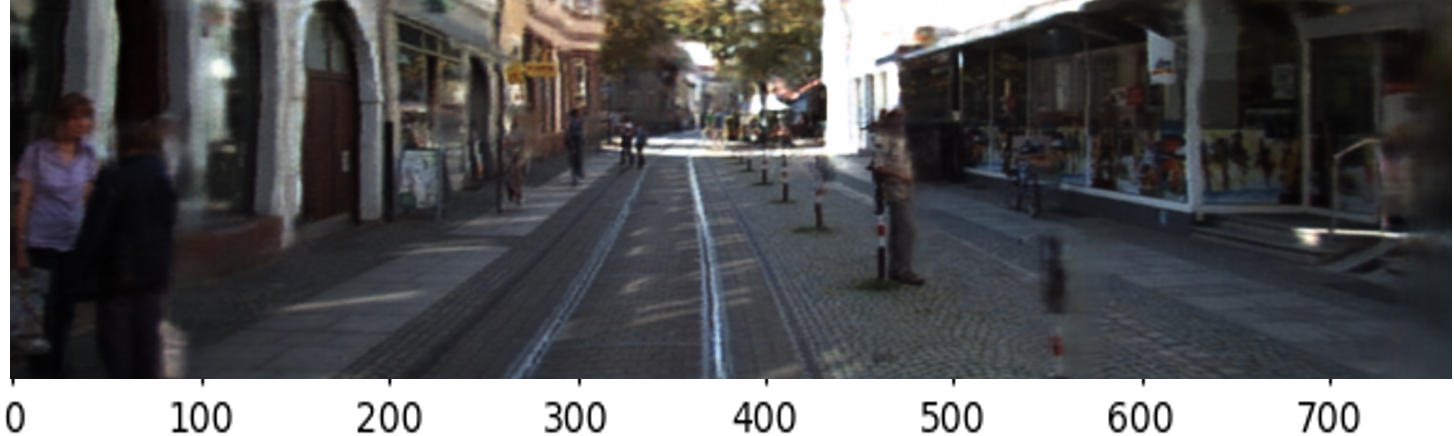} &
        \includegraphics[width=\sz\linewidth]{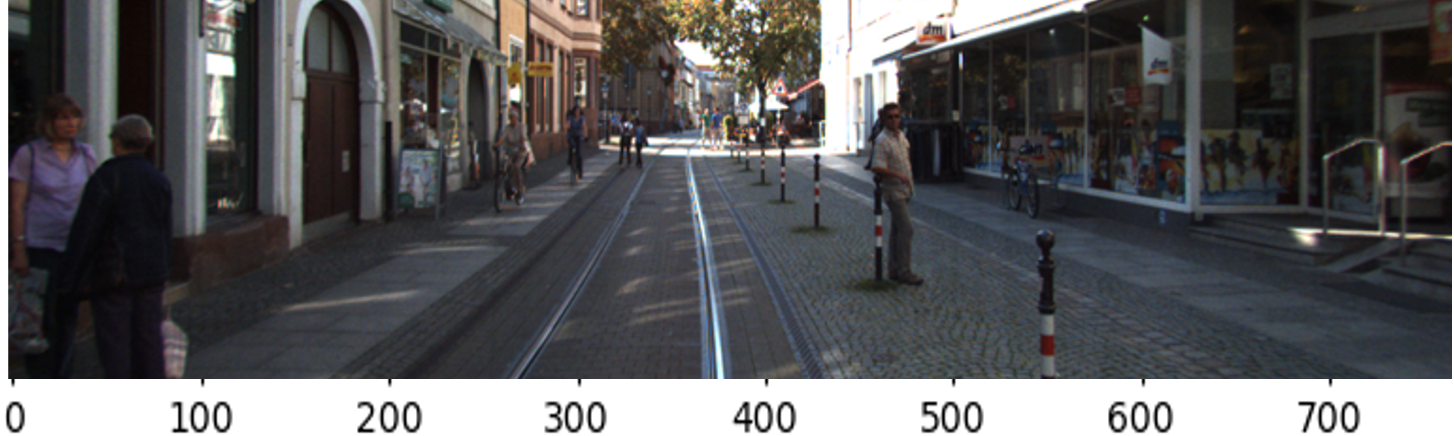}\\[\tabspace]
    \end{tabular} 
    \begin{tabular}{ccc}
        RGB 1 & \textit{DepNet} & Depth GT (for \textit{DepNet} supervision)\\
        \includegraphics[width=\sz\linewidth]{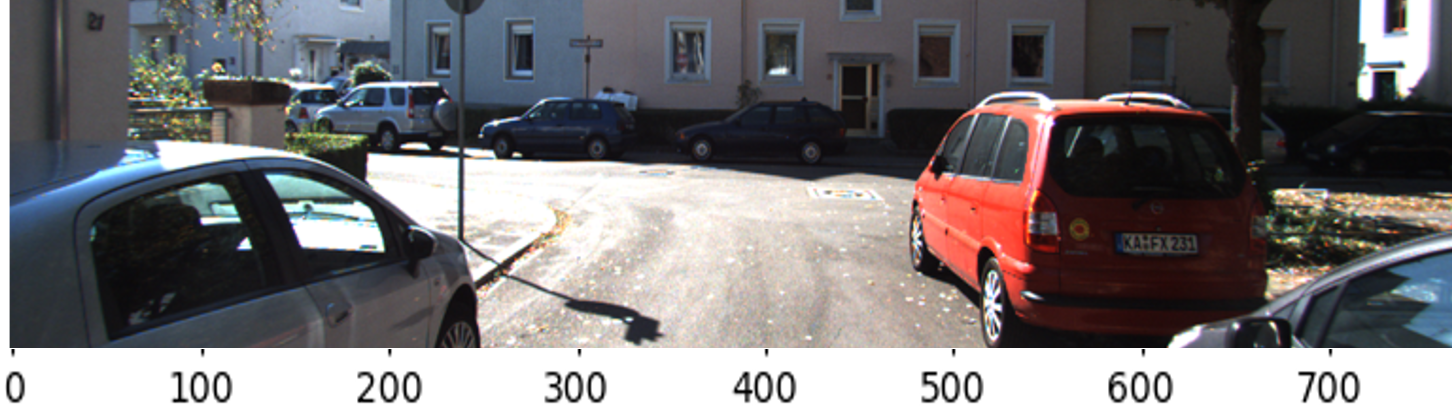} &
        \includegraphics[width=\sz\linewidth]{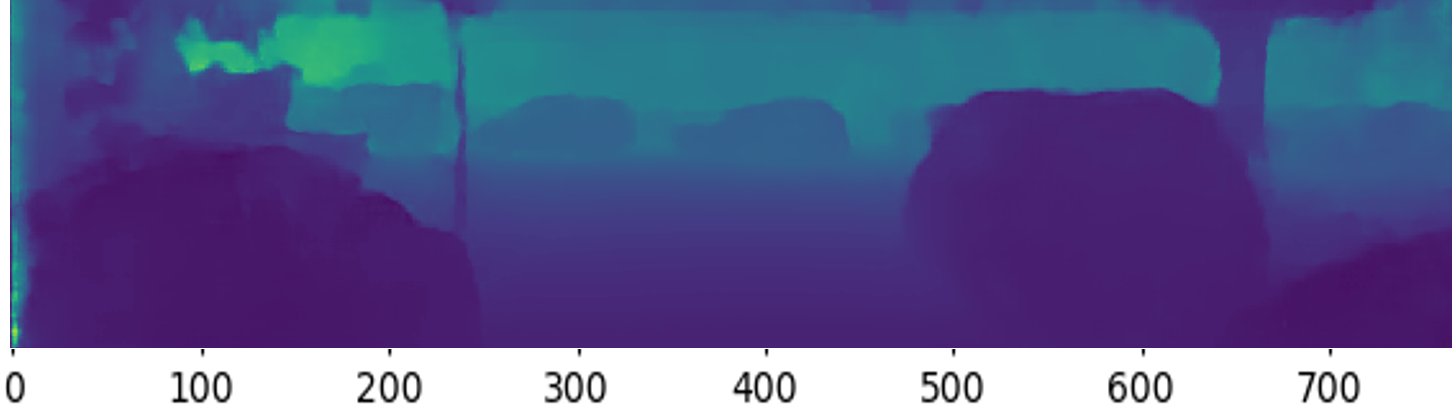} &
        \includegraphics[width=\sz\linewidth]{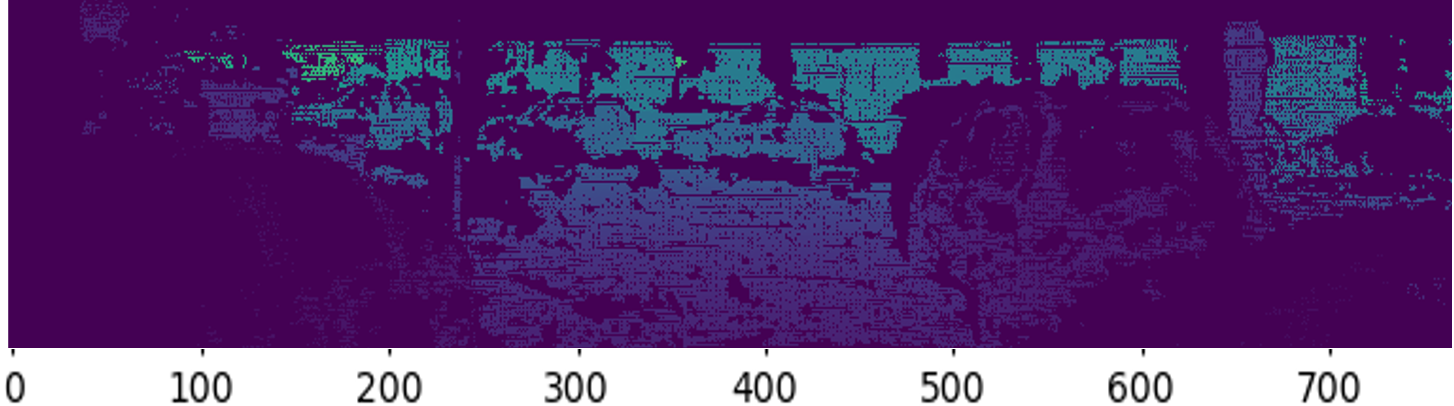}\\   
        View Transformation & \textit{SynNet} & RGB 2 (for \textit{SynNet} supervision)\\
        \includegraphics[width=\sz\linewidth]{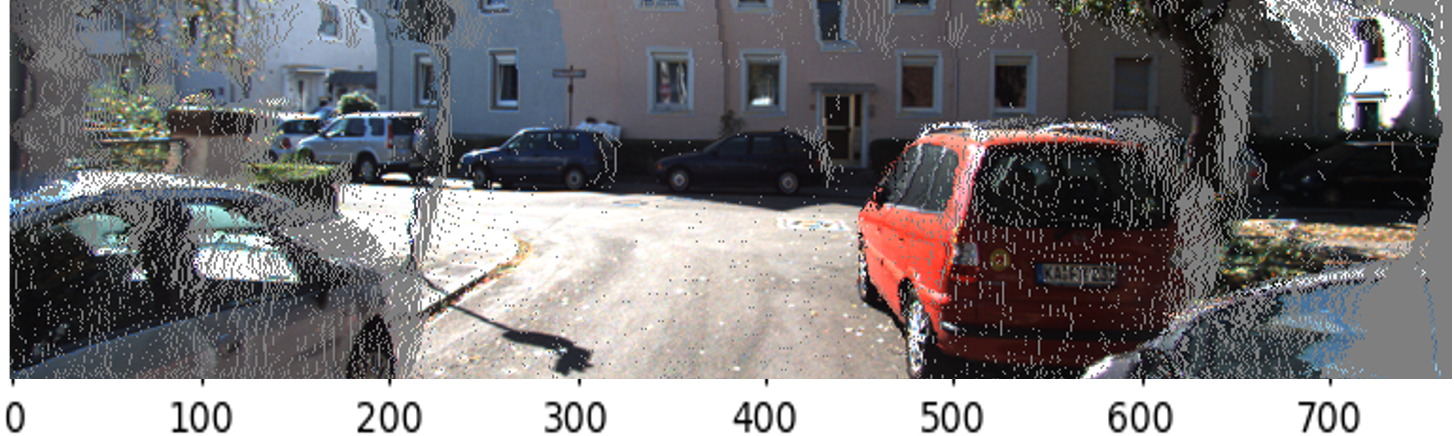} &
        \includegraphics[width=\sz\linewidth]{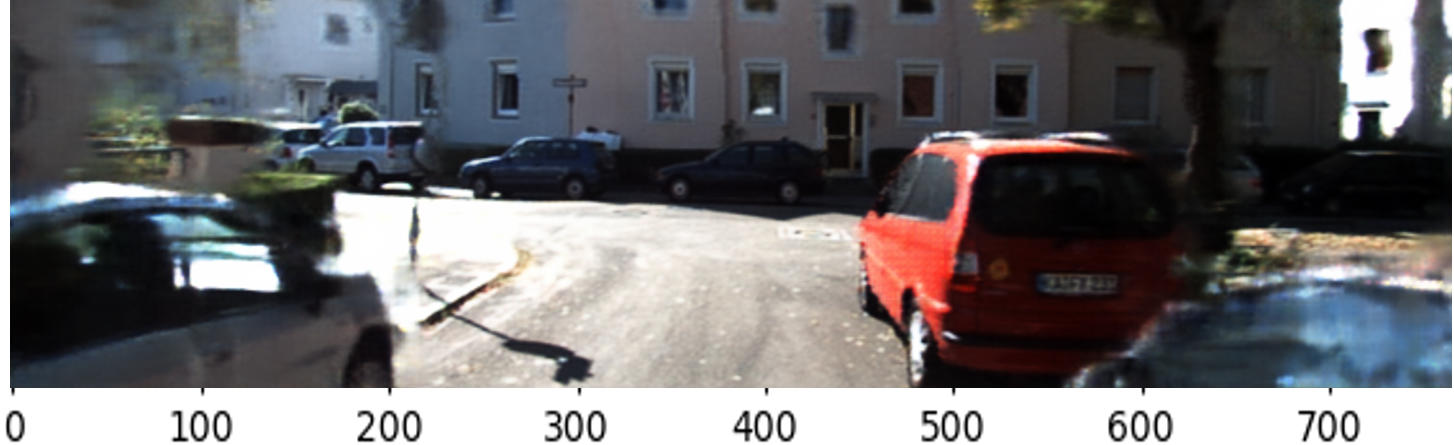} &
        \includegraphics[width=\sz\linewidth]{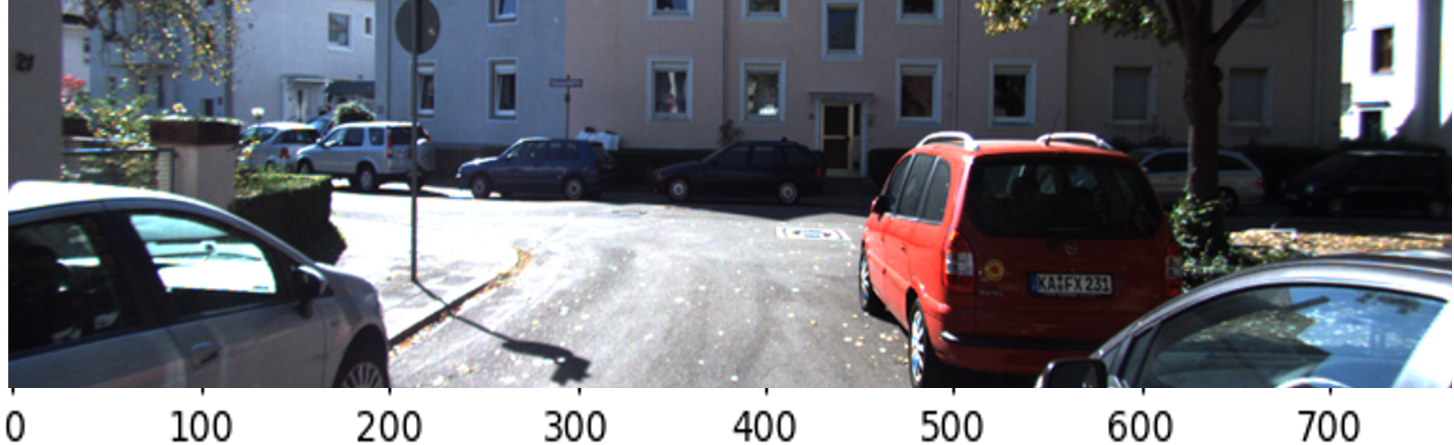}\\[\tabspace]
    \end{tabular} 
    \begin{tabular}{ccc}
        RGB 1 & \textit{DepNet} & Depth GT (for \textit{DepNet} supervision)\\
        \includegraphics[width=\sz\linewidth]{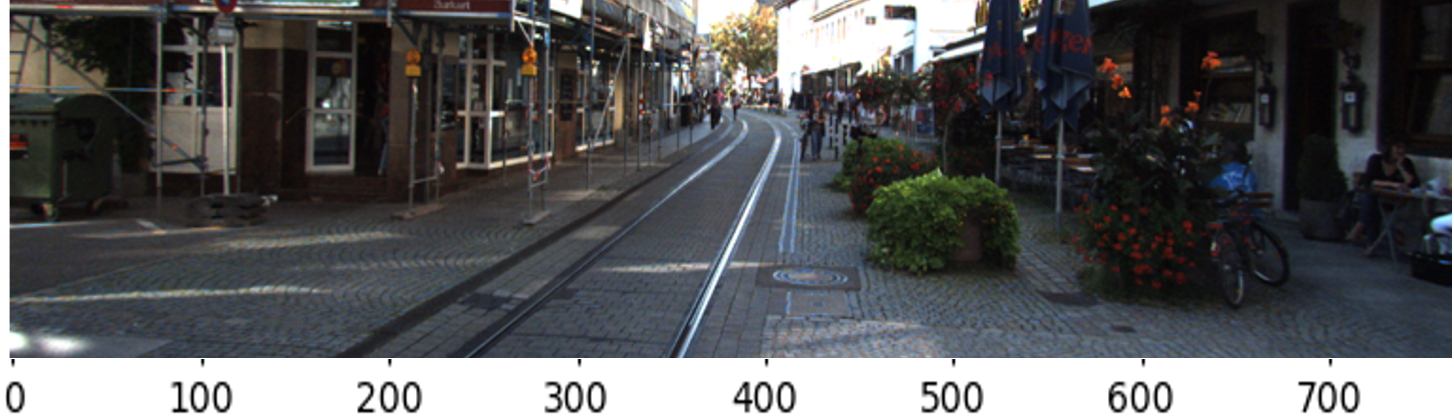} &
        \includegraphics[width=\sz\linewidth]{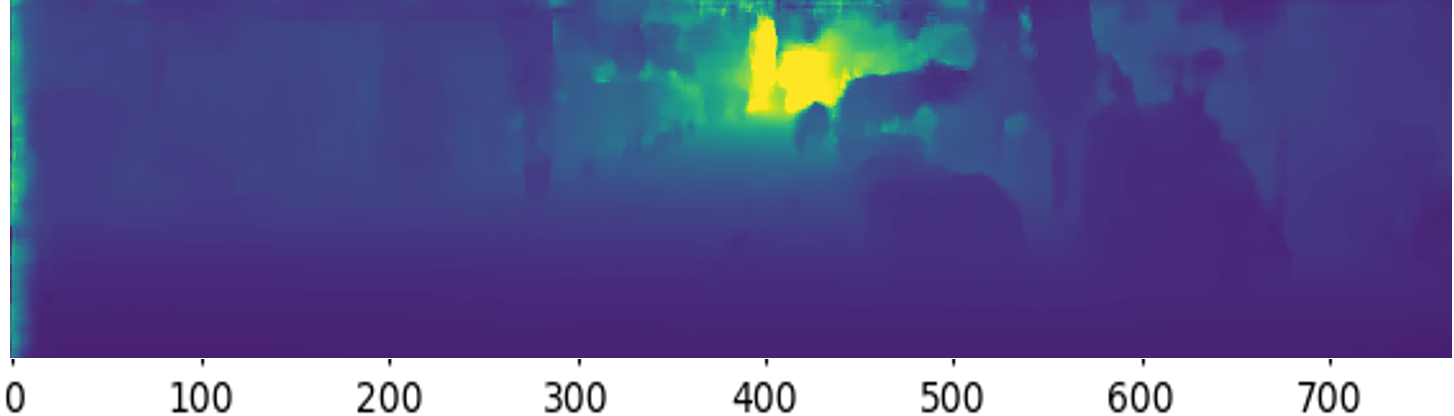} &
        \includegraphics[width=\sz\linewidth]{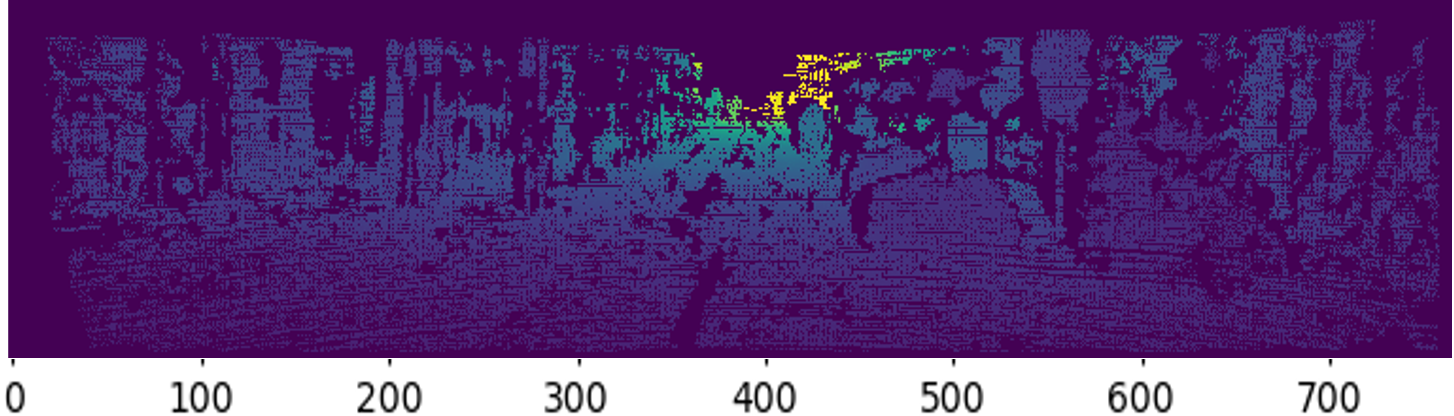}\\   
        View Transformation & \textit{SynNet} & RGB 2 (for \textit{SynNet} supervision)\\
        \includegraphics[width=\sz\linewidth]{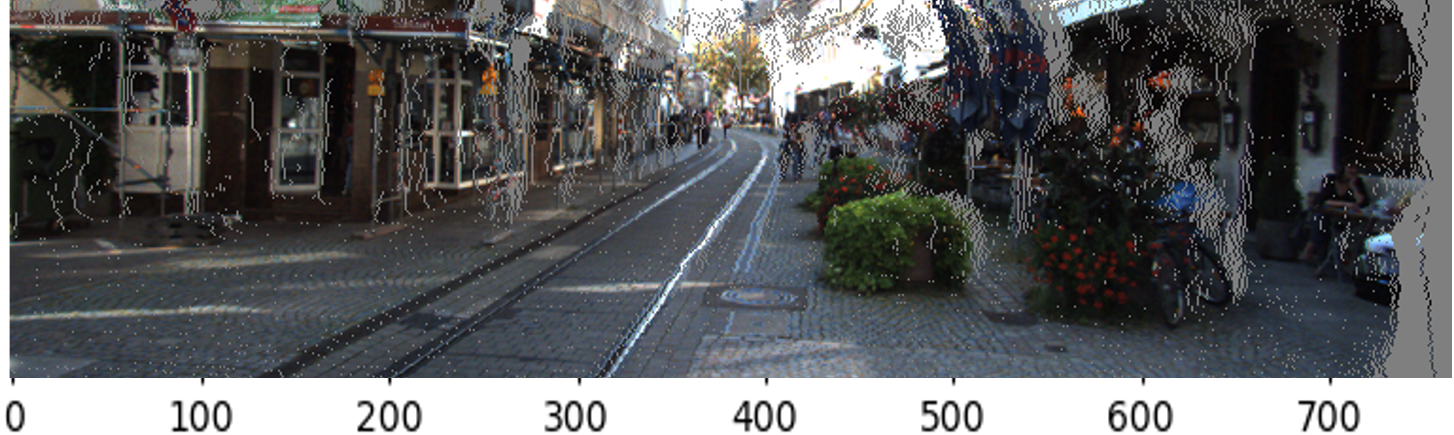} &
        \includegraphics[width=\sz\linewidth]{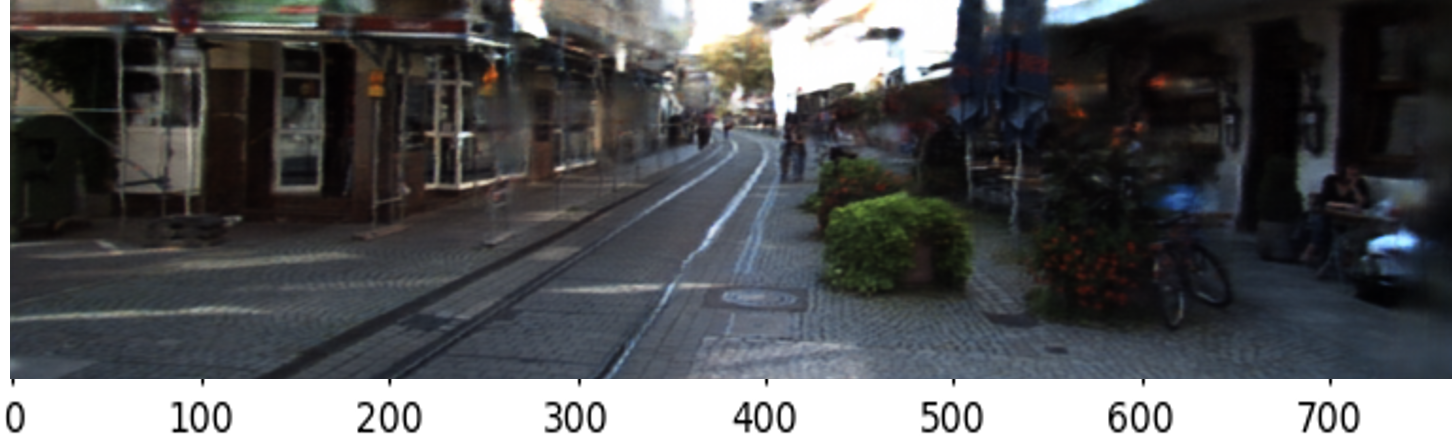} &
        \includegraphics[width=\sz\linewidth]{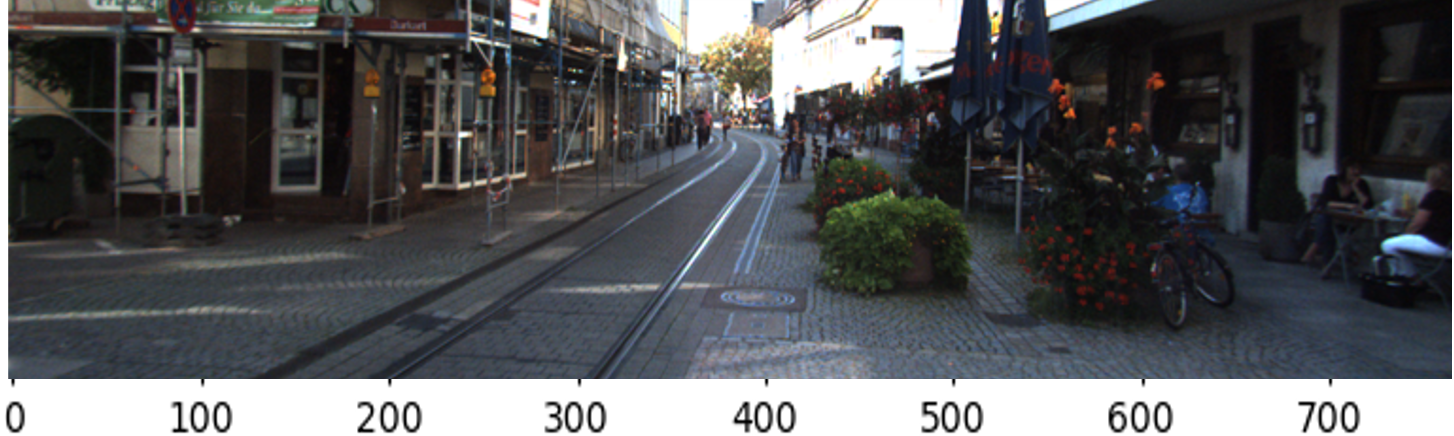}\\[\tabspace]
    \end{tabular} 
    \begin{tabular}{ccc}
        RGB 1 & \textit{DepNet} & Depth GT (for \textit{DepNet} supervision)\\
        \includegraphics[width=\sz\linewidth]{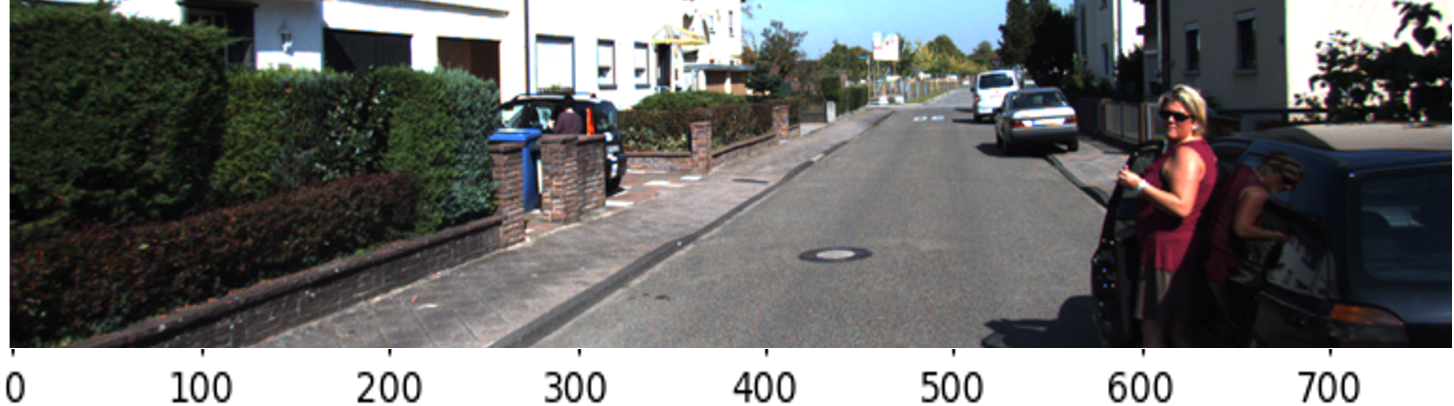} &
        \includegraphics[width=\sz\linewidth]{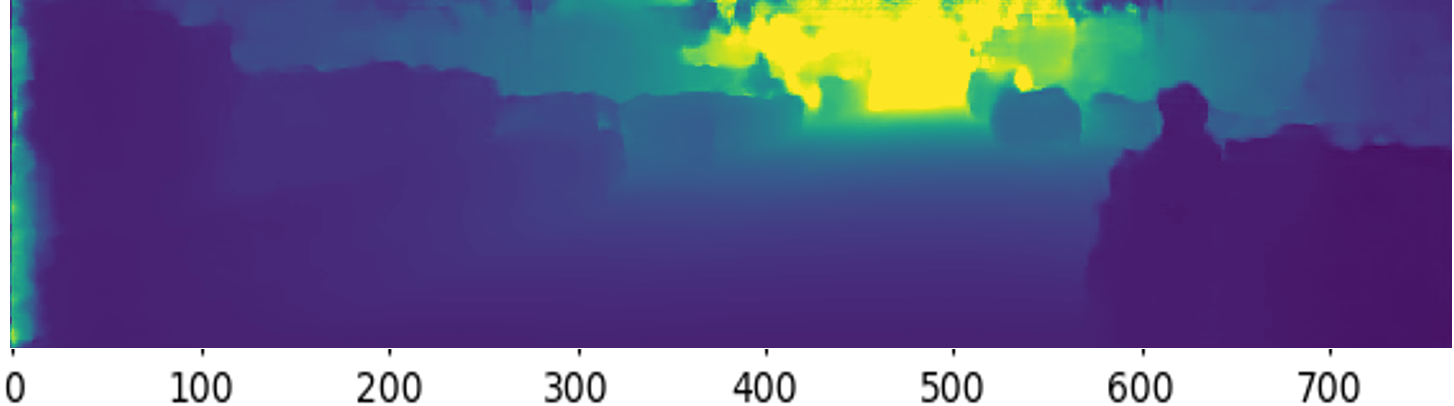} &
        \includegraphics[width=\sz\linewidth]{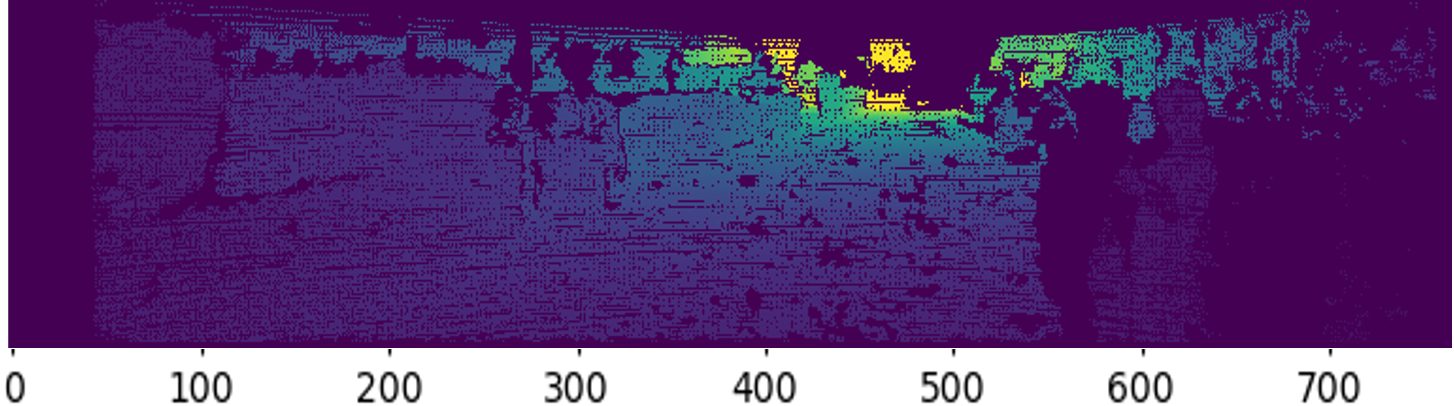}\\   
         View Transformation & \textit{SynNet} & RGB 2 (for \textit{SynNet} supervision)\\
        \includegraphics[width=\sz\linewidth]{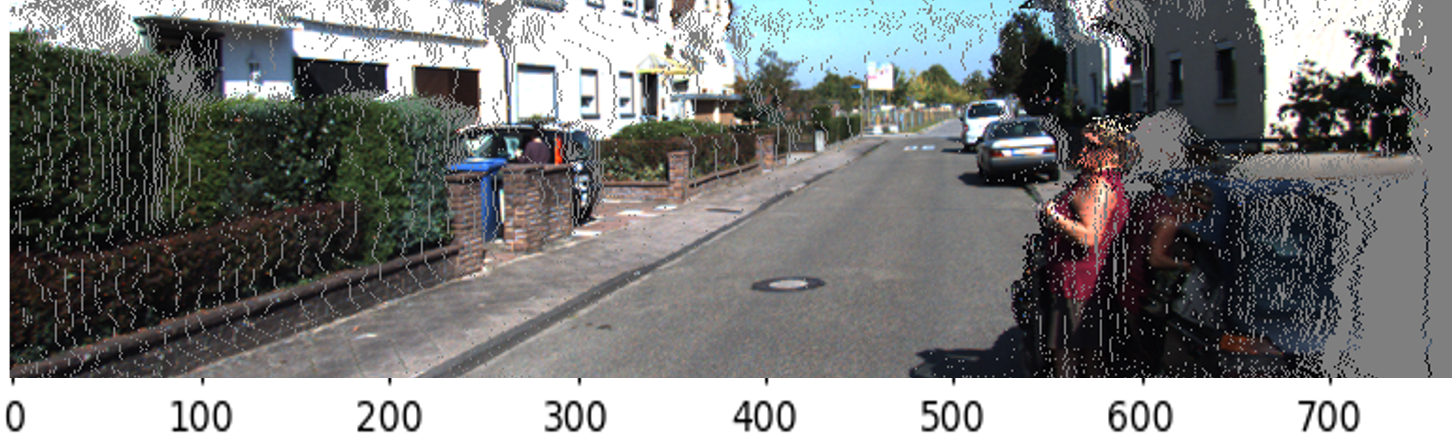} &
        \includegraphics[width=\sz\linewidth]{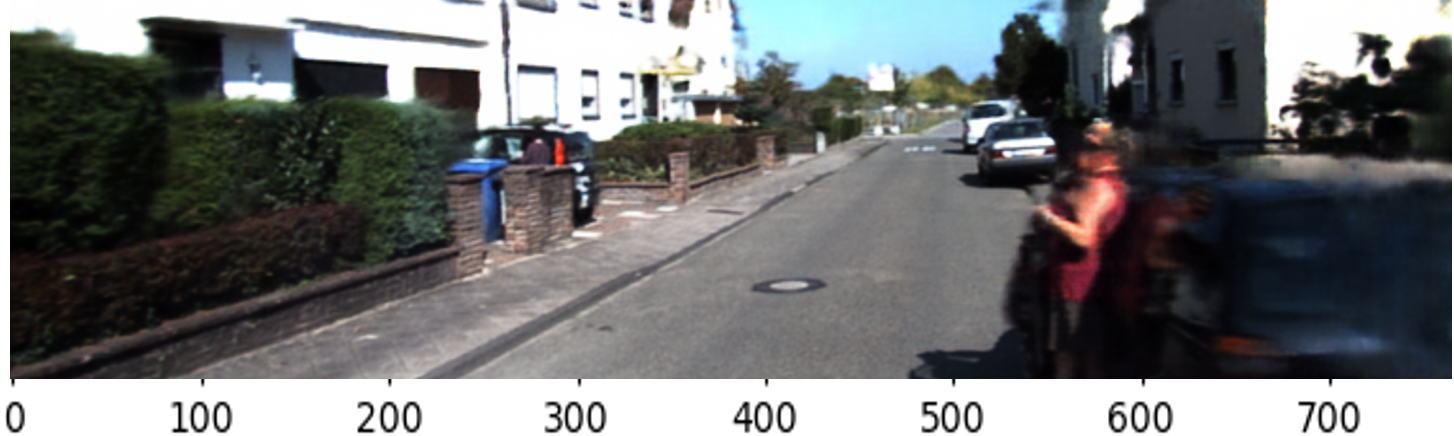} &
        \includegraphics[width=\sz\linewidth]{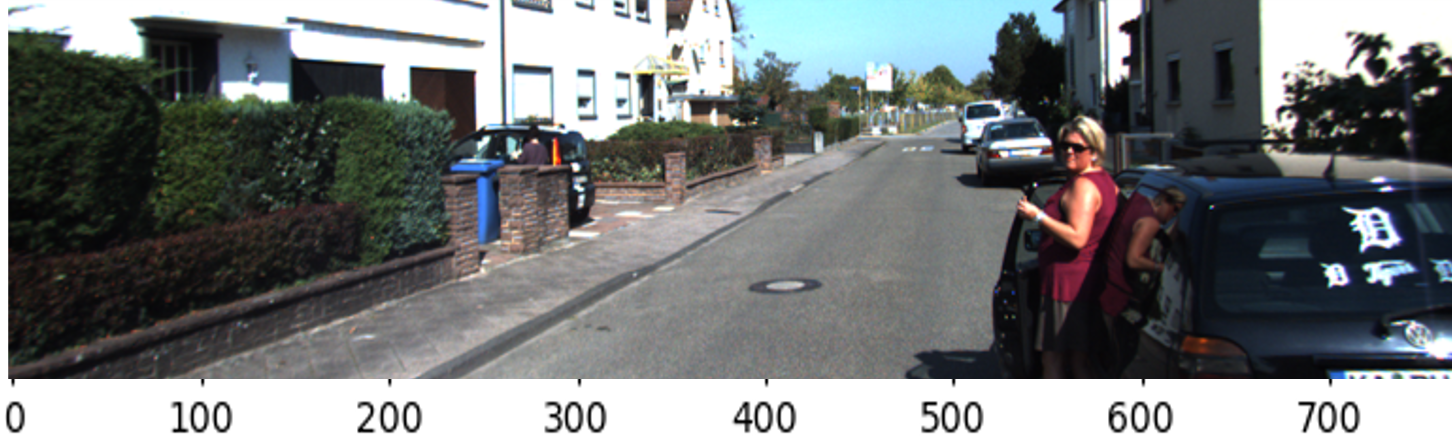}\\[\tabspace]
    \end{tabular} 
    \begin{tabular}{ccc}
        RGB 1 & \textit{DepNet} & Depth GT (for \textit{DepNet} supervision)\\
        \includegraphics[width=\sz\linewidth]{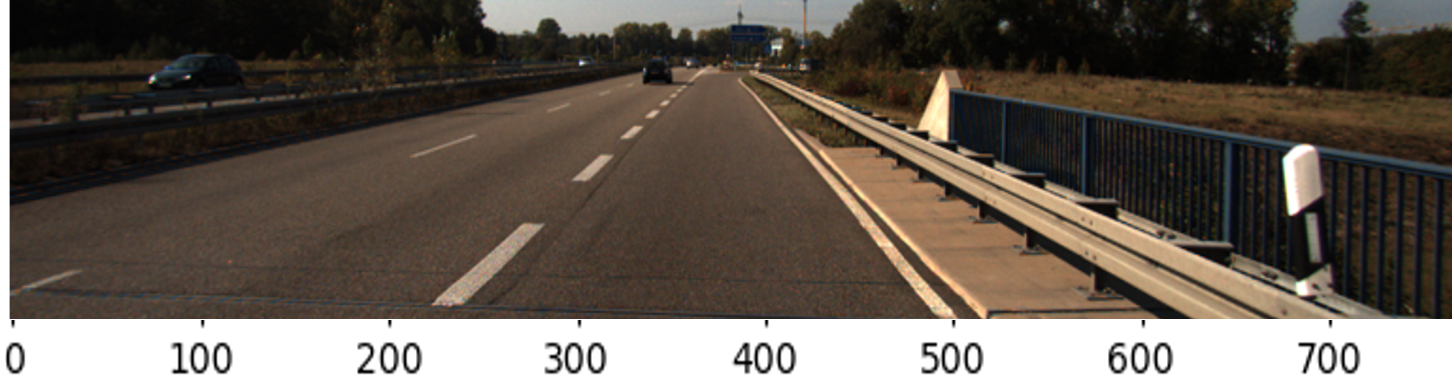} &
        \includegraphics[width=\sz\linewidth]{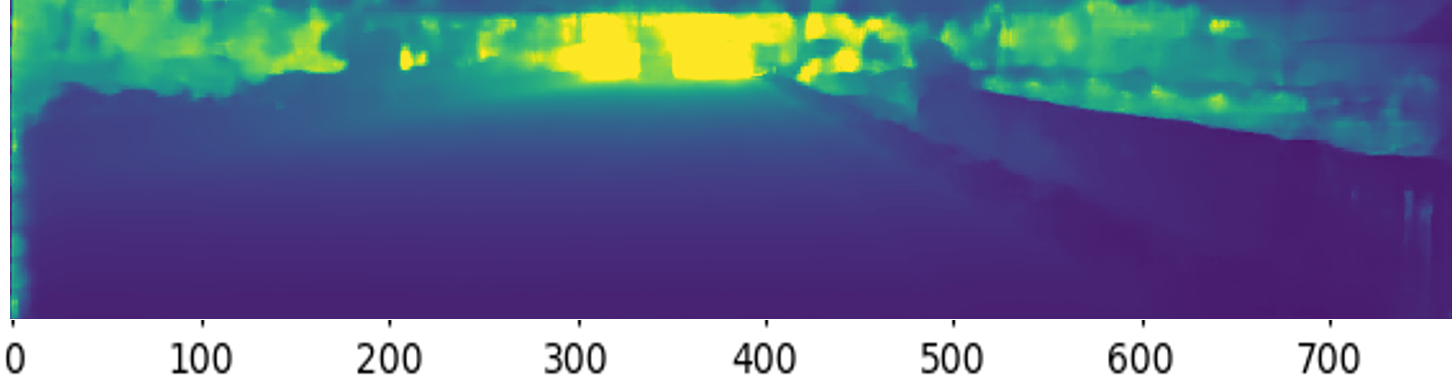} &
        \includegraphics[width=\sz\linewidth]{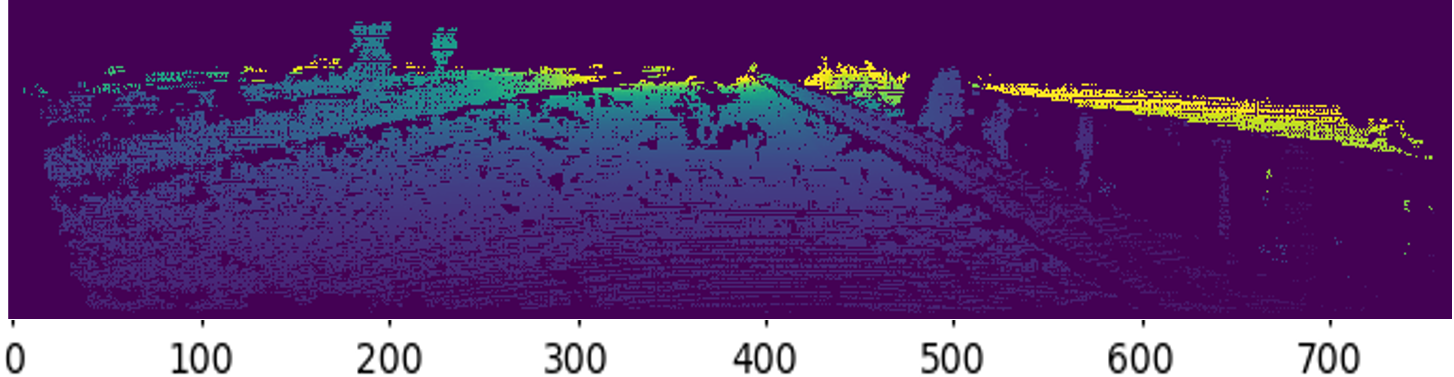}\\   
         View Transformation & \textit{SynNet} & RGB 2 (for \textit{SynNet} supervision)\\
        \includegraphics[width=\sz\linewidth]{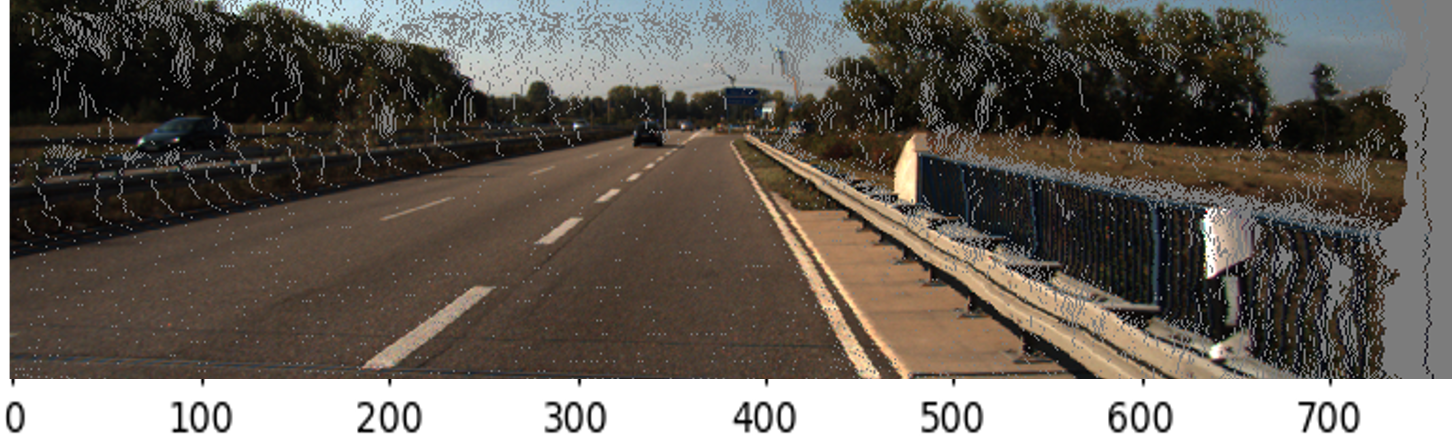} &
        \includegraphics[width=\sz\linewidth]{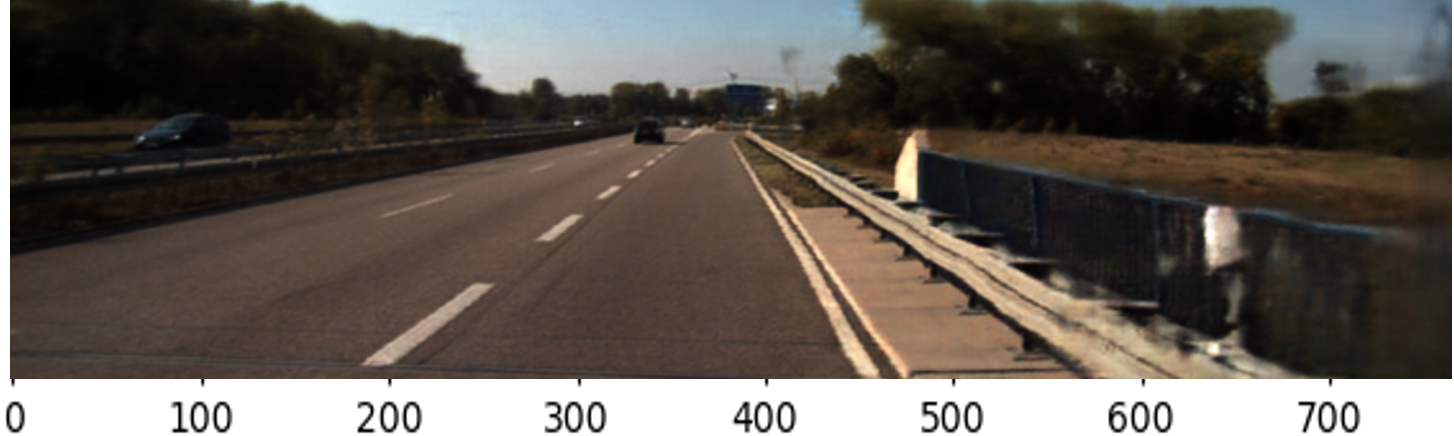} &
        \includegraphics[width=\sz\linewidth]{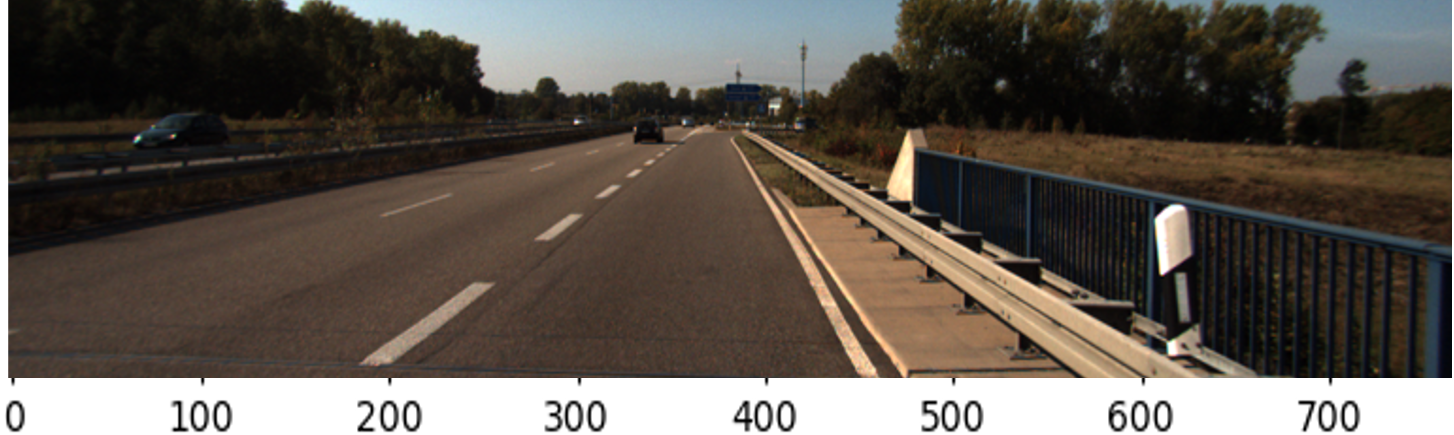}\\[\tabspace]
    \end{tabular} 
    \vspace*{-3pt}
    \caption{Qualitative predictions from each of the steps of the proposed pipeline on the KITTI \cite{Kitti} dataset. RGB 1 and RGB 2 are consecutive views or the corresponding stereo image. Color scale goes from 0 (purple) to 80 meters (yellow). For better appreciation, we recommend to zoom in on each image.}
    \label{fig:Pipeline_2}
    \vspace*{-13pt}
\end{figure*}

\clearpage
\begin{sidewaysfigure}
    \centering
    \footnotesize
    \setlength{\tabcolsep}{1pt}
	\renewcommand{\arraystretch}{0.8}
	\newcommand{\sz}{0.26}
		\newcommand{\sh}{2.55cm}
		\newcommand{\st}{2.09cm}
	    \vspace*{9cm}
	    \hspace*{-20pt}
    \begin{tabular}{cccc}
        RGB & GT & \textit{DepNet} & \textbf{NVS-MonoDepth} (Ours) \\
        \includegraphics[width=\sz\linewidth, height=\st]{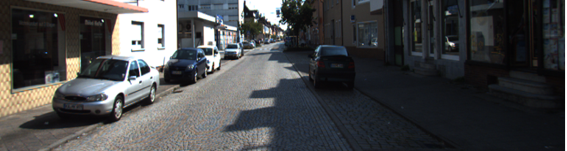} &
        \includegraphics[width=\sz\linewidth, height=\st]{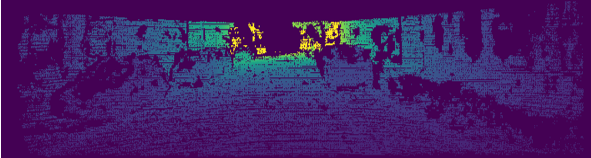} &
        \includegraphics[width=\sz\linewidth, height=\sh]{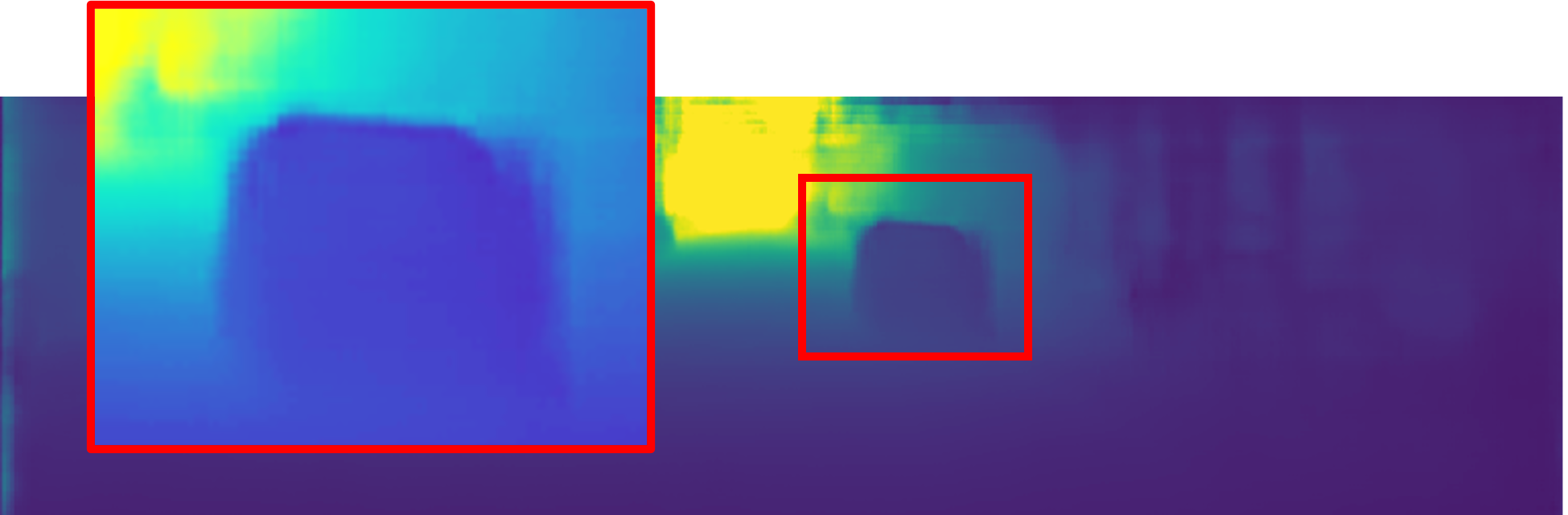} &
        \includegraphics[width=\sz\linewidth, height=2.7cm]{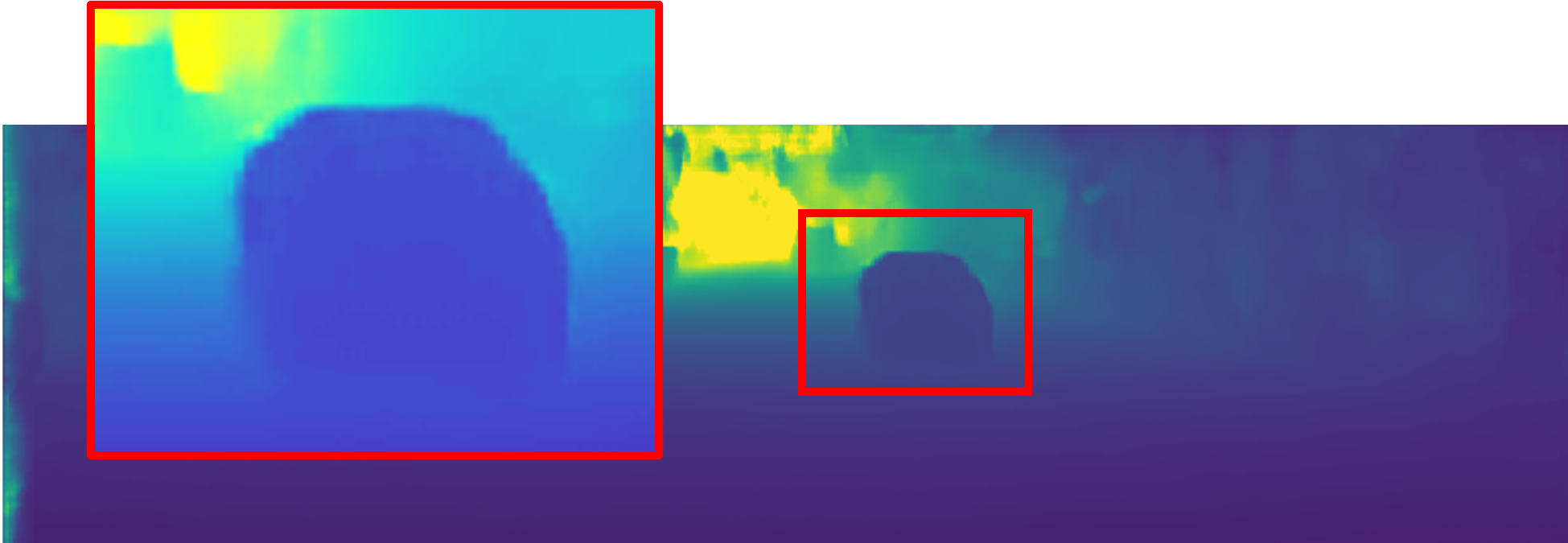} \\

        \includegraphics[width=\sz\linewidth, height=\st]{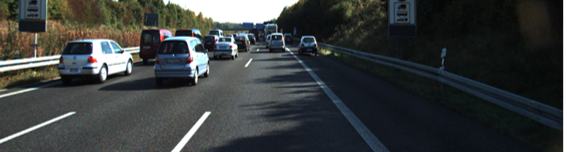} &
        \includegraphics[width=\sz\linewidth, height=\st]{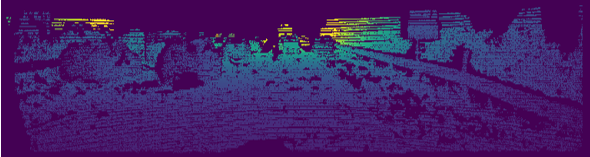} &
        \includegraphics[width=\sz\linewidth, height=\sh]{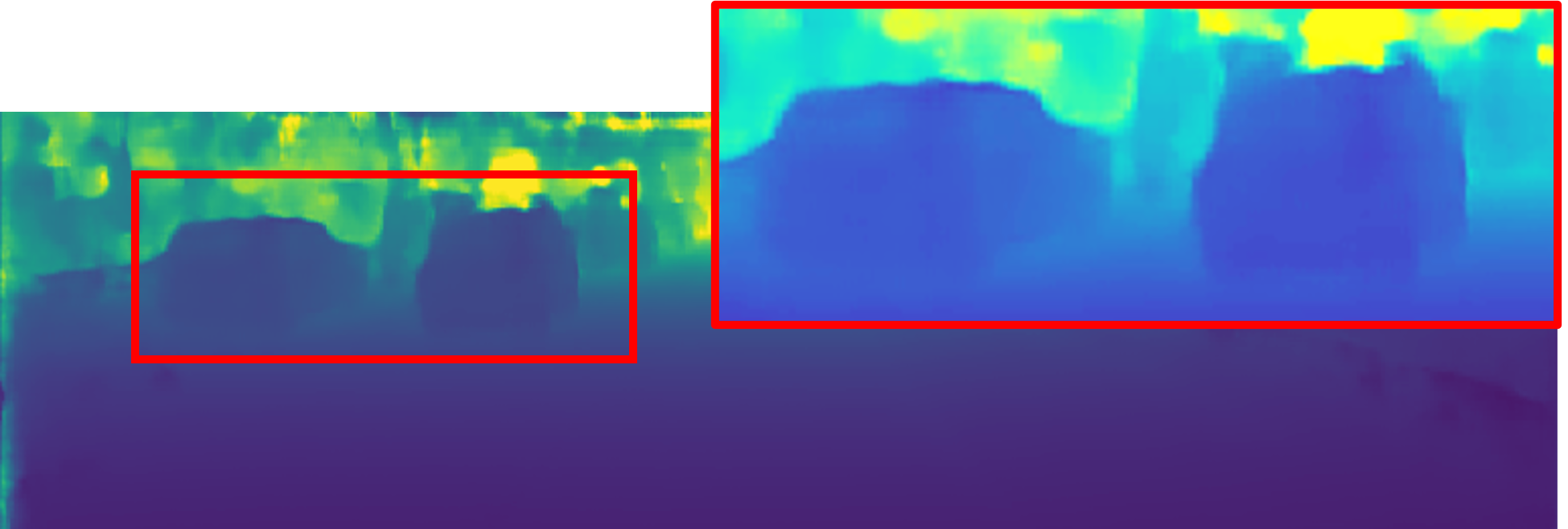} &
        \includegraphics[width=\sz\linewidth, height=\sh]{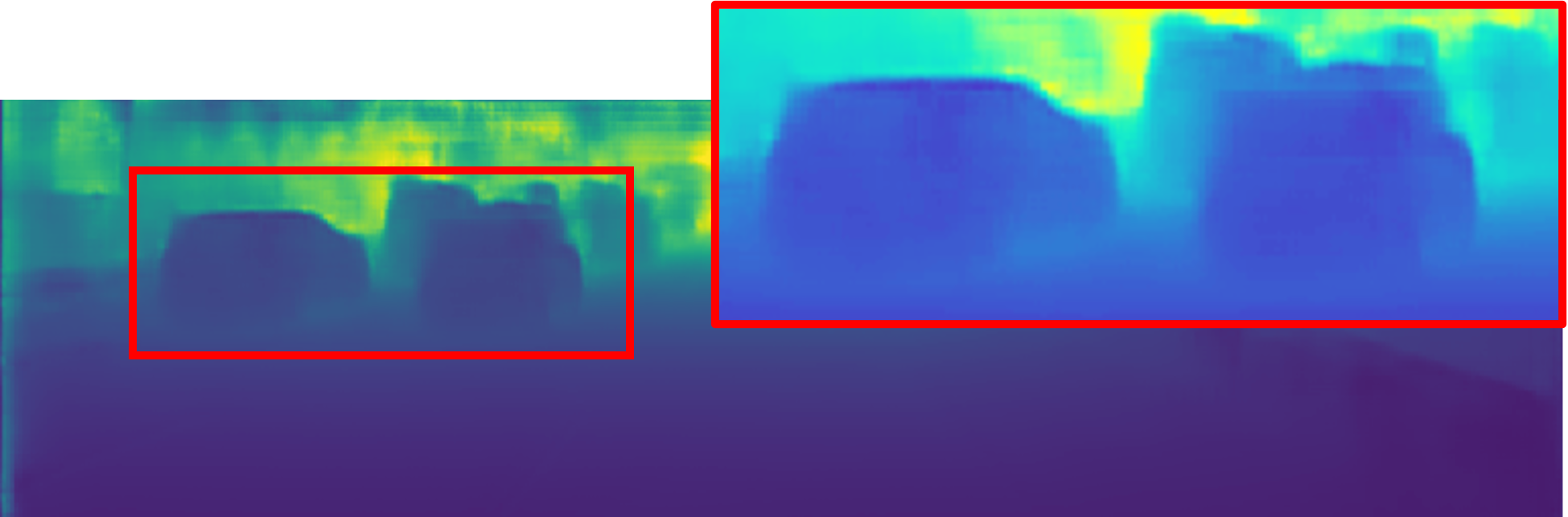} \\

        \includegraphics[width=\sz\linewidth, height=\st]{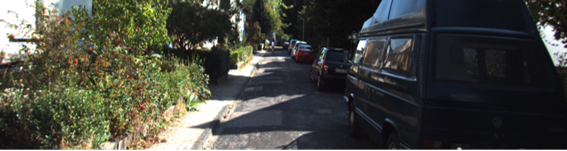} &
        \includegraphics[width=\sz\linewidth, height=\st]{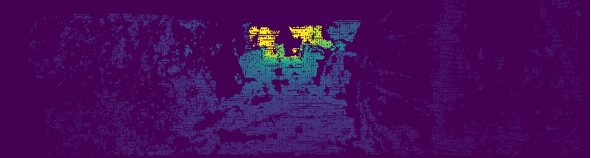} &
        \includegraphics[width=\sz\linewidth, height=\sh]{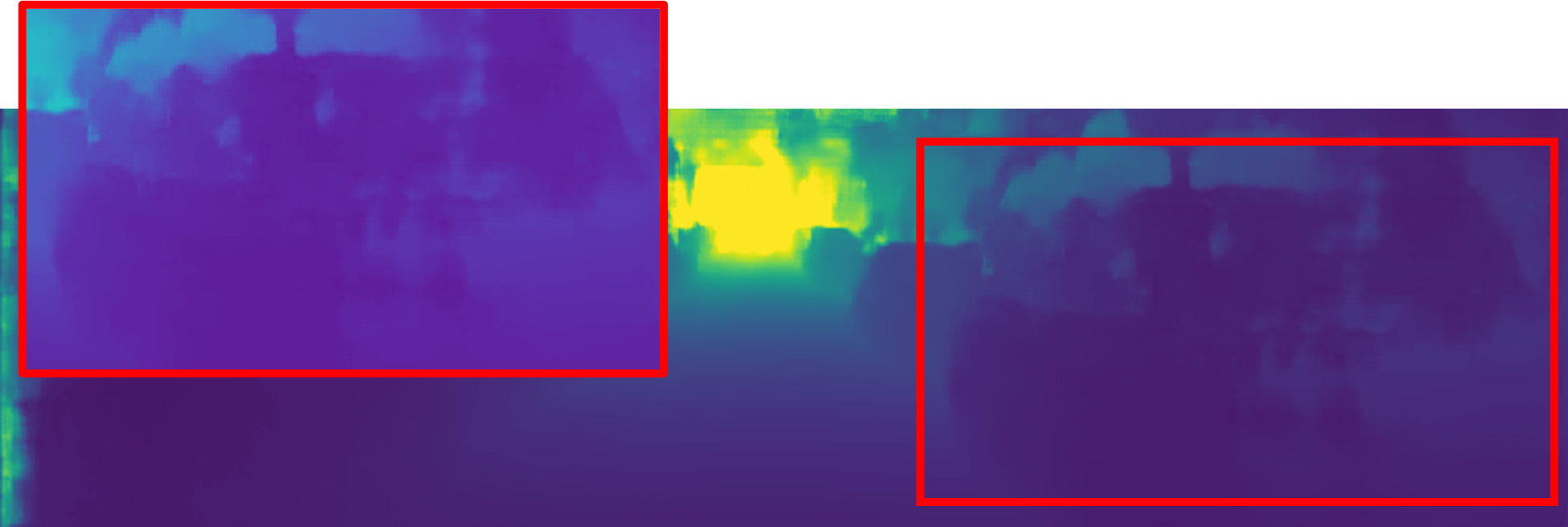} &
        \includegraphics[width=\sz\linewidth, height=\sh]{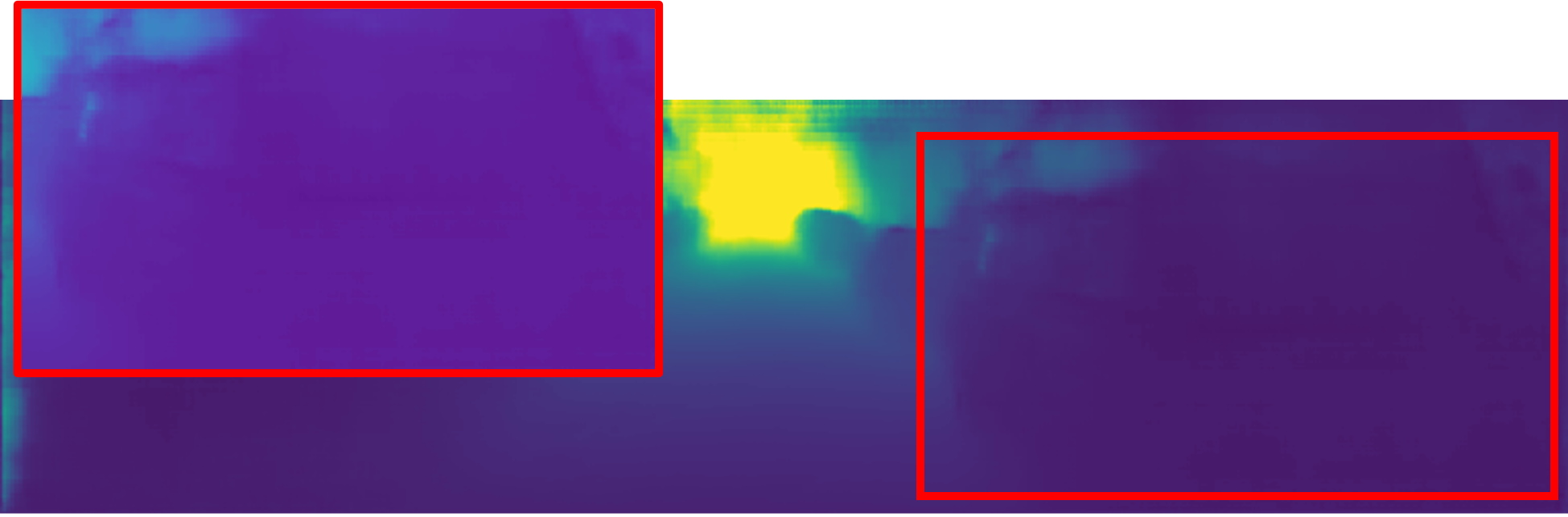} \\

        \includegraphics[width=\sz\linewidth, height=\st]{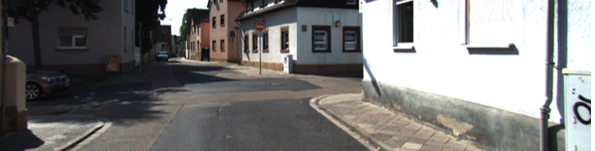} &
        \includegraphics[width=\sz\linewidth, height=\st]{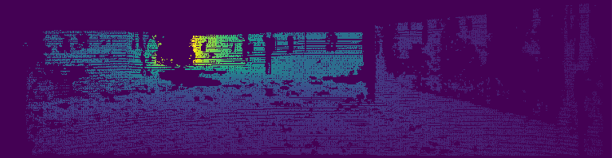} &
        \includegraphics[width=\sz\linewidth, height=\sh]{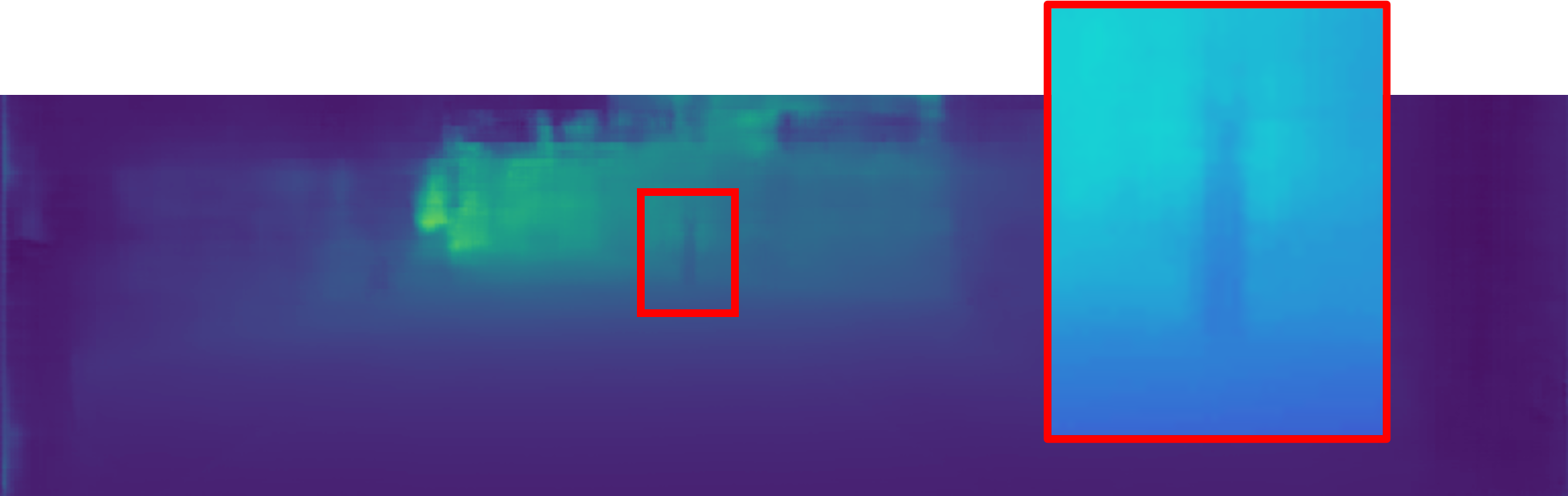} &
        \includegraphics[width=\sz\linewidth, height=\sh]{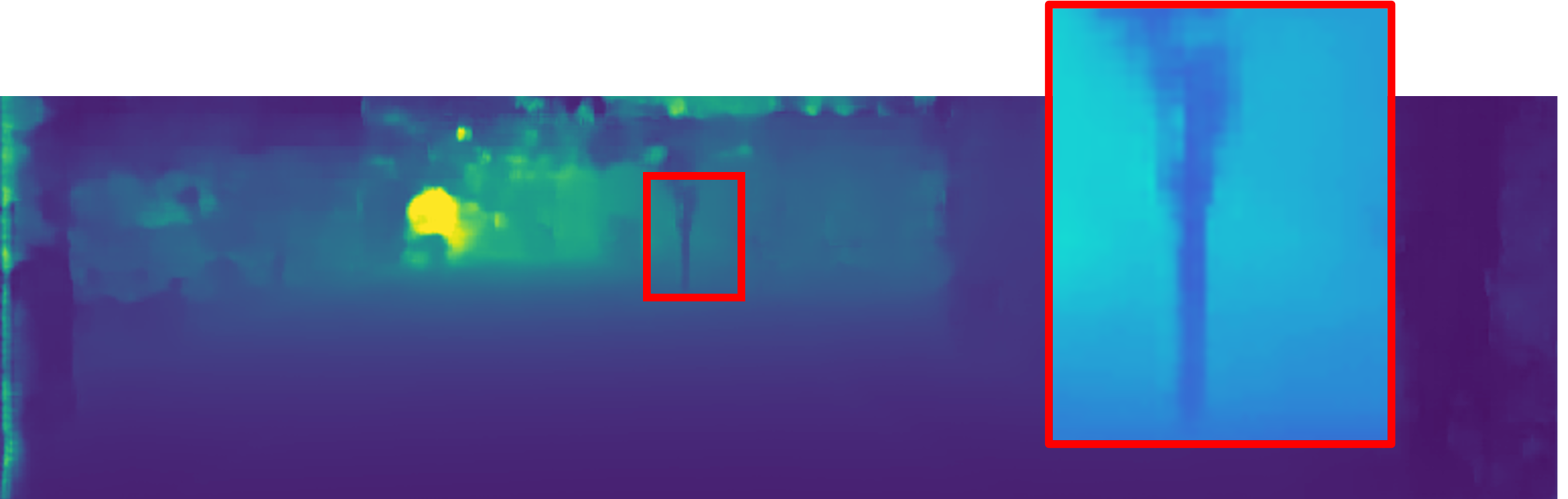} \\

        \includegraphics[width=\sz\linewidth, height=\st]{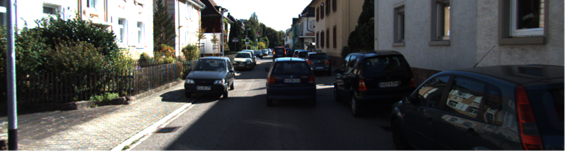} &
        \includegraphics[width=\sz\linewidth, height=\st]{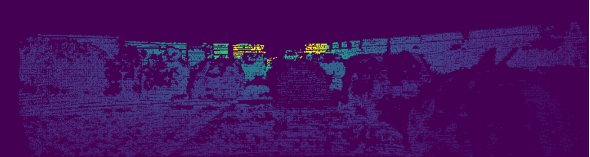} &
        \includegraphics[width=\sz\linewidth, height=\sh]{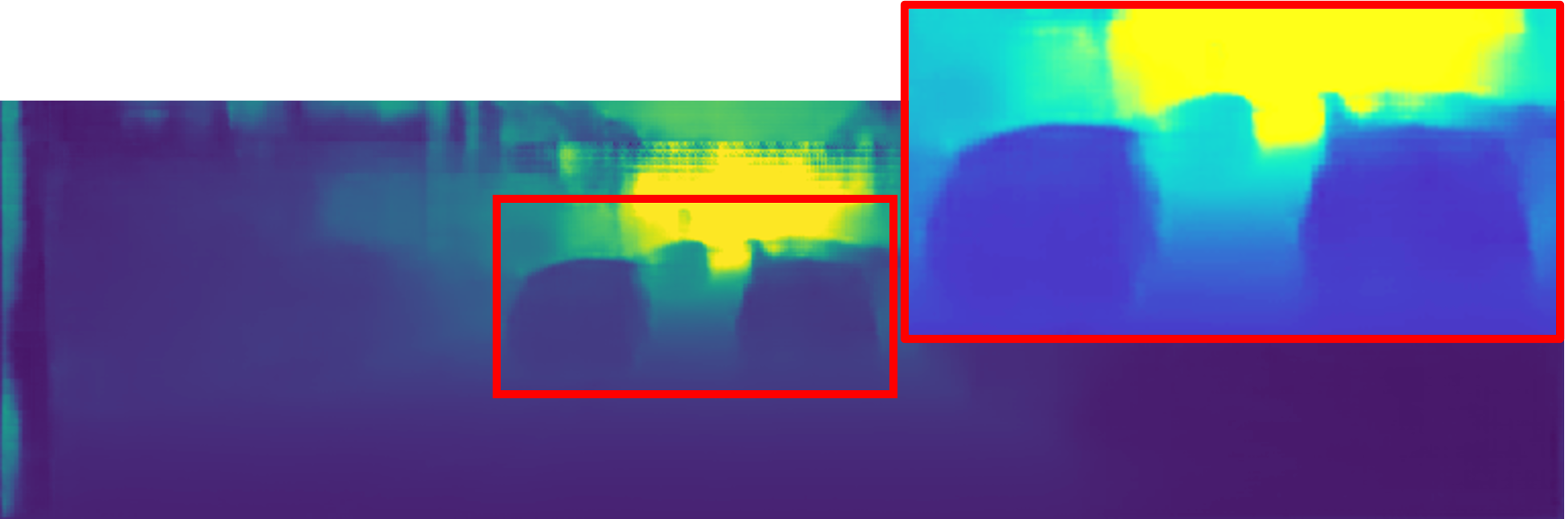} &
        \includegraphics[width=\sz\linewidth, height=\sh]{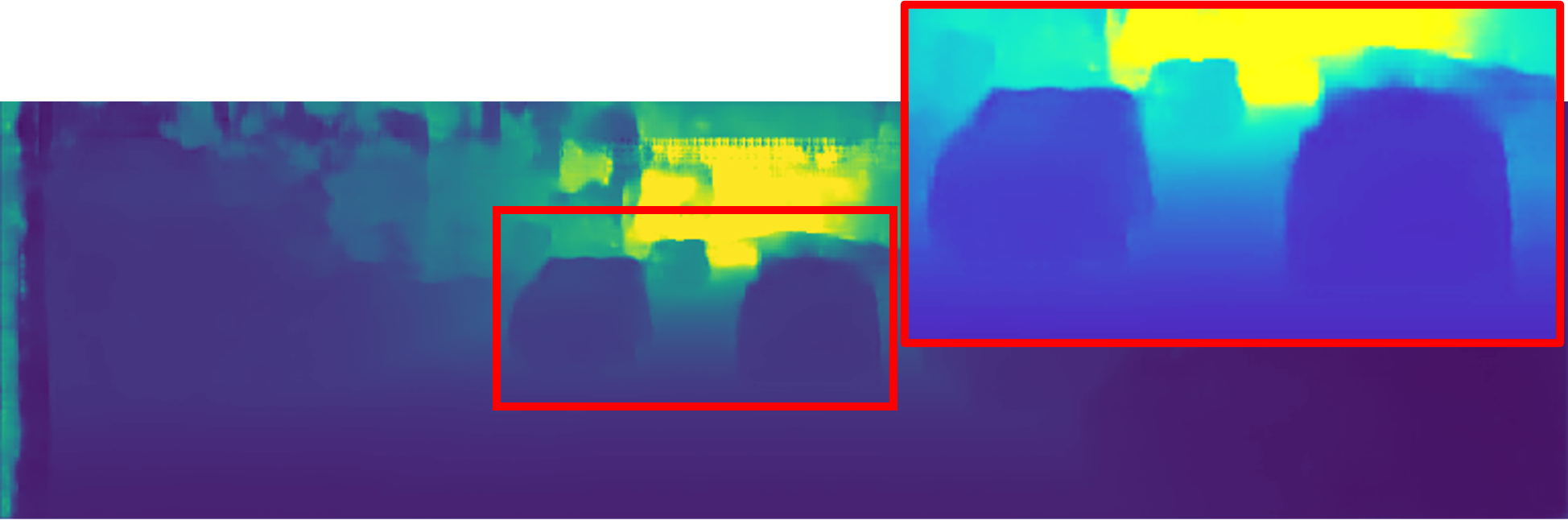} \\

        \includegraphics[width=\sz\linewidth, height=\st]{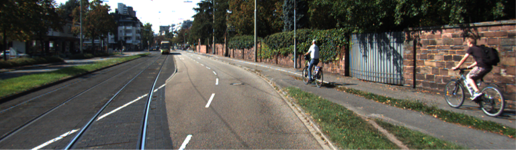} &
        \includegraphics[width=\sz\linewidth, height=\st]{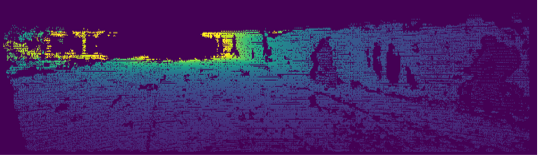} &
        \includegraphics[width=\sz\linewidth, height=\sh]{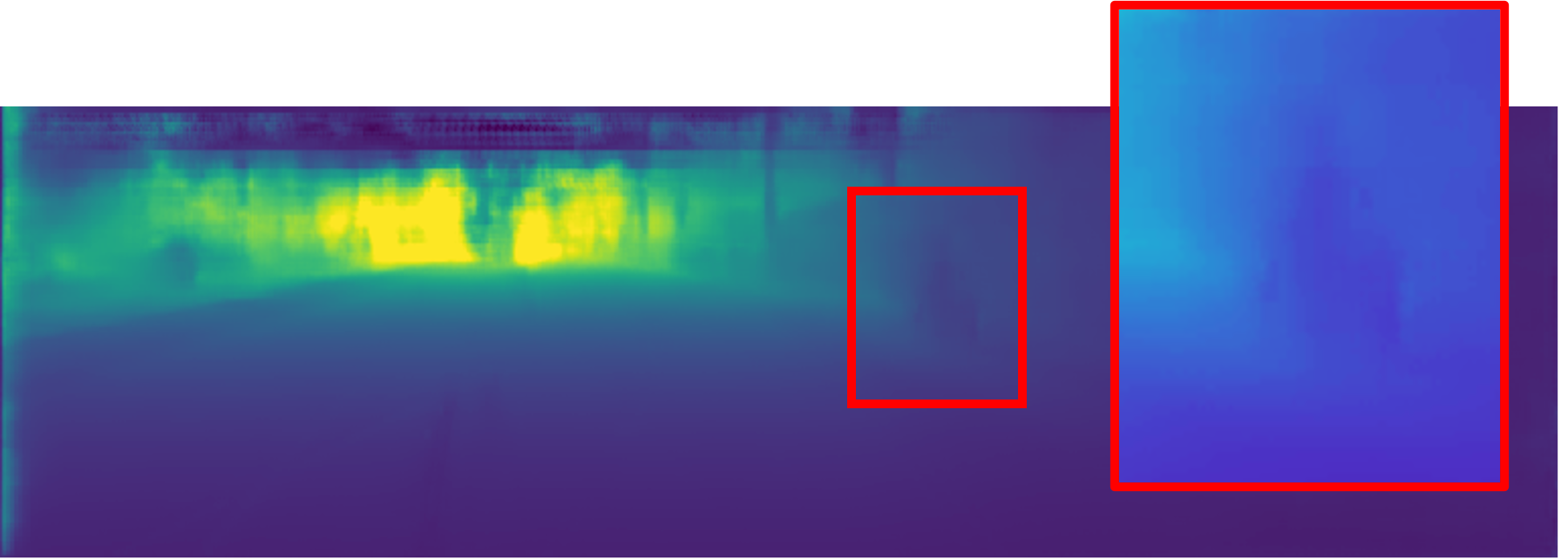} &
        \includegraphics[width=\sz\linewidth, height=\sh]{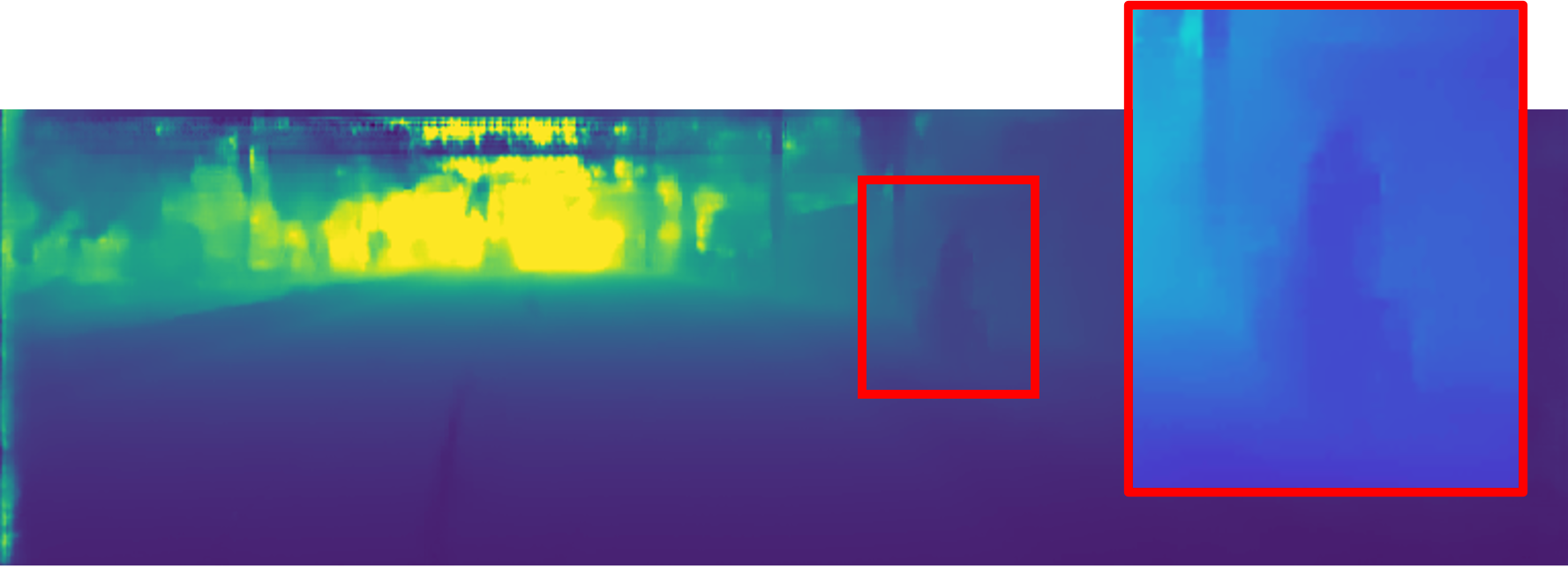} \\
    \end{tabular}
    \caption{\textbf{Qualitative comparison between the vanilla U-Net~\cite{Ronneberger2015} (\textit{DepNet}) and our \textbf{NVS-MonoDepth} variants on the KITTI \cite{Kitti} dataset.} As shown in the close-ups, our method improves the prediction of \textit{DepNet} with better sharpness and finer details. The contrast in the close-ups was adjusted for better visualization.}
    \label{fig:ablation_qualitative_supp}
    \vspace*{\figbottomspacesupp}
\end{sidewaysfigure}
\clearpage